\documentclass{article}

\PassOptionsToPackage{numbers, compress}{natbib}

\usepackage[final]{style/neurips_2022}

\usepackage{titletoc}

\usepackage[utf8]{inputenc} 
\usepackage[T1]{fontenc}    
\usepackage{hyperref}       
\usepackage{url}            
\usepackage{booktabs}       
\usepackage{amsfonts}       
\usepackage{nicefrac}       
\usepackage{microtype}      
\usepackage{xcolor}         

\usepackage{amsmath}

\usepackage{mathtools}
\usepackage{bm}
\usepackage{bbm}
\usepackage{amssymb}
\usepackage{mathtools}
\usepackage{amsthm}
\usepackage[binary-units=true]{siunitx}
\usepackage{lipsum}  
\usepackage{enumitem}

\usepackage{thmtools,thm-restate}

\usepackage[capitalise]{cleveref}

\usepackage{todonotes}

\usepackage{caption}
\usepackage{subcaption}

\usepackage{algorithm}
\usepackage[noend]{algpseudocode}

\usepackage{multirow}

\def\eqref#1{equation~\ref{#1}}

\def\1{\bm{1}}

\def\vzero{{\bm{0}}}

\def\vpsi{{\bm{\psi}}}

\def\vomega{{\bm{\omega}}}
\def\va{{\bm{a}}}
\def\vb{{\bm{b}}}
\def\vc{{\bm{c}}}

\def\vm{{\bm{m}}}

\def\vq{{\bm{q}}}

\def\vt{{\bm{t}}}
\def\vu{{\bm{u}}}
\def\vv{{\bm{v}}}
\def\vw{{\bm{w}}}
\def\vx{{\bm{x}}}

\def\vz{{\bm{z}}}

\def\evq{{q}}

\def\evx{{x}}

\def\mA{{\bm{A}}}
\def\mB{{\bm{B}}}
\def\mC{{\bm{C}}}
\def\mD{{\bm{D}}}
\def\mE{{\bm{E}}}
\def\mF{{\bm{F}}}

\def\mJ{{\bm{J}}}

\def\mP{{\bm{P}}}

\def\mR{{\bm{R}}}
\def\mS{{\bm{S}}}

\def\mU{{\bm{U}}}
\def\mV{{\bm{V}}}
\def\mW{{\bm{W}}}
\def\mX{{\bm{X}}}

\def\mZ{{\bm{Z}}}

\def\mDelta{{\bm{\Delta}}}

\def\mSigma{{\bm{\Sigma}}}

\DeclareMathAlphabet{\mathsfit}{\encodingdefault}{\sfdefault}{m}{sl}
\SetMathAlphabet{\mathsfit}{bold}{\encodingdefault}{\sfdefault}{bx}{n}

\def\sA{{\mathbb{A}}}
\def\sB{{\mathbb{B}}}

\def\sH{{\mathbb{H}}}

\def\sN{{\mathbb{N}}}

\def\sR{{\mathbb{R}}}
\def\sS{{\mathbb{S}}}
\def\sT{{\mathbb{T}}}

\def\sY{{\mathbb{Y}}}

\def\emC{{C}}

\def\ones{{\mathbf{1}}}
\def\zeros{{\mathbf{0}}}
\def\eye{{\mathbf{I}}}
\def\indicator{{\mathbbm{1}}}

\newcommand{\E}{\mathop{\mathbf{E}}}

\declaretheorem{theorem}
\declaretheorem{definition}
\declaretheorem{lemma}

\declaretheorem{proposition}

\newcommand{\changelocaltocdepth}[1]{%
  \addtocontents{toc}{\protect\setcounter{tocdepth}{#1}}%
  \setcounter{tocdepth}{#1}%
}

\newtheorem{innercustomthm}{Theorem}
\newenvironment{customthm}[1]
  {\renewcommand\theinnercustomthm{#1}\innercustomthm}
  {\endinnercustomthm}

\title{Invariance-Aware Randomized Smoothing Certificates}

\author{%
  Jan Schuchardt \\
  Technical University of Munich\\
  \texttt{j.schuchardt@tum.de} \\
  \And
  Stephan G\"unnemann \\
  Technical University of Munich \\
  \texttt{s.guennemann@tum.de} \\
}

\begin{document}




\maketitle

\begin{abstract}
Building models that comply with the invariances inherent to different domains, such as invariance under translation or rotation, is a key aspect of applying machine learning to real world problems like molecular property prediction, medical imaging, protein folding or LiDAR classification. For the first time, we study how the invariances of a model can be leveraged to provably guarantee the robustness of its predictions. We propose a gray-box approach, enhancing the powerful black-box randomized smoothing technique with white-box knowledge about invariances. First, we develop gray-box certificates based on group orbits, which can be applied to arbitrary models with invariance under permutation and Euclidean isometries. Then,  we derive provably tight gray-box certificates. We experimentally demonstrate that the provably tight certificates can offer much stronger guarantees, but that in practical scenarios the orbit-based method is a good approximation.
\end{abstract}

\section{Introduction}\label{section:introduction}
\looseness=-1
It is well-established that machine learning models are susceptible to adversarial attacks~\cite{Szegedy2014,Goodfellow2015,Akhtar2018,Xu2020}.
Even without malevolent actors, adversarial attacks can be considered worst case scenarios in environments with  noisy, erroneous or otherwise corrupted data, thus necessitating robust machine learning methods.

\looseness=-1
Invariance is a central design principle that has so far received little dedicated attention in the realm of adversarially robust machine learning.
Over the past decades, there has been ongoing research into developing machine learning models that comply with the invariances inherent to different data types and tasks.
Prominent recent examples include Deep Sets~\cite{Zaheer2017}, PointNet~\cite{Qi2017}, Group Equivariant CNNs~\cite{Cohen2016}, Spherical CNNs~\cite{Cohen2018} and Graph Convolutional Networks~\cite{Kipf2017},
but the study of invariant models significantly precedes the surge in popularity of deep learning methods~
\cite{Fukushima1979,Giles1987,Burges1996,Chapelle2001,Decoste2002,Haasdonk2005}.

\looseness=-1
For the first time, we explore the following question:
\emph{Can a-priori knowledge about invariances be leveraged in deriving provable guarantees for a model's robustness to adversarial attacks?}

\looseness=-1
Going by a loose categorization of prior work, we could adopt one of two possible approaches for our exploration:
A white-box or black-box one. White-box certificates (e.g.~\cite{Cheng2017,
Katz2017,
Ehlers2017,
Weng2018,
Wong2018,
Gehr2018,
Zhang2018,
Singh2019})
analyze a model's internals, such at its weights and non-linearities, to 
provably guarantee that a prediction does not change under adversarial attack.
Black box certificates -- specifically randomized smoothing~\cite{Liu2018, Lecuyer2019, Cohen2019} --  use statistical methods to provide provable guarantees that hold for all models sharing the same prediction probabilities under a random input distribution, irrespective of their internals.

\looseness=-1
We opt for a randomized smoothing approach, as it allows us to focus on the interplay between invariances and robustness, rather than the specific means of implementing these invariances.
By combining white-box knowledge about invariances with black-box knowledge about prediction probabilities we obtain \emph{gray-box} certificates. 
We first derive a gray-box certificate that follows a post-processing paradigm: It takes an existing black-box certificate and augments the certified region using information about the model's invariances.
This \textit{orbit-based} certificate serves as a baseline for the provably tight gray-box certificates we derive in subsequent sections.

\looseness=-1
For our exploration, we focus on models operating on spatial data, rather than structured data (e.g.~images and sequences) -- both because spatial symmetries can be elegantly formalized using algebraic concepts and because there is an ongoing trend towards using machine learning for real-world applications with inherent spatial invariances, e.g.~molecular property prediction~\cite{Duvenaud2015,Coley2017,Gilmer2017,Schutt2018,Gasteiger2019,Gasteiger2021a,Gasteiger2021b,Gao2022}, LiDAR classification~\cite{Niemeyer2014,Wang2018,He2018,Li2020r}, drug discovery~\cite{Xiong2019,Ramsauer2020,Cai2020}, 
particle physics~\cite{Baldi2014,Guest2018,Bogatskiy2020}
and protein folding~\cite{Qian1988,
Fariselli2001,
Xu2019,
AlQuraishi2019,
Ruff2021}.

\looseness=-1
Our main contributions are
\vspace{-0.2cm}
\begin{itemize}
    \itemsep0pt
    \item the first study on the interplay of invariance and certifiable robustness,
    \item a principled method for deriving tight invariance-aware randomized smoothing certificates,
    \item tight certificates for models invariant to translations and rotations.
\end{itemize}
\vspace{-0.15cm}
We further demonstrate experimentally that the orbit-based certificates offer a good approximation of our tight certificates, if the variance of the smoothing distribution is small.

\section{Related work}

\looseness=-1
\textbf{Invariant machine learning.}
Given the diversity of approaches to invariant machine learning,
and the fact that our approach is model-agnostic,
we refrain from attempting a survey and instead refer to~\cite{Bronstein2021} for a principled, high-level introduction into the realm of learning with invariances and equivariances.

\looseness=-1
\textbf{Invariance and robustness.}
Two recent empirical studies~\cite{Tramer2020,Kamath2021} demonstrate that data augmentation meant to increase robustness to $\ell_p$-norm adversarial attacks reduces robustness to semantically meaningful transformations (e.g.\ rotation) and vice-versa, suggesting an inherent invariance-robustness trade-off.
However, neither study models that are invariant by design.
In~\cite{Singla2021}, the negative effect of translation-invariance on the robustness of image classifiers is investigated.
Note that our work is not meant to resolve potential trade-offs, but to tightly bound the actual robustness of models.

\looseness=-1
\textbf{Gray-box certificates.}
While we propose the first gray-box certificate for invariant models, there exists prior work on combining white-box knowledge with black-box certification.
In~\cite{Mohapatra2020} and~\cite{Levine2020}, knowledge about a classifier's gradients is used to derive tighter randomized smoothing certificates.
In~\cite{Schuchardt2021}, knowledge about a graph neural network's receptive fields is used to derive collective gray-box certificates for multiple predictions.
In~\cite{Scholten2022}, the message-passing scheme of graph neural networks is used for gray-box certification against adversaries that control all features of multiple nodes.

\looseness=-1
\textbf{Adversarial attacks on spatial data.}
Adversarial attacks on spatial data
either modify~\cite{Su2018,Liu2019,Zhou2020,Tsai2020,Lee2020,Zhao2020,Wen2020,Kim2021,Sun2021}, insert~\cite{Kim2021,Xiang2019,Arya2021}, or delete~\cite{Wicker2019,Yang2019,Zheng2019} points in space and have been particularly actively studied for point cloud-classification.
While the earliest work simply adopted gradient-based attacks from the image domain~\cite{Su2018}
more recent work has developed a rich assortment of domain-specific methods, for example to preserve object smoothness~\cite{Wen2020,Huang2020}, leverage critical points~\cite{Wicker2019,Yang2019} 
or craft physically realizable attacks 
(e.g.\ to attack models through LiDAR sensors) 
~\cite{Tsai2020,Cao2019,Tu2020,Abdelfattah2021}.
Of particular note are attacks via isometries (e.g. rotations and translations)~\cite{Zhao2020iso,Shen2021}, whose effectiveness motivates the use of  invariant models.
Note that we certify the robustness of invariant models to arbitrary point modification attacks -- not just isometry attacks (see also ``orthogonal research directions'' below).
Like in other domains, empirical defenses have been proposed~\cite{Liu2019,Zhou2019,Dong2020} and subsequently broken~\cite{Liu2020,Ma2020,Sun2020}, motivating the development of black-box~\cite{Liu2021,Denipitiyage2021,Chu2022} and white-box~\cite{Lorenz2021} robustness certificates for spatial data.
It should be noted that Gaussian randomized smoothing, without invariance information, has already been used in prior work -- either as a baseline~\cite{Liu2021} or as a special case of the respective certificate~\cite{Chu2022}.

\looseness=-1
\textbf{Orthogonal research directions.}
One related but orthogonal research direction is transformation-specific certification~\cite{Singh2019,Chu2022,Lorenz2021,Balunovic2019,Ruoss2021,Mohapatra2021,Li2021,Fischer2021,Alfarra2022,Muravev2022}.
There, a model is assumed to potentially change its prediction under adversarial parametric transformations (e.g.~rotations)
and one certifies robustness for specific parameter ranges.
White-box methods~\cite{Singh2019,Lorenz2021,Balunovic2019,Ruoss2021,Mohapatra2021} over-approximate the set of inputs reachable by a transformation and then propagate it through a model using existing white-box techniques.
Black-box methods -- namely transformation-specific randomized smoothing ~\cite{Li2021,Fischer2020} -- randomize the transformation parameters to provide robustness guarantees for arbitrary models via statistical methods.
Later work generalized this principle to spatial data~\cite{Chu2022}, vector-field deformations~\cite{Alfarra2022} and multiplicative parameters~\cite{Muravev2022}.
Different from these methods, we assume our model to be invariant under a set of transformations, i.e.~never change its prediction. We use this property as a tool for certification against arbitrary perturbations.
Aside from that, there exist white-box certification techniques for specific \textit{operations with invariances} (e.g.\ global max-pooling~\cite{Lorenz2021}, message passing~\cite{Zugner2019,Zugner2020} and batch normalization~\cite{Boopathy2019}).
But prior work does not use or even discuss this property -- it treats these invariant operations as coincidental building blocks of the models it is trying to certify.
Our work is the first to study  \textit{invariance itself} in the context of provable robustness and how to leverage it for certification.

\section{Background}
 
\subsection{Randomized smoothing}\label{section:background_randomized_smoothing}
\looseness=-1
Randomized smoothing is a black-box certification technique that can be adapted to
various data types, tasks and threat models 
\cite{Lee2019,Bojchevski2020,Levine2021,
Jia2019,Chiang2020,Kumar2021,Fischer2021,Li2020,Wang2020,Rosenfeld2020}.
Instead of directly certifying a classifier $g$, it constructs a smoothed classifier $f$ that returns the most likely prediction under random perturbations of its input. It then certifies the robustness of this smoothed classifier.
We present the tight Gaussian smoothing certificate derived by~\citet{Cohen2019} 
and its generalization to matrix data~\cite{Liu2021,Chu2022}.

\looseness=-1
Assume a continuous $\smash{\left(N \times D\right)}$-dimensional input space $\smash{\sR^{N \times D}}$, label set $\smash{\sY}$ and \textit{base classifier} $\smash{g : \sR^{N \times D} \rightarrow \sY}$.
Let $\smash{\mu_\mX(\mZ) = \prod_{d=1}^D\mathcal{N}\left(\mZ_{:,d} \mid \mX_{:,d}, \sigma^2 \eye_N\right)}$ be the isotropic matrix normal distribution with mean $\mX$ and standard deviation $\sigma$. Let $\smash{p_{\mX,y} = \Pr_{\mZ \sim \mu_\mX} \left[g(\mZ) = y\right]}$ be the probability of $g$ predicting class $y$ under this \textit {smoothing distribution}.
One can then define a \textit{smoothed classifier} 
$\smash{f(\mX) = \mathrm{argmax}_{y \in \sY} p_{\mX,y}}$ that returns the most likely prediction of $g$ under $\mu_\mX$.

\looseness=-1
Let $y^* = f(\mX)$ be a smoothed prediction and $\smash{\mX' = \mX + \mDelta}$
a
perturbed input.
One can show $\smash{f(\mX') = y^*}$ by proving that,
for perturbed input $\mX'$, $\smash{y^*}$ is more likely than all other classes combined. That is, $p_{\mX', y^*} > 0.5$.
A tighter certificate can be obtained by proving that $\smash{y^*}$ is more likely than the second most likely class, i.e.~$p_{\mX', y^*} > \max_{y' \neq y^*} p_{\mX', y'}$.
For the sake of exposition, we use the first approach throughout the main text and generalize all results to the second one in~\cref{appendix:multiclass}.
One can lower-bound $p_{\mX', y^*}$ by finding the \textit{worst-case classifier} from a set of functions $\sH$ with $g \in \sH$:
\begin{equation}\label{eq:cohen_worst_case_relaxation}
    p_{\mX', y^*} \geq \min_{h \in \sH} \Pr_{\mZ \sim \mu_{\mX'}} \left[h(\mZ) = y^*\right].
\end{equation}
For
$\smash{\sH = \left\{h : \sR^{N \times D} \rightarrow \sY \mid
  \Pr_{\mZ \sim \mu_\mX} \left[h(\mZ) = y^*\right]
   \geq
  p_{\mX,y^*}
\right\}}$,
the
classifiers that are at least as likely as $g$ to classify $\mX$ as $y^*$, the exact solution is given by the Neyman-Pearson lemma~\cite{Neyman1933} (see~\cref{appendix:neyman_pearson}). 
The optimal value is 
$\Phi\left(\Phi^{-1}\left(p_{\mX,y^*}\right) - \frac{||\Delta||_2}{\sigma}\right)$, where $\Phi$ is the standard-normal CDF and $||\cdot||_2$ is the Frobenius norm.
If $\smash{||\mDelta||_2 < \sigma \Phi^{-1}\left(p_{\mX,y^*}\right)}$, then $\smash{p_{\mX', y^*} > 0.5}$ and the prediction is provably robust.
Because~\cref{eq:cohen_worst_case_relaxation} was solved exactly, this is a \textit{tight certificate}, i.e.\ the best possible certificate that can be obtained by only using black-box knowledge about prediction probability $p_{\mX,y^*}$
.

\looseness=-1
\textbf{Probabilistic certificates.}
For neural networks, the prediction probability $p_{\mX,y^*}$ can usually not be computed analytically. Instead, one has to use Monte Carlo sampling to compute a lower confidence bound $\underline{p_{\vx',y^*}}$ that holds with high probability $1 - \alpha$. The resulting certificate is a probabilistic one.

\subsection{Group invariance}\label{section:background_invariance}
Let $\smash{f : \sR^{N \times D} \rightarrow \sY}$ be a classifier and $\sT$ a group, i.e.\ a set and associated operator $\cdot : \sT \times \sT \rightarrow \sT$ that is closed and associative under the operation, has an inverse $t^{-1}$ for each element $t \in \sT$ and features an identity element $e$.
Further let $\sT$ act on the input space via a \textit{group action} $\circ : \sT \times \sR^{N \times D} \rightarrow \sR^{N \times D}$ that preserves the group structure, i.e.\ $(t \cdot t') \circ \mX = t \circ (t' \circ \mX)$.
Group actions naturally partition the domain into \textit{orbits}, i.e.\ sets that can be reached by applying group actions to inputs:
\begin{restatable}[Orbits]{definition}{definitionequivalenceclasses}\label{definition:equivalence_class}
The orbit of an input $\mX \in \sR^{N \times D}$ w.r.t.\ a group $\sT$ is
$[\mX]_\sT = \left\{t \circ \mX \mid t \in \sT \right\}$.
\end{restatable}
Classifier $f$ is said to be invariant under group $\sT$ if $\forall \mX \in \sR^{N \times D}, \forall \mX' \in [\mX]_\sT: f(\mX) = f(\mX')$.

\subsection{Haar measures}\label{section:haar_measure}
Haar measures~\cite{Haar1933} are a generalization of Lebesgue measures for integration over groups.
Their key property is invariance, meaning the group operator does not affect the measure.
\begin{definition}[Right Haar measure]
    Let $\eta$ be a finite, regular measure on the Borel subsets of $\sT$.
    If $\eta(\sS) = \eta(\{s \cdot t \mid s \in \sS\})$ for all $t \in \sT$ and Borel subsets $\sS \subseteq \sT$, 
    then $\eta$ is a right Haar measure.
\end{definition}
For the groups we consider, the Haar measure is unique up to a multiplicative constant~\cite{Cartan1940,Alfsen1963}.
In~\cref{section:certification_methodology} we use Haar measures as a notational tool to summarize certificates for different groups.
However, an understanding of measure theory is not required to follow any part of the derivations.

\looseness=-1
\section{Problem setting}\label{section:problem_setting}
We consider a similar setup to that described in~\cref{section:background_randomized_smoothing}, i.e.~we have a smoothed classifier $\smash{f : \sR^{N \times D} \rightarrow \sY}$ that is the result of randomly smoothing a base classifier $g$ with an isotropic matrix normal distribution $\mu_\mX$.
Given a clean prediction $\smash{y^* = f(\mX)}$ with clean prediction probability $\smash{p_{\mX, y^*}}$, we want to determine whether
$\smash{f(\mX') = y^*}$ for an adversarially perturbed input $\smash{\mX' = \mX + \mDelta}$.

Different from all prior work, we additionally assume that the \emph{base classifier} $g$ is invariant under a group $\sT$.
This does not necessarily mean that the \emph{smoothed classifier} $f$ shares the same invariances.
But, we can use the isotropy of smoothing distribution $\mu_\mX$ to prove (see~\cref{appendix:invariance_preservation}) that randomized smoothing preserves invariance under Euclidean isotropies (rotation, reflection and translation) and permutation, which we shall leverage in~\cref{section:postprocessing}.
\begin{restatable}{theorem}{invariancepreservation}\label{theorem:invariance_preservation}
    Let base classifier $g : \sR^{N \times D} \rightarrow \sY$ be invariant under group $\sT$ with $\sT \subseteq \mathit{E}(D)$
    or $\sT \subseteq \mathit{S}(N)$, 
    where $\mathit{E}(D)$ is the Euclidean group  and $\mathit{S}(N)$ is the permutation group.
    Then the isotropically smoothed classifier $f$, as defined in~\cref{section:background_randomized_smoothing}, is also invariant under $\sT$.
\end{restatable}
Note that the above result also holds for subgroups of $\mathit{E}(D)$, such as the translation group $\mathit{T}(D)$, the rotation group $\mathit{SO}(D)$ and the roto-translation group $\mathit{SE}(D)$.
We define these groups and their actions on $\sR^{N \times D}$ more formally in~\cref{appendix:group_definitions}.

\section{Orbit-based gray-box certificates}\label{section:postprocessing}

\begin{figure}
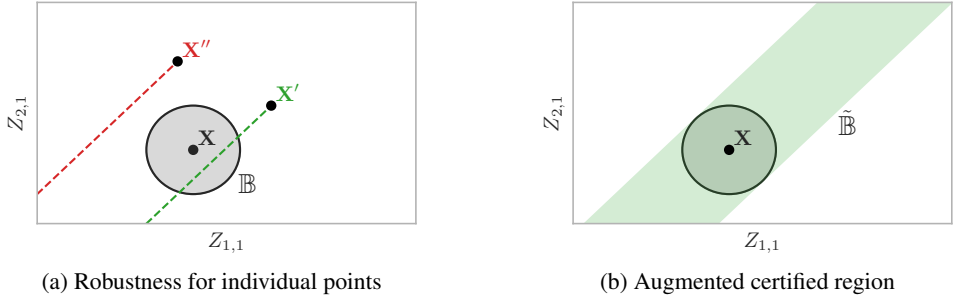

    \vspace{-0.7cm}
     \centering
     \begin{subfigure}[b]{0.49\textwidth}
         \centering
         \resizebox{0.8\textwidth}{!}{\input{figures/explainy_figure/preprocessing.pgf}}
         \caption{Robustness for individual points}
         \label{fig:explainy_figure_preprocessing}
     \end{subfigure}
     \hfill
     \begin{subfigure}[b]{0.49\textwidth}
         \centering
         \resizebox{0.8\textwidth}{!}{\input{figures/explainy_figure/postprocessing.pgf}}
         \caption{Augmented certified region}
         \label{fig:explainy_figure_postprocessing}
     \end{subfigure}
     \caption{Orbit-based gray-box certificate for translation invariance and $N=2, D=1$. a.) Input $\mX'$ can be translated into certified region $\sB$. It is not an adversarial example. Input $\mX''$ can not be translated into $\sB$. It might be an adversarial example. b.) The prediction $y^* = f(\mX)$ is certifiably robust to all perturbed inputs from augmented region $\tilde{\sB}$, the union over all translations of $\sB$.}
     \label{fig:explainy_figure}
    \vspace{-0.5cm}
\end{figure}

\looseness=-1
Since we are the first to consider robustness certification for invariant models, we begin by defining a baseline that other certificates can be benchmarked against:
It is based on the insight that certificates correspond to sets of \textit{inputs} $\sB \subseteq \sR^{N \times D}$ that preserve clean prediction $y^*$, and group $\sT$ corresponds to sets of \textit{transformations} that preserve predictions.
By transitivity, a perturbed input $\mX'$ that can reach $\sB$ via a transformation from $\sT$, i.e.\ $\exists t \in \sT: t \circ \mX' \in \sB$, 
must fulfill $f(\mX') = y^*$ and 
cannot be an adversarial example.
Combining all such inputs, i.e.\ combining the orbits of elements of $\sB$ (see~\cref{definition:equivalence_class}), yields an augmented certified region $\tilde{\sB} \supseteq \sB$  (proof in \cref{appendix:postprocessing_main_proof}):
\begin{restatable}[Orbit-based certificates]{theorem}{postprocessingmain}\label{theorem:postprocessing_main}
    Let $f \in \sR^{N \times D} \rightarrow \sY$ be invariant under a group $\sT$.
    Let $y^* = f(\mX)$ be a prediction that is certifiably robust to a set of perturbed inputs $\sB \subseteq \sR^{N \times D}$,
    i.e. $\forall \mZ \in \sB: f(\mZ) = y^*$.
    Let  $\tilde{\sB} =  \cup_{\mZ \in \sB} [\mZ]_\sT$.
    Then $\forall \mX' \in \tilde{\sB} : f(\mX') = y^*$.
\end{restatable}
This orbit-based approach is illustrated in~\cref{fig:explainy_figure}.
While we focus on randomized smoothing, \cref{theorem:postprocessing_main} holds for arbitrary models and certified regions $\sB$.
However, obtaining $\sB$ via other means may not be possible. For instance, there exist no white-box certificates for rotation-invariant models.

Because randomized smoothing preserves invariance to Euclidean isometries and permutation (see~\cref{theorem:invariance_preservation}), we may apply the orbit-based approach
with $\sT \subseteq \mathit{E}(D)$ or $\sT \subseteq \mathit{S}(N)$ 
to the certified region $\sB = \left\{\mX + \mDelta \mid ||\mDelta ||_2 < r \right\}$ with $r = \sigma \Phi^{-1}\left(p_{\mX,y^*}\right)$ that we derived in~\cref{section:background_randomized_smoothing}.

While $\tilde{\sB} =  \cup_{\mZ \in \sB} [\mZ]_\sT$,
is a valid certificate, 
one may desire an equivalent, but more explicit characterization of the certified region $\tilde{\sB}$ to determine whether a specific perturbed input preserves the prediction $y^*$.
We discuss such explicit characterizations in~\cref{appendix:postprocessing}.
In particular, 
translation invariance (i.e.\ $\sT = \mathit{T}(D)$)
leads to certified region
$\tilde{\sB} = \left\{\mX + \mDelta \mid || \mDelta - \ones_N \overline{\mDelta}||_2 <  r \right\}$,
where $\overline{\mDelta} \in \sR^{1 \times D}$ are column-wise averages and $r$ is defined as above.
Rotation invariance (i.e.\ $\sT = \mathit{SO}(D)$) leads to certified region
$\tilde{\sB} = \left\{\mX' \mid || \mX' \mR^T - \mX ||_2 <  r \right\}$,
where $\mR$  is an optimal rotation matrix defined by the singular value decomposition of $\mX^T \mX' $~\cite{Schonemann1966,Kabsch1976} (see~\cref{appendix:preprocessing_rotation}).

\section{Tight gray-box certificates}\label{section:tight_certificates}
\looseness=-1
Now that we have established a baseline for gray-box certification, one may naturally wonder about its optimality.
We answer this by deriving tight gray-box certificates, i.e.~the best certificates that can be obtained for prediction $y^*$ using only the invariances of base classifier $g$ under group $\sT$ and its prediction probability $p_{\mX,y^*}$ under clean smoothing distribution $\mu_\mX$.
Similar to~\cref{section:background_randomized_smoothing}, we do so by finding a worst-case classifier.
In addition to constraining its  clean prediction probability, we constrain the classifier to be invariant under $\sT$, 
i.e.~solve
$\min_{h \in \sH_\sT} \Pr_{\mZ \sim \mu_{\mX'}} \left[h(\mZ) = y^*\right]$
with
\begin{equation}\label{eq:classifier_set_invariant}
\sH_\sT = \left\{h : \sR^{N \times D} \rightarrow \sY \mid 
  \Pr_{\mZ \sim \mu_\mX}
  \left[h(\mZ) = y^*\right]
   \geq
  p_{\mX,y^*}
  ,\forall \mZ, \forall \mZ' \in \left[\mZ\right]_\sT : h(\mZ) = h(\mZ')
\right\},
\end{equation}
where $\left[\mZ\right]_\sT$ is the orbit of $\mZ$ w.r.t.\ $\sT$ (see~\cref{definition:equivalence_class}).
In the following, we show how to solve the above optimization problem and then apply our certification methodology to specific invariances.

\subsection{Certification methodology}\label{section:certification_methodology}
\looseness=-1
To work with invariance constraints, it is convenient to not think of $h$ as a function, but a family of variables $(h_\mZ) \in \sY$ indexed by $\sR^{N \times D}$. The invariance constraint states that all variables from an orbit should have the same value, i.e.~$\forall \mZ, \forall \mZ' \in \left[\mZ\right]_\sT  : h_\mZ = h_{\mZ'}$.
Intuitively, the constraint can be enforced by replacing all these variables with a single variable.
We propose to formalize this idea by using canonical maps, which map all inputs from an orbit to a single, distinct representative:
\begin{restatable}[Canonical map]{definition}{canonicalmap}\label{definition:canonical_map}
A canonical map for invariance under a group of transformations $\sT$ is a function $\gamma:\sR^{N \times D} \rightarrow \sR^{N \times D}$ with
\begin{align}
    &\forall \mZ \in \sR^{N \times D}: \gamma(\mZ) \in \left[\mZ\right]_\sT,\label{eq:canonical_map_1}
    \\
    &\forall \mZ \in \sR^{N \times D}, \forall \mZ' \in \left[\mZ\right]_\sT : \gamma(\mZ) = \gamma(\mZ').\label{eq:canonical_map_2}
\end{align}
\end{restatable}
In~\cref{appendix:tight_equivalence_class_theorem}, we prove that canonical maps let us discard the invariance constraints:
\begin{restatable}{lemma}{canonicalcert}\label{theorem:canonical_cert}
    Let $g : \sR^{N \times D} \rightarrow \sY$ be invariant under  group $\sT$ and let $\sH_\sT$ be defined as in~\cref{eq:classifier_set_invariant}.
    If $\gamma:\sR^{N \times D} \rightarrow \sR^{N \times D}$ is a canonical map for invariance under $\sT$, then
    \begin{equation*}
        \min_{h \in \sH_\sT} \Pr_{\mZ \sim \mu_{\mX'}} \left[h(\mZ) = y^*\right]
        =
        \min_{h : \sR^{N \times D} \rightarrow \sY}   \Pr_{\mZ \sim \mu_{\mX'}} \left[h(\gamma(\mZ)) = y^*\right]
        \text{s.t.} \Pr_{\mZ \sim \mu_{\mX}} \left[h(\gamma(\mZ)) = y^*\right] \geq p_{\mX, y^*}.
    \end{equation*}
\end{restatable}
Like in~\cref{section:background_randomized_smoothing}, the 
optimization problem without invariance constraints from~\cref{theorem:canonical_cert} could now be solved exactly using the Neyman-Pearson lemma -- if we could find the  distribution of $\gamma(\mZ)$, i.e.\ the distribution of representatives\footnote{which, by~\cref{definition:canonical_map}, is equivalent to a distribution over orbits.}.
This 
can be achieved for groups $\mathit{T}(D)$, $\mathit{SO}(D)$ and  $\mathit{SE}(D)$ via careful change of variables into an alternative parameterization of $\sR^{N \times D}$ (see~\cref{appendix:tight_methodology}).

This leads us to our main result,
which we derive more formally in~\cref{appendix:tight_methodology}.
In the following, let $\langle \mA, \mB\rangle_\mathrm{F}$ be the Frobenius inner product $\sum_{n=1}^N (\mA_{n})^T \mB_{n}$ 
and recall from~\cref{section:haar_measure} that a right Haar measure is a measure for integration over a group $\sT$.
\begin{restatable}{theorem}{maincert}\label{theorem:main}
    Let $g : \sR^{N \times D} \rightarrow \sY$ be invariant under $\sT$ with $\sT$ chosen from $\{\mathit{T}(D), \mathit{SO}(D), \mathit{SE}(D)\}$.
    For $\mathit{SO}(D)$ and $\mathit{SE}(D)$, let $D \in \{2,3\}$.
    Let $\sH_\sT$ be defined as in~\cref{eq:classifier_set_invariant} 
    and $\eta$ be a right Haar measure on $\sT$. 
    Define the indicator function $h^* : \sR^{N \times D} \rightarrow \{0,1\}$ with 
    \begin{align}
        h^*(\mZ) = 
        \indicator \left[\frac{\beta_{\mX'}(\mZ)}{\beta_{\mX}(\mZ)} \leq \kappa \right]
        &\text{ , where }
        \beta_{\mX}(\mZ) = \int_{t \in \sT} \exp\left(
        \left\langle t \circ \mZ, \mX \right\rangle_\mathrm{F} \mathbin{/} \sigma^2
        \right) \ d \eta(t) 
        \label{eq:main_result_classifier}
        \\
        & \text{ and } \kappa \in \sR \text{ such that } 
        \E_{\mZ \sim \mu_{\mX}} \left[h^*(\mZ)\right] = p_{\mX, y^*}.
        \label{eq:main_result_constraint}
    \end{align}
    Then
    \begin{equation}
        \min_{h \in \sH_\sT} \Pr_{\mZ \sim \mu_{\mX'}} \left[h(\mZ) = y^*\right]
        =
        \E_{\mZ \sim \mu_{\mX'}} \left[h^*(\mZ)\right]
        \label{eq:main_result_objective}.
    \end{equation}
\end{restatable}
The indicator function $h^*$ in~\cref{eq:main_result_classifier} corresponds to the worst-case invariant classifier for clean prediction probability $p_{\mX,y^*}$.
To classify an input sample $\mZ$, it integrates Gaussian kernels of $(t \circ \mZ, \mX')$ and $(t \circ \mZ, \mX)$ over group $\sT$ (e.g.\ over all possible rotations). If the ratio of these integrals is below a threshold $\kappa$, it classifies $\mZ$ as $y^*$.
The constraint in~\cref{eq:main_result_constraint} ensures that the probability of predicting $y^*$ under the clean smoothing distribution $\mu_\mX$ matches that of the actual base classifier $g$.
The expected value with respect to perturbed smoothing distribution $\mu_{\mX'}$ in~\cref{eq:main_result_objective} is the optimal value of our optimization problem. As discussed in~\cref{section:background_randomized_smoothing}, the prediction is certifiably robust if this optimal value is greater than $\nicefrac{1}{2}$.

\textbf{Applying the certificate} to a group $\sT$ requires three steps:
1.) Calculating the Haar integrals in~\cref{eq:main_result_classifier}.
2.) Solving~\cref{eq:main_result_constraint} for threshold $\kappa$.
3.) Evaluating the expected value in~\cref{eq:main_result_objective}.

\looseness=-1
\textbf{Connection to prior work.}
This result differs from black-box randomized smoothing with Gaussian noise, where the worst-case classifier is a linear model~\cite{Cohen2019}.
The \textit{group averaging} performed by the worst-case invariant classifier, i.e.\ integrating a function over group $\sT$, is a key technique for building invariant models~\cite{Cohen2016,Bruna2013,Murphy2018,Murphy2019,Puny2021,Yarotsky2022}.
Group-averaged kernels have been proposed in~\cite{Haasdonk2005}.
It is fascinating to see them naturally materialize from nothing but an invariance constraint.
\subsection{Translation invariance}\label{section:tight_translation}
\looseness=-1
In the case of translation invariance (i.e.\ $\sT = \mathit{T}(D)$), we can evaluate the worst-case classifier (\cref{eq:main_result_classifier}, solve for threshold $\kappa$ (\cref{eq:main_result_constraint}) and evaluate the perturbed prediction probability (\cref{eq:main_result_objective}) analytically.
This leads to the following result (proof in~\cref{appendix:tight_translation}):
\begin{restatable}{theorem}{translationcert}\label{theorem:translation_cert}
    \looseness=-1
    Let $g : \sR^{N \times D} \rightarrow \sY$ be invariant under $\sT = \mathit{T}(D)$ and $\sH_\sT$ be defined as in~\cref{eq:classifier_set_invariant}.
    Then
    \begin{equation*}
        \min_{h \in \sH_\sT} \Pr_{\mZ \sim \mu_{\mX'}} \left[h(\mZ) = y^*\right]
        =
        \Phi\left(\Phi^{-1} \left( p_{\mX,y^*}  \right) -
        \frac{1}{\sigma}
        \left\lvert\left\lvert \mDelta - \ones_N \overline{\mDelta}\right\rvert\right\rvert_2 
        \right),
    \end{equation*}
    where $\overline{\mDelta} \in \sR^{1 \times D}$ are the column-wise averages of $\Delta = \mX' - \mX$ and $\sigma$ is the  standard deviation of the isotropic matrix normal smoothing distribution $\mu_\mX$.
\end{restatable}
\textbf{Certificate parameters.}
For a fixed smoothing standard deviation $\sigma$, 
the certificate depends on a single parameter: The norm $\left\lvert\left\lvert \mDelta - \ones_N \overline{\mDelta}\right\rvert\right\rvert_2$ of the mean-centered perturbation matrix $\mDelta$.

\textbf{Comparison to the orbit-based certificate.} 
Substituting into
robustness condition $\smash{\min_{h \in \sH_\sT} \Pr_{\mZ \sim \mu_{\mX'}} \left[h(\mZ) = y^*\right] > \frac{1}{2}}$
shows that $f(\mX') = y^*$
if  $\left\lvert\left\lvert \mDelta - \ones_N \overline{\mDelta}\right\rvert\right\rvert_2 < \sigma \Phi^{-1}\left(p_{\mX,y^*}\right)$.
This result, obtained via our tight certification methodology, is identical to the orbit-based certificate for translation invariance from the end of~\cref{section:postprocessing}.
In other words: Despite its simplicity, the orbit-based certificate is the best possible gray-box certificate
for translation invariance.

\subsection{Rotation invariance in 2D}\label{section:rotation_2d_tight}
\looseness=-1
Considering the previous result, one may  suspect the orbit-based certificate to be tight for arbitrary invariances. This is  not the case. For instance, it is not tight for rotation invariance ($\sT = \mathit{SO}(D)$):
\begin{restatable}{theorem}{rotationnontightness}\label{theorem:rotation_non_tightness}
    Let $g : \sR^{N \times D} \rightarrow \sY$ be invariant under $\sT = \mathit{SO}(D)$ and $\sH_\sT$ be defined as in~\cref{eq:classifier_set_invariant}.
    Assume that perturbed input $\mX'$ is not obtained via rotation of $\mX$, i.e.\ 
    $\nexists \mR \in \mathit{SO}(D) : \mX' = \mX \mR^T$. Further assume that $p_{\mX,y^*} \in (0,1)$. Then, for all $\mR \in \mathit{SO}(D)$:
    \begin{equation}
        \min_{h \in \sH_\sT} \Pr_{\mZ \sim \mu_{\mX'}} \left[h(\mZ) = y^*\right]
        >
        \Phi\left(\Phi^{-1}\left(p_{\mX,y^*}\right)
        - \frac{1}{\sigma}\left\lvert\left\lvert \mX'\mR^T - \mX \right\rvert\right\rvert_2
        \right).
    \end{equation}
\end{restatable}
In other words: The tight certificate is strictly stronger. Proof in~\cref{appendix:rotation_non_tightness}.

Next, we apply~\cref{theorem:main} to obtain the strictly stronger, tight certificate for rotation invariance in 2D. 
In the following, let $\mR(\theta) \in \mathit{SO}(2)$
be 
the matrix that rotates counter-clockwise by angle $\theta$
and 
 $\mathcal{I}_0$ be the modified Bessel function of the first kind and order $0$. 
The Haar integral from~\cref{eq:main_result_classifier} that defines the worst-case classifier can be evaluated analytically (see~\cref{appendix:tight_rotation_2d})):
\begin{equation}\label{eq:worst_case_2d_closed}
\beta_\mX(\mZ) = \mathcal{I}_0
    \left(
    \frac{1}{\sigma^2}
    \sqrt{
        \left\langle \mZ, \mX \right\rangle_\mathrm{F}^2
        +
        \left\langle \mZ, \mX \mR(-\nicefrac{\pi}{2})^T \right\rangle_\mathrm{F}^2
    }
    \right),
\end{equation}
We can substitute this into~\cref{theorem:main} and use the fact that~\cref{eq:worst_case_2d_closed} depends on linear transformations of the matrix normal random variable $\mZ$ to obtain the following certificate (proof in~\cref{appendix:tight_rotation_2d}):
\begin{restatable}{theorem}{2drotationcert}\label{theorem:2d_rotation_cert}
    Let $g : \sR^{N \times 2} \rightarrow \sY$ be invariant under $\sT = \mathit{SO}(2)$ and $\sH_\sT$ be defined as in~\cref{eq:classifier_set_invariant}.
    Define the indicator function $\rho : \sR^4 \rightarrow \{0,1\}$ with
    \begin{align}
        \rho(\vq) & =
        \indicator \left[\mathcal{I}_0\left(\sqrt{\evq_1^2 + \evq_2^2}\right) \mathbin{/} 
            \mathcal{I}_0\left(\sqrt{\evq_3^2 + \evq_4^2}\right) \leq \kappa\right],
        \label{eq:2d_rotation_classifier}
       \\
        & \text{ with } \kappa \in \sR \text{ such that } 
        \E_{\vq \sim \mathcal{N}\left(
            \vm^{(2)},
            \mSigma
        \right)} \left[\rho(\vq)\right] = p_{\mX, y^*}.
        \label{eq:2d_rotation_threshold}
    \end{align}
    Then
    \begin{equation}
        \min_{h \in \sH_\sT} \Pr_{\mZ \sim \mu_{\mX'}} \left[h(\mZ) = y^*\right]
        =
        \E_{\vq \sim \mathcal{N}\left(
            \vm^{(1)},
            \mSigma
        \right)} \left[\rho(\vq) \right]
        \label{eq:2d_rotation_objective}.
    \end{equation}
    where 
    $\vm^{(1)}, \vm^{(2)} \in \sR^{4}$, $\mSigma \in \sR^{4\times4}$ are linear combinations (see~\cref{appendix:tight_rotation_2d})  of  $||\mX||_2^2 \mathbin{/} \sigma^2$, $||\mDelta||_2^2 \mathbin{/} \sigma^2$ and parameters
    $\epsilon_1 = \left\langle \mX, \mDelta \right\rangle_\mathrm{F}$, 
    $\epsilon_2 = \langle \mX \mR\left(-\nicefrac{\pi}{2}\right)^T, \mDelta \rangle_\mathrm{F}$.
\end{restatable}
\looseness=-1
\textbf{Certificate parameters.}
This certificate depends on the perturbation norm $\smash{||\mDelta||_2}$ and 
clean data norm $\smash{||\mX||_2}$, relative to the smoothing standard deviation $\sigma$. It further depends on parameters
$\epsilon_1$  and 
$\epsilon_2$, which are 
Frobenius inner products between $\mDelta$ and the clean input $\mX$ before and after a rotation by $-\nicefrac{\pi}{2}$. These parameters capture the orientation of $\mX' = \mX + \mDelta$ relative to $\mX$, as one would expect from a rotation-invariance aware certificate.

\looseness=-1
\textbf{Monte Carlo evaluation.}
Evidently, we do not have a closed-form
expression for the expectations in~\cref{eq:2d_rotation_objective,eq:2d_rotation_threshold}.
However, recall from~\cref{section:background_randomized_smoothing} that randomized smoothing already involves an intractable expectation, namely the prediction probability $p_{\mX,y^*}$.
One has to use
 Monte Carlo 
sampling to compute a lower confidence bound that hold with high probability $1- \alpha$.
We
adopt the same approach: First, we lower-bound threshold $\kappa$ from~\cref{eq:2d_rotation_threshold}, i.e.\ make the classifier slightly less likely to predict $y^*$.
Then, we lower-bound bound expectation 
$\E_{\vq \sim \mathcal{N}\left(
            \vm^{(1)},
            \mSigma
        \right)} \left[\rho(\vq) \right]$ from~\cref{eq:2d_rotation_objective}.
Because the resulting value is slightly smaller than than optimal value of our optimization problem $\min_{h \in \sH_\sT} \Pr_{\mZ \sim \mu_{\mX'}} \left[h(\mZ) = y^*\right]$, this procedure yields a valid certificate.
We discuss the full algorithm and how to ensure that all bounds simultaneously hold in~\cref{appendix:monte_carlo_certification}.
Because the expectations only require sampling from four-dimensional normal distributions and do not
depend on base classifier $g$, one can use a large number of samples to obtain narrow bounds at little computational cost
(e.g.~$0.59\si{\second}$ for $100000$ samples
per confidence bound
on an Intel Xeon E5-2630 CPU).

\looseness=-1
\subsection{Rotation invariance in 3D}\label{section:tight_rotation_3d}
To evaluate the tight certificate for 3D rotation invariance (i.e.\ $\sT = \mathit{SO}(3)$), we adopt the same Monte Carlo evaluation approach discussed in~\cref{section:rotation_2d_tight}.
Different from 2D rotation invariance, evaluating the worst-case classifier analytically is not tractable.
It can however be evaluated by numerical integration over an Euler angle parameterization of rotation group $\mathit{SO}(3)$ (see~\cref{appendix:tight_rotation_3d}).

\subsection{Roto-translation invariance in 2D and 3D}\label{section:tight_translation_rotation}
In~\cref{appendix:tight_roto_translation} we prove that additionally enforcing translation invariance (i.e.\ $\sT = \mathit{SE}(D)$)  is equivalent to centering $\mX$ and $\mDelta$ before evaluating the certificates for rotation invariance. This is consistent with our result from~\cref{section:tight_translation}, i.e.\ the orbit-based approach being optimal for translation.

\section{Limitations and broader impact}\label{limitations}
\looseness=-1
The main limitation of our work lies in its exploratory nature.
We have derived the first certificates for models invariant under arbitrary Euclidean isometries and/or permutation.
While these invariances are of key importance to many practical applications
, there is a vast swath of invariances that we have not covered, such as spatio-temporal invariances~\cite{Bogatskiy2020}, invariance under graph isomorphisms~\cite{Kipf2017} or invariances for planar images~\cite{Cohen2016}.
Furthermore, we have not yet derived tight certificates for reflections and permutations (though our certification methodology could conceivably be applied to them).

\textbf{Broader impact.}
With the growing prevalence of machine learning in safety-critical and sensitive domains like autonomous driving~\cite{Grigorescu2020,Muhammad2020} or healthcare~\cite{Miotto2018,Norgeot2019}, trustworthy models promise to become increasingly important.
Certification is one pillar of
trustworthiness, alongside
concepts like
fairness~\cite{Corbett2018,Mehrabi2021} or differential privacy~\cite{Dwork2008,Ji2014}.
Unique to our work is that we certify
models with spatial invariances. These invariances naturally occur in the physical sciences, including those with a direct societal impact like pharmacology and biochemistry.
Our work can be seen as a first step towards provable trustworthiness for tasks like drug discovery~\cite{Xiong2019,Ramsauer2020,Cai2020} or protein folding~\cite{Qian1988,
Fariselli2001,
Xu2019,
AlQuraishi2019,
Ruff2021}.

\section{Experimental evaluation}\label{section:main_experiments}
\looseness=-1
We already know that the orbit-based certificate for translation invariance is
tight and
certifies robustness for
an infinitely larger volume
than black-box randomized smoothing (see~\cref{fig:explainy_figure}).
Therefore, we focus our
experiments 
on the certificates for rotation invariance.
Recall that the orbit-based certificate
guarantees robustness for the set 
$\tilde{\sB} = \left\{\mX' \mid || \mX' \mR^T - \mX ||_2 < r \right\}$ with $\smash{r = \sigma \Phi^{-1}\left(p_{\mX,y^*}\right)}$ and optimal rotation matrix $\mR$.
In other words: It certifies robustness for perturbations with rotational components that can be eliminated to bring $\mX'$ into distance $r$ of $\mX$.
We want to understand whether the tight certificates offer any benefit beyond that, or if the strict inequality in~\cref{theorem:rotation_non_tightness} is due to some negligible $\epsilon$.
To this end, we first thoroughly examine the four-dimensional parameter space of the tight certificate for rotation invariance in 2D, 
before applying our certificates to rotation invariant point cloud classifiers.

\looseness=-1
All parameters and experimental details are specified in~\cref{appendix:full_experimental_setup}. 
We use $10000$ samples per confidence bound and set $\alpha=0.001$, i.e.~all certificates hold with $99.9\%$ probability.
A reference implementation will be made available at \href{https://www.cs.cit.tum.de/daml/invariance-smoothing/}{https://www.cs.cit.tum.de/daml/invariance-smoothing}.

\subsection{Tight certificate parameter space}\label{section:main_experiments_inverse}
\looseness=-1
The tight certificate for 2D rotation invariance depends on $||\mX||_2 \mathbin{/} \sigma$, $||\mDelta||_2 \mathbin{/} \sigma$ and parameters $\epsilon_1$ and $\epsilon_2$, which capture the orientation of the perturbed point cloud and fulfill 
$\sqrt{\epsilon_1^2 + \epsilon_2^2} \leq ||\mX||_2 \cdot ||\mDelta||_2$
(see~\cref{appendix:tight_certificate_parameters}).
To avoid clutter, we define $\smash{\tilde{\epsilon}_{k} \vcentcolon= \epsilon_{k} \mathbin{/} \left(||\mX||_2 \cdot ||\mDelta||_2\right)}$.
As our metric for this section, we report $p_\mathrm{min}$, the smallest probability $p_{\mX,y^*}$ for which a prediction can still be certified
\footnote{We discuss how to compute these inverse certificates in~\cref{appendix:inverse_certificates}.}
\begin{figure}
\begin{minipage}[t]{0.49\textwidth}
    \centering
     \input{figures/experiments/inverse/parallel/inverse_parallel_0-5.pgf}
\caption{Comparison of tight and orbit-based certificates applied to adversarial scaling for $\sigma=0.5$ and varying $||\mDelta||$ and $||\mX||$ (smaller $p_\mathrm{min}$ is better). As $||\mX||$ increases, the difference between the certificates shrinks.}
     \label{fig:main_inverse_parallel}
\end{minipage}
\hfill
\begin{minipage}[t]{0.49\textwidth}
\centering
    \includegraphics{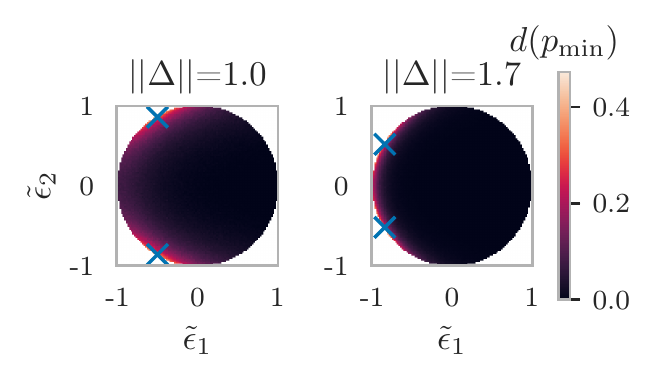}
\caption{Difference in $p_\mathrm{min}$ between the tight certificate for 2D rotation invariance and the black-box baseline for  $\sigma=0.5$, $||\mX||=1.0$. Crosses correspond to adversarial rotations.
Large $d(p_\mathrm{min})$ can be observed near adversarial rotations.
}
\label{fig:main_inverse_multiple_inner_cross_delta}
\end{minipage}
\vspace{-0.55cm}
\end{figure}

\looseness=-1
\textbf{Adversarial scaling.} First, we assume that $\smash{\mX' = (1 + c) \mX}$, i.e.~the input is adversarially scaled.
In this case, we have $\tilde{\epsilon}_1 = 1$ and $\tilde{\epsilon}_2 = 0$.
We then vary $||\mDelta||_2$ and $||\mX||_2$ and evaluate our certificates.
Note that such attacks have no rotational component, i.e.\ the orbit-based certificate is identical to the black-box one.
\cref{fig:main_inverse_parallel} shows that, even in the absence of rotations, the tight certificate can yield significantly stronger guarantees.
For $\sigma=0.5$ and $||\mX||_2=0.01$, the baseline  can only certify robustness for a  prediction with $\smash{p_{\mX,y^*} = 0.8}$ if $||\mDelta||_2\leq 0.4$. The tight certificate can certify robustness up to $||\mDelta||_2 = 0.73$.
However, the gap shrinks, as the norm of the clean data increases.

\begin{figure}
    \centering
    \includegraphics[scale=0.95]{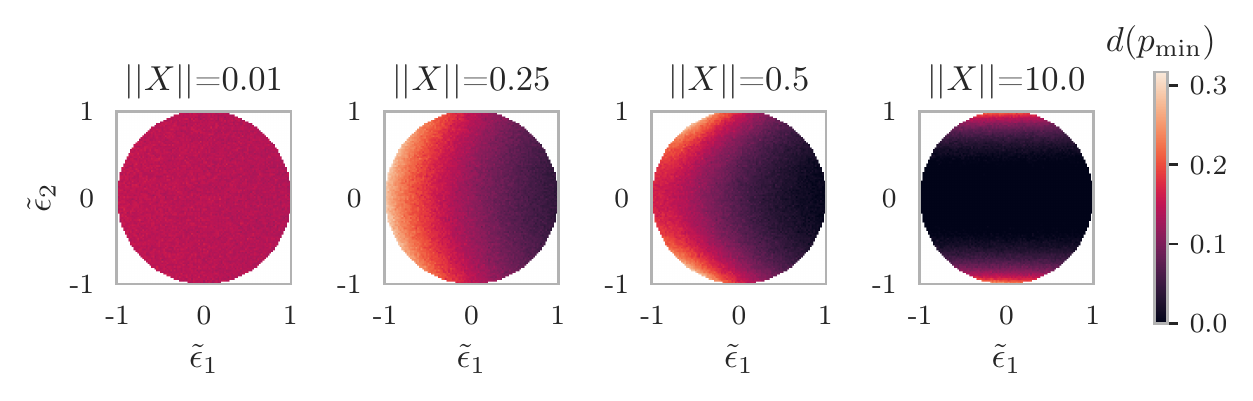}
    \vspace{-0.2cm}
    \caption{Difference in $p_\mathrm{min}$ between the tight certificate for 2D rotation invariance and the black-box baseline for $\sigma=0.5$, $||\mDelta||=0.5$ under varying $||\mX||$, $\tilde{\epsilon}_1$ and $\tilde{\epsilon}_2$.
    As $||\mX||$ increases, the regions where the tight certificate outperforms the black-box baseline shrink.}
    \label{fig:main_inverse_multiple_inner_cross_clean}
\end{figure}

\looseness=-1
\textbf{Effect of data norm.} We would like to see if this is a pervasive pattern. To this end, we fix $||\mDelta||_2$, gradually increase $||\mX||_2$, and evaluate the tight certificate for $\tilde{\epsilon}_1$, $\tilde{\epsilon}_2$ on  a $100 \times 100$ rasterization of $\smash{[0,1] \times [0,1]}$.
We then measure the difference $d(p_\mathrm{min})$ to the black-box certificate (we will discuss how these results relate to the orbit-based one shortly).
\cref{fig:main_inverse_multiple_inner_cross_clean} shows that for small $||\mX||_2$, the tight certificate outperforms the black-box certificate for arbitrary perturbations of norm $||\mDelta||_2$. But, as $||\mX||_2$ increases, the regions where it outperforms the black-box one shrink.

\begin{figure}
\begin{minipage}{0.49\textwidth}
\centering
    \input{figures/experiments/smoothed_accuracy/smoothed_accuracy.pgf}
\caption{Test set accuracy of smoothed  translation and rotation invariant point cloud classifiers on the ModelNet40 and MNIST pointcloud datasets, under varying standard deviation $\sigma$.}
\label{fig:main_smoothed_accuracy}
\end{minipage}
\hfill
\begin{minipage}{0.49\textwidth}
\centering
    \input{figures/experiments/forward/mnist_parallel/0.15.pgf}
\caption{Comparison of certificates for adversarial scaling of MNIST with EnsPointNet and $\sigma=0.15$. The tight and orbit-based certificates yield similar certified accuracies.
}
\label{fig:main_forward_mnist_parallel}
\end{minipage}
\vspace{-0.55cm}
\end{figure}

\looseness=-1
\textbf{Rotational components.}
Simple algebra (see~\cref{appendix:tight_certificate_parameters}) shows that for any combination of $||\mX||_2$ and $||\mDelta||_2$, 
adversarial rotations, i.e.\ $\mX' = \mX \mR^T$ such that $||\mX' - \mX|| = ||\mDelta||$, 
correspond to two specific points with $\sqrt{\tilde{\epsilon}_1^2 + \tilde{\epsilon}_2^2} = 1$ in the certificate's parameter space.
In~\cref{fig:main_inverse_multiple_inner_cross_delta} we fix $\sigma=0.5$ and $||\mX||_2 = 1$, vary $||\mDelta||$ and highlight these two points.
We observe that the tight certificate only outperforms the black-box certificate for values of $\tilde{\epsilon_1}, \tilde{\epsilon_2}$ that are close to adversarial rotations.
However, robustness to such attacks can also be readily certified by the orbit-based approach.
Combined with our previous observations, and because the certificate only depends on norms relative to smoothing standard deviation $\sigma$, i.e. $||\mX||_2 \mathbin{/} \sigma$ and $||\mDelta||_2 \mathbin{/} \sigma$, we expect the tight and the orbit-based certificate to perform similarly well, assuming that the smoothing standard deviation $\sigma$ is small  relative to  $||\mX||_2$.

\looseness=-1
\subsection{Application to point cloud classification}\label{section:main_experiments_forward}
To verify whether our observations hold in practice, we apply our certificates to point cloud classification. We consider two datasets: 3D point cloud representations of ModelNet40~\cite{Wu2015}, which consists of CAD models from 40 different categories, and 2D point cloud representations of MNIST~\cite{Lecun1998}. We apply the same pre-processing steps as in~\cite{Qi2017}.
Certification is performed on the default test sets.
As our base classifiers, we use rotation and translation (i.e.\ $\mathit{SE}(D)$) invariant versions of two well-established models
that are used in prior work on robustness certification for point clouds: PointNet~\cite{Qi2017} and DGCNN~\cite{Wang2019}.
To implement the invariances, we center the input data, perform principal component analysis, apply the model to all possible poses (see discussion in~\cite{Xiao2020,Li2021attn}) and average the output logits (EnsPointNet and EnsDGCNN).
In addition, we consider a more refined model~\cite{Xiao2020} that combines canonical poses via a self-attention mechanism (AttnPointNet).
Throughout this section, we report both probabilistic upper and lower bounds for the Monte Carlo evaluation of the tight certificates (recall~\cref{section:rotation_2d_tight}).

\looseness=-1
\textbf{Practical smoothing parameters.} \cref{fig:main_smoothed_accuracy} shows the test set accuracies of randomly smoothed models under varying standard deviations $\sigma$. Values of $\sigma$ that preserve an accuracy above $50\%$ are small, relative to the average norm of the test sets ($10.67$ for MNIST, $19.17$ for ModelNet40).
Going by our previous results, we expect the tight and orbit-based certificates to perform similarly well.

\looseness=-1
\textbf{Adversarial scaling.} In~\cref{fig:main_forward_mnist_parallel} we again consider adversarial scaling, i.e.~attacks without rotational components, but applied to the MNIST point cloud dataset with $\sigma=0.15$. We report the certified accuracy, i.e.~the percentage of correct and provably robust predictions, for certification with ($\mathit{SE}(2)$) and without ($\mathit{SO}(2)$) translation invariance. The tight and orbit-based certificate yield similar results. 
We further observe that enforcing translation invariance increases the certified accuracy and extends the range of $||\Delta||$ for which robustness can be certified.

\begin{figure}
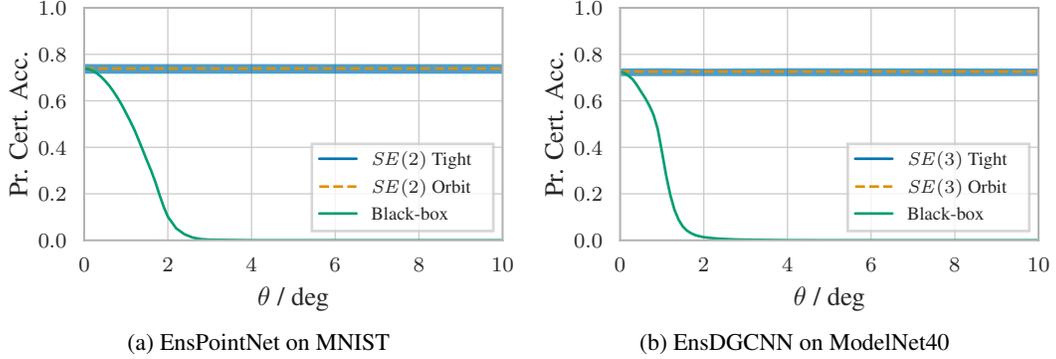

     \centering
     \begin{subfigure}[b]{0.49\textwidth}
         \centering
    \input{figures/experiments/forward/rotation/mnist/pointnet/0.1/0.1_main.pgf}
         \caption{EnsPointNet on MNIST}
     \end{subfigure}
     \hfill
     \begin{subfigure}[b]{0.49\textwidth}
         \centering
    \input{figures/experiments/forward/rotation/modelnet/dgcnn/arbitrary/0.1/0.1_main.pgf}
         \caption{EnsDGCNN on ModelNet40}
     \end{subfigure}
     \caption{Comparison of tight, orbit-based and black-box certificates for randomly perturbed inputs with $||\mDelta|| = \sigma = 0.1$, rotated by angle $\theta$. The gray-box certificates effectively eliminate the induced rotation, while the black-box method cannot certify robustness for large $\theta$.}
     \label{fig:main_forward_multiple_inner_cross}
     \vspace{-0.55cm}
\end{figure}

\looseness=-1
\textbf{Rotational components.} Finally, we study perturbations with rotational components.
We fix $||\mDelta||$, randomly sample perturbations of the specified norm 
and then rotate $\smash{\mZ = \mX + \mDelta}$ by a specified angle $\smash{\theta \in [0,10^{\circ}]}$ (in the case of ModelNet40, around one randomly chosen axis).  
For each element of the test set and each $\theta$, we generate $10$ such samples.
We then compute the percentage of samples $\mX'$ for which $f(\mX)$ is correct and $f(\mX') = f(\mX)$ is provably guaranteed (``probabilistic certified accuracy'').
\ref{fig:main_forward_multiple_inner_cross} shows results for MNIST and ModelNet40 evaluated with $||\mDelta|| = \sigma = 0.1$.
The black-box baseline's probabilistic certified accuracy drops close to $0$ for $\theta=2^{\circ}$. The gray-box certificates are almost constant in $\theta$, i.e.~effectively eliminate any induced rotation.
However, the tight certificate did not offer any meaningful benefit beyond that.
Using the lower bound for Monte Carlo evaluation, there was not a single sample for which only the tight certificate could guarantee robustness.

\looseness=-1
In ~\cref{appendix:experiments} we repeat the experiments from this and the previous section for various other combinations of parameter values. All results are consistent with the ones presented here, confirming that the orbit-based approach offers a good approximation of the tight certificates in practice.

\section{Conclusion}
\looseness=-1
For the first time, we have studied the use of invariances for robustness certification.
We proposed a gray-box approach, combining white-box knowledge about invariances with black-box randomized smoothing.
We have derived a orbit-based procedure for certification that can be applied to arbitrary models with invariance to permutations and Euclidean isometries.
We have proven that the orbit-based certificate for translation invariance is tight and derived strictly stronger certificates for rotation invariance.
Our experiments are to be interpreted in two ways:
Firstly, the fact that it is possible to derive tight invariance-aware certificates and that there exist scenarios in which they offer stronger guarantees for arbitrary perturbations should be an exciting inspiration for future work.
Secondly,  the fact that the orbit-based certificates are easily interpretable and offer good approximations of our tight certificates should invite their application to real-world tasks with inherent invariances.

\textbf{Acknowledgements.}
The authors would like to thank Johannes Gasteiger for valuable discussions on invariant deep learning and Lukas Gosch for constructive criticism of the manuscript.
This research was supported by the German Research Foundation, grant GU 1409/4-1.

\bibliographystyle{unsrtnat} 
\bibliography{references/references} 



\appendix

\startcontents[section]
\printcontents[section]{l}{0}{\setcounter{tocdepth}{2}}

\clearpage

\section{Additional experiments}\label{appendix:experiments}

In the following, we repeat the experiments from~\cref{section:main_experiments} for a wider range of certificate and model parameters.
In~\cref{appendix:experiments_inverse}, we investigate the parameter space of the tight certificate for rotation invariance in 2D.
In~\cref{appendix:experiments_forward}, we again apply our certificates to rotation invariant point cloud classifiers,
this time considering all three models (EnsPointNet, AttnPointNet and EnsDGCNN) and different smoothing distribution parameters.
As before, we take $10000$ samples per confidence bound and set $\alpha=0.001$, i.e.~all certificates hold with $99.9\%$ probability.

All results support our main conclusion from~\cref{section:main_experiments}:
The tight certificates for rotation invariance can offer much stronger robustness guarantees than their orbit-based counterparts.
However, in practical scenarios, where the smoothing standard deviation is small relative to the norm of the clean data,
the orbit-based approach offers a very good approximation.

\clearpage

\subsection{Tight certificate parameter space}\label{appendix:experiments_inverse}
For this section, recall that the tight certificate for 2D rotation invariance depends on $||\mX||_2$, $||\mDelta||_2$ and parameters $\epsilon_1 = \left\langle \mX, \mDelta \right\rangle_\mathrm{F}$, 
    $\epsilon_2 = \left\langle \mX \mR\left(\nicefrac{-\pi}{2}\right)^T, \mDelta \right\rangle_\mathrm{F}$,
    as well as smoothing standard deviation $\sigma$.
We define $\tilde{\epsilon}_1 = \frac{\epsilon_1}{||\mX||\cdot ||\mDelta||}$ and
$\tilde{\epsilon}_2 = \frac{\epsilon_2}{||\mX||\cdot ||\mDelta||}$.
Note that $\sqrt{\tilde{\epsilon}_1^2 + \tilde{\epsilon}_2^2} \leq 1$ (see also~\cref{appendix:tight_certificate_parameters}).

\subsubsection{Adversarial scaling}
We begin with adversarial scaling, which corresponds to $\tilde{\epsilon}_1 = 1, \tilde{\epsilon}_2 = 0$. We consider $\sigma \in \{0.1,0.15,0.2, 0.25, 0.5, 1.0\}$. For each $\sigma$, we vary $||\mX||_2$ and $||\mDelta||$
and compute $p_\mathrm{min}$, the smallest value of $p_{\mX,y^*}$ for which a prediction can still be certified (see~\cref{appendix:inverse_certificates}).
Lower $p_\mathrm{min}$ mean that the certificate can guarantee robustness  for models that are less consistent in their predictions.

\begin{figure}[H]
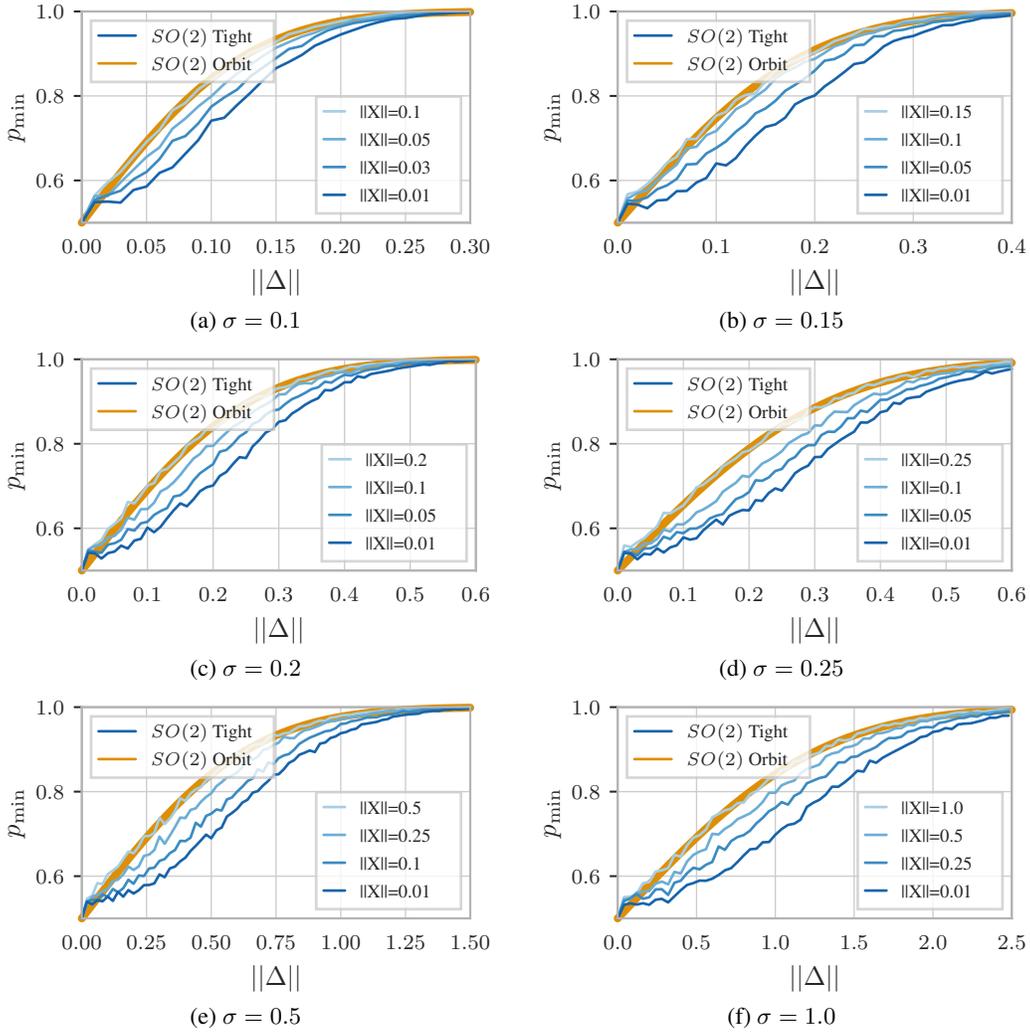

     \centering
     
     \begin{subfigure}[b]{0.49\textwidth}
         \centering
         \input{figures/experiments/inverse/parallel/inverse_parallel_0-1.pgf}
         \vskip-0.2cm
         \caption{$\sigma=0.1$}
     \end{subfigure}
     \hfill
     \begin{subfigure}[b]{0.49\textwidth}
         \centering
         \input{figures/experiments/inverse/parallel/inverse_parallel_0-15.pgf}
         \vskip-0.2cm
         \caption{$\sigma=0.15$}
     \end{subfigure}
     \begin{subfigure}[b]{0.49\textwidth}
         \centering
         \input{figures/experiments/inverse/parallel/inverse_parallel_0-2.pgf}
         \vskip-0.2cm
         \caption{$\sigma=0.2$}
     \end{subfigure}
     \hfill
     \begin{subfigure}[b]{0.49\textwidth}
         \centering
         \input{figures/experiments/inverse/parallel/inverse_parallel_0-25.pgf}
         \vskip-0.2cm
         \caption{$\sigma=0.25$}
     \end{subfigure}
     \begin{subfigure}[b]{0.49\textwidth}
         \centering
         \input{figures/experiments/inverse/parallel/inverse_parallel_0-5.pgf}
         \vskip-0.2cm
         \caption{$\sigma=0.5$}
     \end{subfigure}
     \hfill
     \begin{subfigure}[b]{0.49\textwidth}
         \centering
         \input{figures/experiments/inverse/parallel/inverse_parallel_1.pgf}
         \vskip-0.2cm
         \caption{$\sigma=1.0$}
     \end{subfigure}
     \caption{Comparison of tight and orbit-based certificates for rotation invariance in 2D,  applied to adversarial scaling for varying $||\mDelta||$, $||\mX||$ and smoothing standard deviation $\sigma$ (smaller $p_\mathrm{min}$ is better). As $||\mX||$ increases, the difference between the certificates shrinks.}
     \label{fig:appendix_inverse_parallel}
\end{figure}

\cref{fig:appendix_inverse_parallel} shows that, if the clean data norm is small, e.g.\ $||\mX|| = 0.01$, then the tight certificate yields much lower $p_\mathrm{min}$ than the orbit-based one, save for very small and very large values of $||\mDelta||$.
That is, the tight certificate for rotation invariance is beneficial even for adversarial perturbations that do not have  any rotational components.
However, as $||\mX||$ approaches $\sigma$, this difference shrinks, i.e.\ both approaches offer guarantees of similar strength.

\clearpage

\subsubsection{Effect of clean data norm on certificates for arbitrary perturbations}
Next, we study whether the same effect can be observed with arbitrary perturbations, i.e.\ arbitrary values of $\tilde{\epsilon}_1$ and $\tilde{\epsilon}_2$.
In~\cref{fig:main_inverse_multiple_inner_cross_clean} from~\cref{section:main_experiments_inverse}, we fixed a specific combination of $\sigma$ and $||\mDelta||$, varied $||\mX||_2$ and evaluated the tight certificate for $\tilde{\epsilon}_1, \tilde{\epsilon}_2$ taken from a $100 \times 100$ rasterization of $[0,1]\times [0,1]$.
We then measured the difference $d(p_\mathrm{min})$ to the black-box certificate.

Here, we consider different combinations of $\sigma$, $||\mDelta||$ and $||\mX||$.
We can reduce the dimensionality of the parameter space that needs to be explored by observing that both the tight gray-box and the black-box certificate do not depend on the absolute value of these parameters, but their value relative to $\sigma$.
The black-box baseline has $\smash{p_\mathrm{min} = \Phi(||\mDelta||_2 \mathbin{/} \sigma})$.
As specified in~\cref{appendix:tight_rotation_2d}, the tight certificate depends on $||\mX||_2^2 \mathbin{/} \sigma^2$, $||\mDelta||_2^2 \mathbin{/} \sigma^2$,
 $\epsilon_1 \mathbin{/} \sigma^2 = \frac{1}{\sigma^2} ||\mX||_2 \cdot ||\mDelta||_2 \cdot \cos(\theta_1)$ and $\frac{1}{\sigma^2}   \mathbin{/} ||\mX||_2 \cdot ||\mDelta||_2 \cdot \cos(\theta_1)$, where $\theta_1, \theta_2$ are angles between $2N$-dimensional vectors.
Therefore, we can fix an arbitrary value of $\sigma$ and then choose $||\mX||_2$ and $||\mDelta||_2$ relative to it.

We set set $\sigma = 1$ and consider $||\mDelta||_2 \in \{\frac{1}{4} \sigma,  \frac{1}{2} \sigma, \sigma, 2 \sigma, 3 \sigma\}$,
i.e.\ perturbation norms that are much smaller, similar to or much larger than the smoothing standard deviation.
For each $||\mDelta||_2$, we evaluate the tight certificate $||\mX||_2 \in \{\frac{1}{100} \sigma, \frac{1}{2} \sigma, \sigma, 10 \sigma\}$, i.e.\ clean data norms that are much smaller, similar to or much larger than the smoothing standard deviation.

\begin{figure}[H]
     \centering
     \begin{subfigure}[b]{0.99\textwidth}
         \centering
         \includegraphics{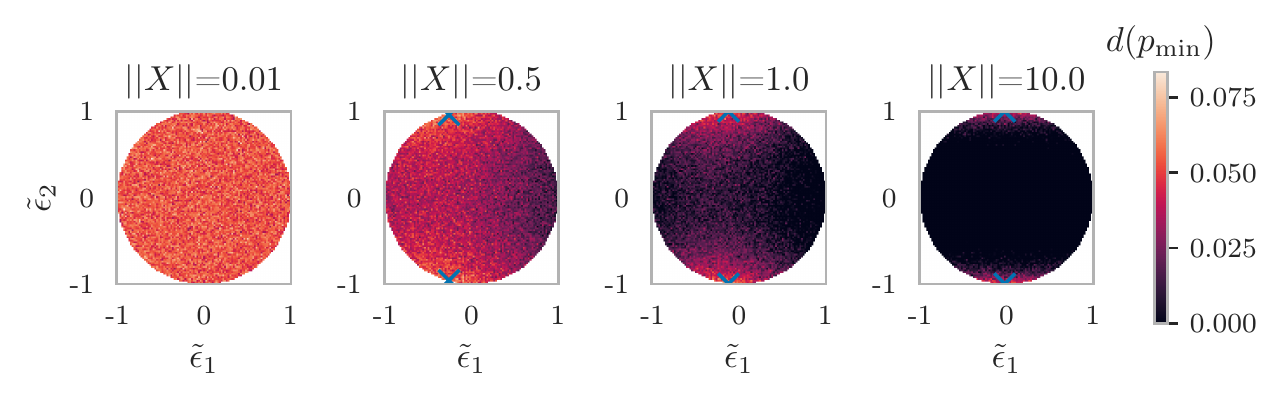}
         \vskip-0.25cm
         \caption{$||\mDelta||=\sigma \mathbin{/} 4$}
     \end{subfigure}
     \vskip-0.1cm
     \begin{subfigure}[b]{0.99\textwidth}
         \centering
         \includegraphics{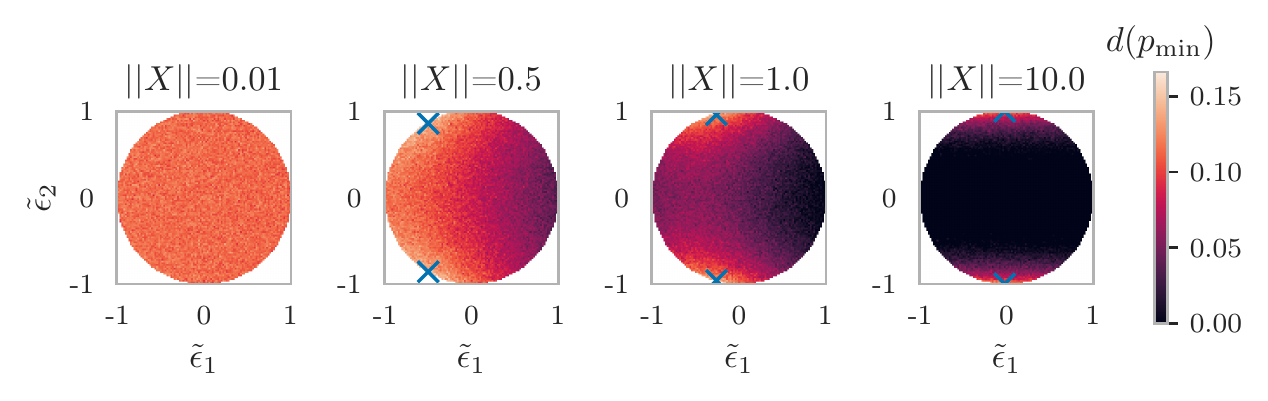}
         \vskip-0.25cm
         \caption{$||\mDelta||=\sigma \mathbin{/} 2$}
     \end{subfigure}
     \vskip-0.1cm
     \begin{subfigure}[b]{0.99\textwidth}
         \centering
         \includegraphics{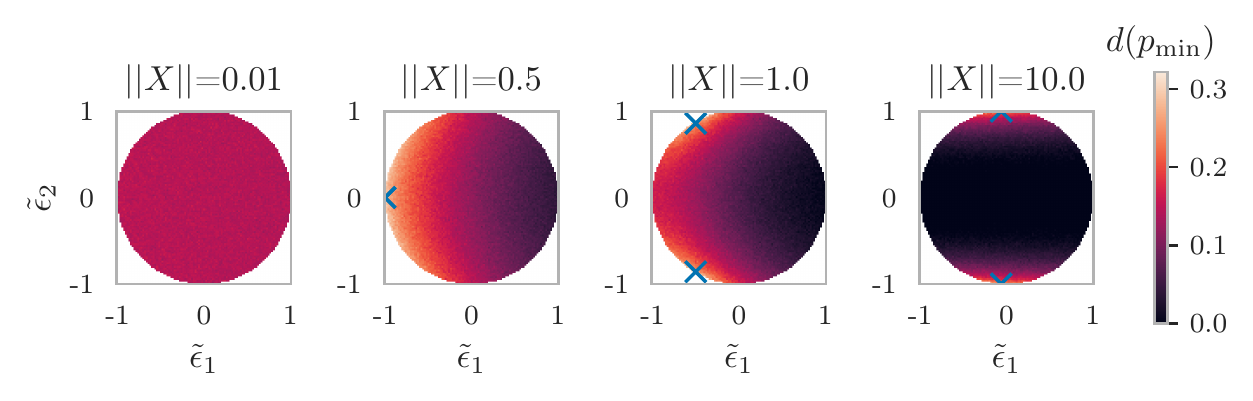}
         \vskip-0.25cm
         \caption{$||\mDelta||=\sigma$}
     \end{subfigure}
     \vskip-0.1cm
     \begin{subfigure}[b]{0.99\textwidth}
         \centering
         \includegraphics{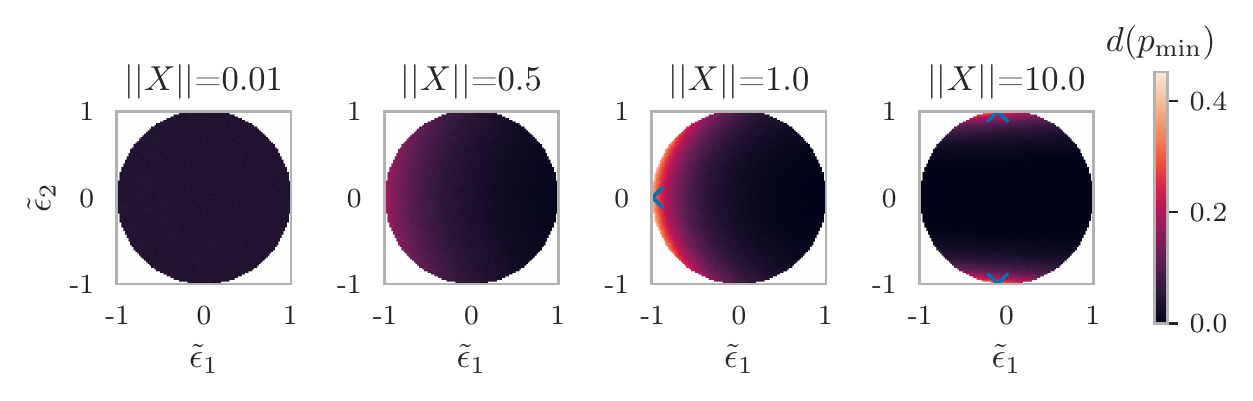}
         \vskip-0.25cm
         \caption{$||\mDelta||=2 \sigma$}
     \end{subfigure}
     \vskip-0.1cm
     \begin{subfigure}[b]{0.99\textwidth}
         \centering
         \includegraphics{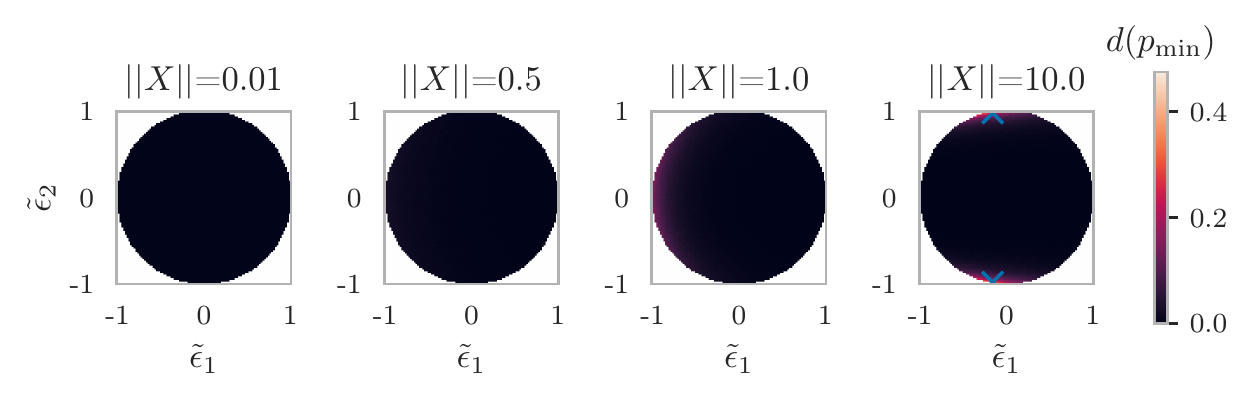}
         \vskip-0.25cm
         \caption{$||\mDelta||=3 \sigma$}
     \end{subfigure}
     \caption{Difference in $p_\mathrm{min}$ between the tight certificate for 2D rotation invariance and the black-box baseline for $\sigma=1$ under varying $||\mX||$, $||\mDelta||$, $\tilde{\epsilon}_1$ and $\tilde{\epsilon}_2$.
     Blue crosses indicate adversarial rotations.
     }
     \label{fig:appendix_inverse_multiple_inner_cross_1}
\end{figure}

\cref{fig:appendix_inverse_multiple_inner_cross_1} shows the resulting $d(p_\mathrm{min})$.
In the case where $||\mDelta||_2 \leq 2 ||\mX||_2$ we have additionally highlighted the two values of $(\tilde{\epsilon}_1, \tilde{\epsilon}_2)$ that correspond to  adversarial rotations with blue crosses (see~\cref{appendix:tight_certificate_parameters})
Note that, to improve readability, the colorbar is scaled differently for each choice of $||\mDelta||$, i.e.\ each row. Within each row, the same colorbar is used.

We can make four observations.

Firstly, in the case that $||\mDelta||_2 \leq \sigma$ and $||\mX|| = 0.01$ the tight certificate yields lower (i.e.\ better) $p_\mathrm{min}$ for arbitrary perturbations, not just those corresponding to adversarial rotations.

Secondly, as the clean data norm $||\mX||_2$ increases, the region where the tight certificate outperforms the black-box baseline shrinks.
It concentrates around values of $(\tilde{\epsilon}_1, \tilde{\epsilon}_2)$ corresponding to adversarial rotations.

Thirdly, the difference in $p_\mathrm{min}$ is not as drastic when $||\mDelta||_2$ is small.
This is to be expected. For instance, with $||\mDelta||_2 = \sigma \mathbin{/} 4$, the black-box certificate has $p_\mathrm{min} = \Phi( \sigma \mathbin{/} 4) \approx 0.6$. Here, the difference to the smallest possible value $0.5$ is small, meaning there is little room for improvement.
We already observed this in~\cref{fig:appendix_inverse_parallel}.

Finally, when $||\mDelta||_2$ is large relative to $\sigma$, and so large that $\mDelta$ cannot be the result of an adversarial rotation (e.g.\ $||\mDelta||_2 = 3, ||\mX||_2 = 0.01$), then the tight certificate is almost identical to the black-box baseline.
But, as $||\mX||_2$ increases and an adversarial rotation becomes possible (e.g.\ $||\mDelta||_2 = 3$, $||\mX||_2 = 10$) 
then the tight certificate does again yield much better $p_\mathrm{min}$ -- but only for values of $(\tilde{\epsilon}_1, \tilde{\epsilon}_2)$ that are very close to adversarial rotations.

In summary, these observations support our claim that, if $||\mX||$ is large relative to $\sigma$, then the tight certificate only improves upon the black-box baseline for perturbations that are similar to adversarial rotations.
Combined with the fact that the orbit-based gray-box certificate for rotation invariance also accounts for rotational components,
this suggests that the orbit-based certificate should be a good approximation in practice.

\begin{figure}[H]
     \centering
     \begin{subfigure}[b]{0.99\textwidth}
         \centering
         \includegraphics{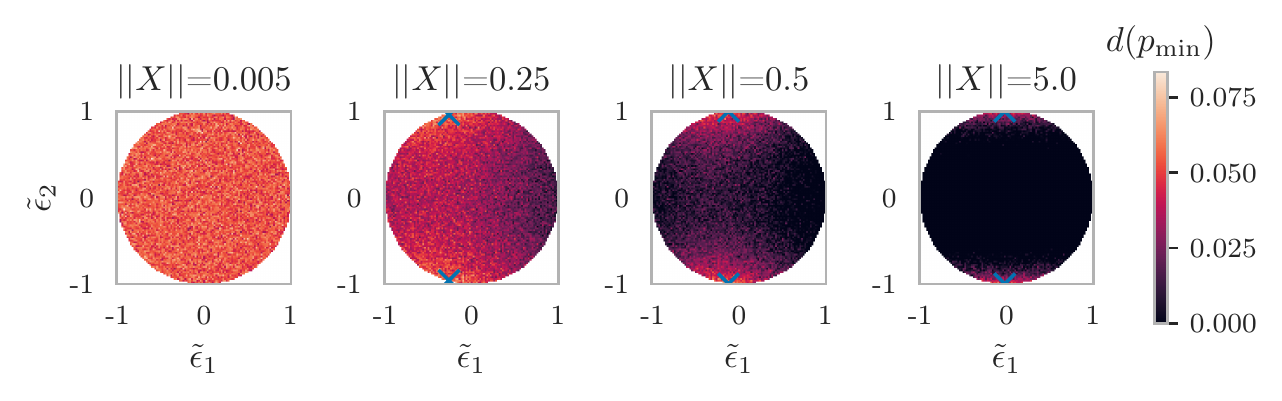}
         \vskip-0.2cm
         \caption{$||\mDelta||=\sigma \mathbin{/} 4$}
     \end{subfigure}
     \vskip-0.1cm
     \begin{subfigure}[b]{0.99\textwidth}
         \centering
         \includegraphics{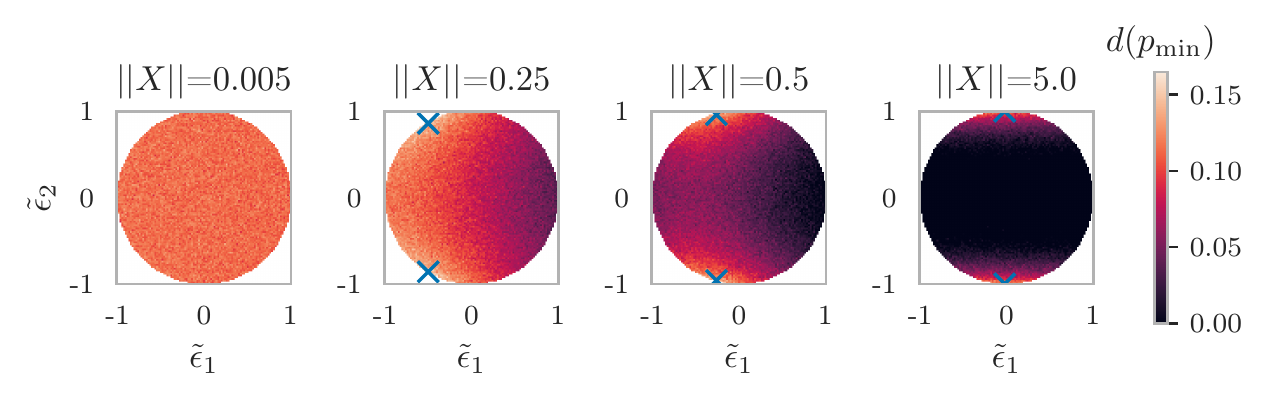}
         \vskip-0.2cm
         \caption{$||\mDelta||=\sigma \mathbin{/} 2$}
     \end{subfigure}
     \vskip-0.1cm
     \begin{subfigure}[b]{0.99\textwidth}
         \centering
         \includegraphics{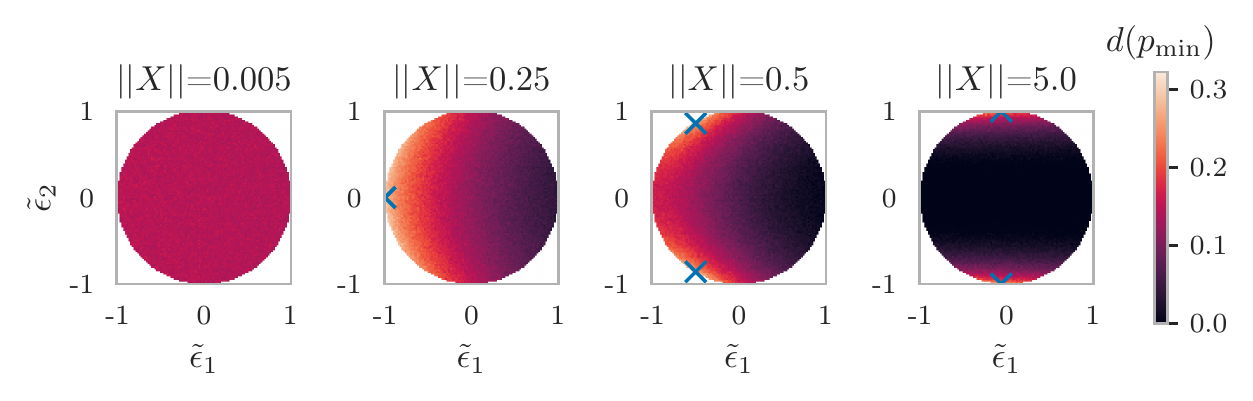}
         \vskip-0.2cm
         \caption{$||\mDelta||=\sigma$}
     \end{subfigure}
     \vskip-0.1cm
     \begin{subfigure}[b]{0.99\textwidth}
         \centering
         \includegraphics{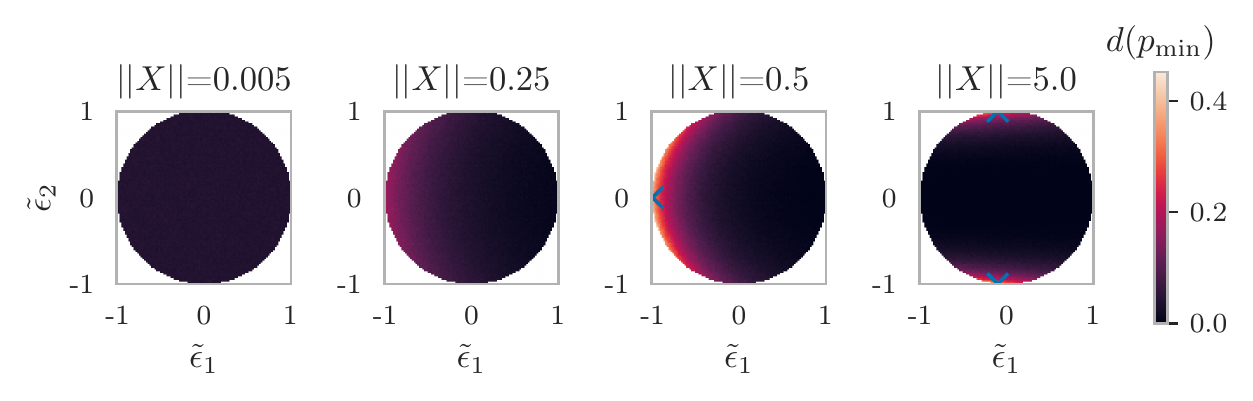}
         \vskip-0.2cm
         \caption{$||\mDelta||=2 \sigma$}
     \end{subfigure}
     \vskip-0.1cm
     \begin{subfigure}[b]{0.99\textwidth}
         \centering
         \includegraphics{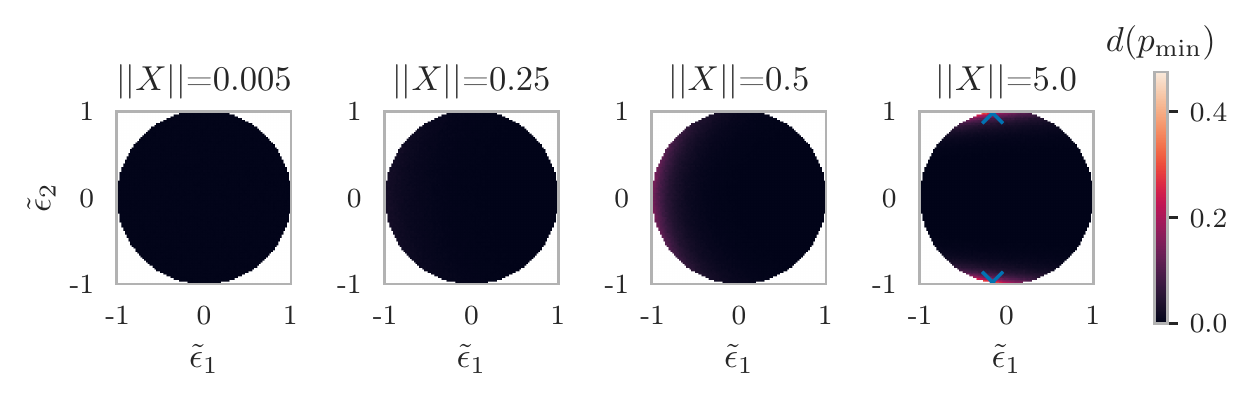}
         \vskip-0.2cm
         \caption{$||\mDelta||=3 \sigma$}
     \end{subfigure}
     \caption{Difference in $p_\mathrm{min}$ between the tight certificate for 2D rotation invariance and the black-box baseline for $\sigma=1 \mathbin{/} 2$ under varying $||\mX||$, $||\mDelta||$, $\tilde{\epsilon}_1$ and $\tilde{\epsilon}_2$.
     The results are identical to~\cref{fig:appendix_inverse_multiple_inner_cross_1}, because the certificate only depends on the value of the parameters relative to $\sigma$.}
     \label{fig:appendix_inverse_multiple_inner_cross_2}
\end{figure}

In~\cref{fig:appendix_inverse_multiple_inner_cross_2}, we repeat the same experiments with $\sigma$, $||\mX||_2$ and $||\mDelta||_2$ scaled by a factor of $\frac{1}{2}$, to verify that it is indeed sufficient to only consider a single, arbitrary value of $\sigma$ and the relative value of all other parameters.
As expected, the results are identical to~\cref{fig:appendix_inverse_multiple_inner_cross_1}.

\begin{figure}[H]
     \centering
     \begin{subfigure}[b]{0.49\textwidth}
         \centering
         \includegraphics{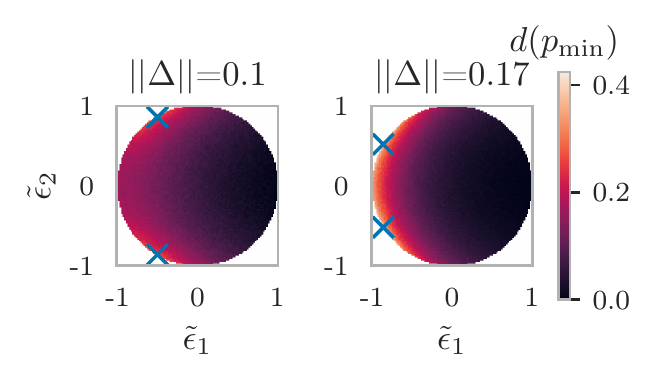}
         \vskip-0.1cm
         \caption{$\sigma = 0.1$, $||\mX||=0.1$}
     \end{subfigure}
     \hfill
     \begin{subfigure}[b]{0.49\textwidth}
         \centering
         \includegraphics{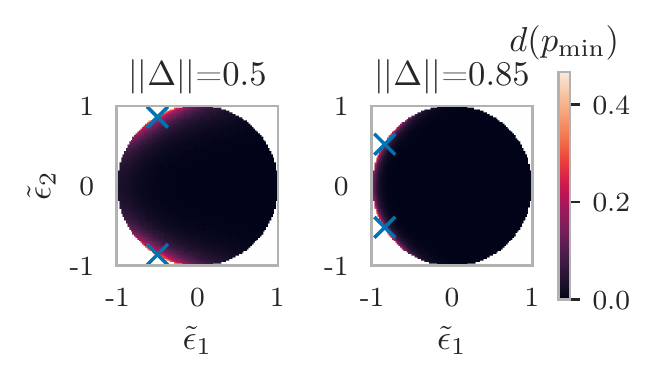}
         \vskip-0.1cm
         \caption{$\sigma=0.2$, $||\mX||=0.5$}
     \end{subfigure}
     \begin{subfigure}[b]{0.49\textwidth}
         \centering
         \includegraphics{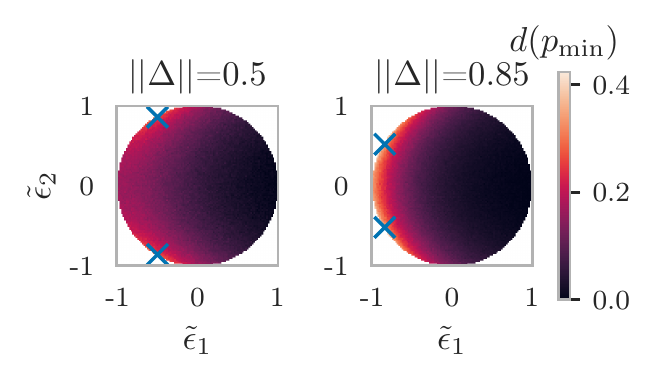}
         \vskip-0.1cm
         \caption{$\sigma = 0.5$, $||\mX||=0.5$}
     \end{subfigure}
     \hfill
     \begin{subfigure}[b]{0.49\textwidth}
         \centering
         \includegraphics{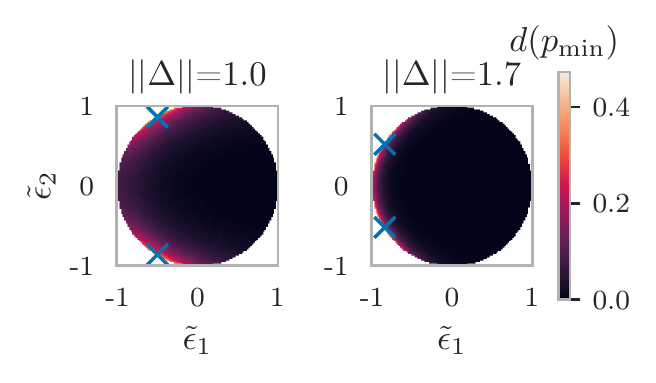}
         \vskip-0.1cm
         \caption{$\sigma=0.5$, $||\mX||=1.0$}
     \end{subfigure}
     \begin{subfigure}[b]{0.49\textwidth}
         \centering
         \includegraphics{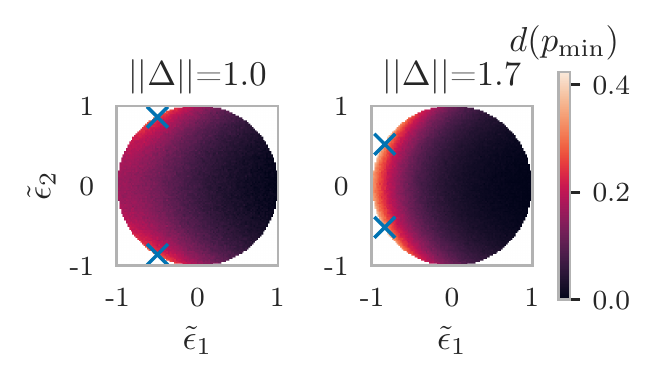}
         \vskip-0.1cm
         \caption{$\sigma = 1.0$, $||\mX||=1.0$}
     \end{subfigure}
     \hfill
     \begin{subfigure}[b]{0.49\textwidth}
         \centering
         \includegraphics{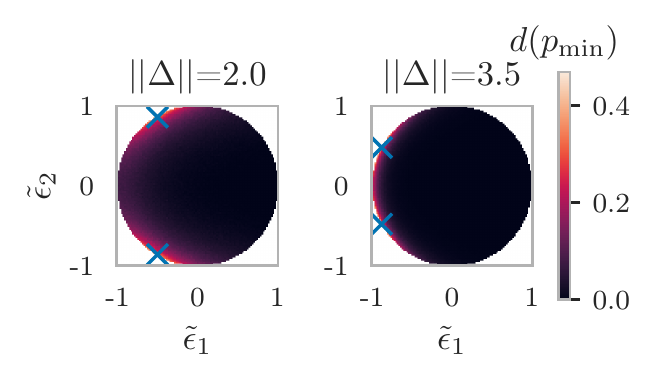}
         \vskip-0.1cm
         \caption{$\sigma=1.0$, $||\mX||=2.0$}
     \end{subfigure}
     \caption{Difference in $p_\mathrm{min}$ between the tight certificate for 2D rotation invariance and the black-box baseline for different combinations of $\sigma$, $||\mX||_2$ and $||\mDelta||_2$. Parameter $||\mDelta||_2$ is chosen such that adversarial rotations (indicated with blue crosses) correspond to $\tilde{\epsilon}= \approx 0.5$ or $\tilde{\epsilon}_2 \approx \pm 0.5$.
     The regions with large $p_\mathrm{min}$ are concentrated around adversarial rotations.}
     \label{fig:appendix_inverse_multiple_inner_cross_3}
\end{figure}

In~\cref{fig:appendix_inverse_multiple_inner_cross_3} we perform the same experiment for fixed values of $\sigma$ and $||\mX||_2$
and two values of $||\mDelta||_2$ chosen such that adversarial rotations  correspond to $\tilde{\epsilon}= \approx 0.5$ and  $\tilde{\epsilon}_2 \approx \pm 0.5$, respectively (similar to~\cref{fig:main_inverse_multiple_inner_cross_delta} from~\cref{section:main_experiments_inverse}).
In other words: The adversarial rotations move in $30^{\circ}$-increments in the parameter space.
Again, we can observe that the regions with large $p_\mathrm{min}$ move with the adversarial rotations.

\clearpage

\clearpage

\subsection{Application to point cloud classification}\label{appendix:experiments_forward}
Next, we repeat the experiments from~\cref{section:main_experiments_forward} for diverse combinations of parameters and classifier architectures.
We focus on the case $\sigma \leq 0.25$, where the randomly smoothed models achieve at least $50\%$ accuracy (see~\cref{fig:main_smoothed_accuracy}), as even smaller accuracies are too low to be of any practical use.

We first consider adversarial scaling on the point clouds constructed from MNIST (\cref{appendix:experiments_forward_scaling_mnist}).
We then consider perturbations with random rotation components on MNIST (\cref{appendix:experiments_forward_rotation_mnist}) and ModelNet40 (\cref{appendix:experiments_forward_axis_rotation_modelnet}).
For the tight certificates, we show both probabilistic upper and lower bounds obtained via Monte Carlo evaluation (recall~\cref{section:rotation_2d_tight}).
Note that only the lower bound is guaranteed to be a valid certificate with high probability.

\subsubsection{Adversarial scaling on MNIST}\label{appendix:experiments_forward_scaling_mnist}
\begin{figure}[H]
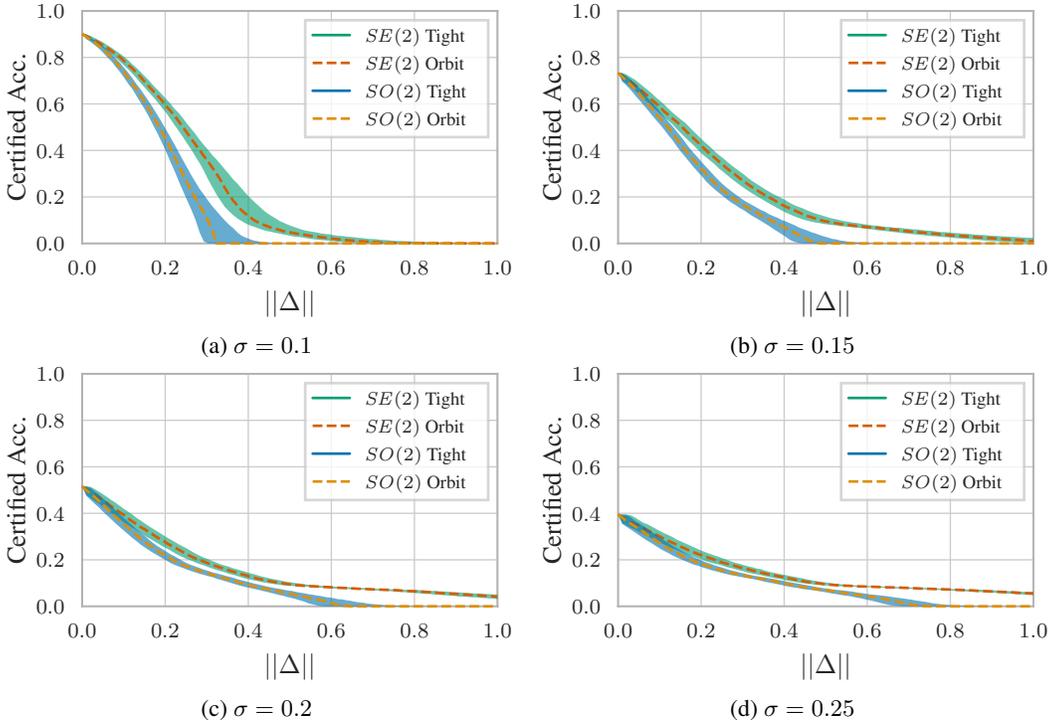

     \centering
     \begin{subfigure}[b]{0.49\textwidth}
         \centering
         \input{figures/experiments/forward/mnist_parallel/0.1.pgf}
         \caption{$\sigma=0.1$}
     \end{subfigure}
     \hfill
     \begin{subfigure}[b]{0.49\textwidth}
         \centering
         \input{figures/experiments/forward/mnist_parallel/0.15.pgf}
         \caption{$\sigma=0.15$}
     \end{subfigure}
     \begin{subfigure}[b]{0.49\textwidth}
         \centering
         \input{figures/experiments/forward/mnist_parallel/0.2.pgf}
         \caption{$\sigma=0.2$}
     \end{subfigure}
     \hfill
     \begin{subfigure}[b]{0.49\textwidth}
         \centering
         \input{figures/experiments/forward/mnist_parallel/0.25.pgf}
         \caption{$\sigma=0.25$}
     \end{subfigure}
     \caption{Comparison of certificates for adversarial scaling of MNIST with EnsPointNet and $\sigma \in \{0.1, 0.15, 0.2, 0.25\}$. Tight and orbit-based certificates yield similar certified accuracies.}
     \label{fig:appendix_experiments_forward_scaling_mnist}
\end{figure}

\cref{fig:appendix_experiments_forward_scaling_mnist} shows the certified accuracy (i.e.\ the number of correct and certifiably robust predictions)
under adversarial scaling (i.e.\ $\tilde{\epsilon}_1 = 1, \tilde{\epsilon}_2 = 0$)
for the randomly smoothed EnsPointNet with different $\sigma$ on the MNIST test set.
We evaluate the tight and orbit-based certificate for rotation invariance ($\mathit{SO}(2)$) and those for simultaneous rotation and translation invariance ($\mathit{SE}(2)$).

In all cases, the tight and orbit-based certificates yield similar certified accuracies.
The certified accuracies of the tight certificate deviate slightly from the orbit-based ones, because we compute a probabilistic bound that holds with high probability $1-\alpha$, rather than evaluating it exactly.
The gap between the probalistic lower bound and the orbit-based certificate is particularly large for $\sigma=0.1$, which can be explained by our observations from~\cref{appendix:experiments_inverse}:
The smaller $\sigma$ is, relative to $||\mX||_2$ (which is defined by each datapoint of the test set and thus constant), the smaller the benefit of using it over the orbit-based baseline becomes.

Additionally enforcing translation invariance yields stronger robustness guarantees.
For instance, with $\sigma=0.2$, the $\mathit{SE}(2)$ certificates can still certify some predictions for $||\mDelta||_2 = 0.8$, whereas the $\mathit{SO}(2)$ certificates can not certify robustness beyond $||\mDelta||_2 \approx 0.62$.

\clearpage

\subsubsection{Perturbations with random rotation components on MNIST}\label{appendix:experiments_forward_rotation_mnist}

Like in~\cref{section:main_experiments_forward}, we construct perturbations with rotational components by fixing $||\mDelta||$, randomly sampling perturbations of the specified norm 
and then rotating $\smash{\mZ = X + \mDelta}$ by a specified angle $\theta$.
We obtain the original, unrotated perturbations via sampling from an isotropic matrix normal distribution and then scaling the sample to be of the desired norm.
Per value of $\theta$, we sample ten such perturbed inputs per element of  the test set (i.e.\ $100000$ per $\theta$).
We then compare the tight and the orbit-based certificate for simultaneous rotation and translation invariance ($\mathit{SE}$), as well as the black-box baseline. 
As our metric, we use probabilistic certified accuracy, i.e.\  the percentage of samples $\mX'$ for which $f(\mX)$ is correct and $f(\mX') = f(\mX)$ is provably guaranteed.

\begin{figure}[H]
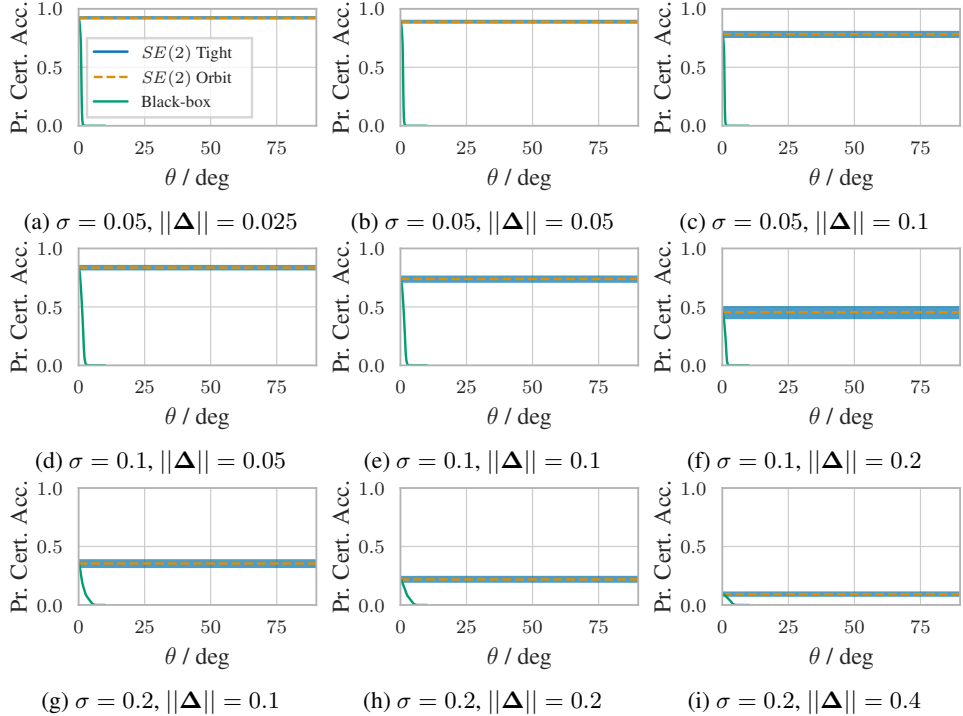

     \centering
     \begin{subfigure}[b]{0.3\textwidth}
         \centering
         \resizebox{\textwidth}{!}{\input{figures/experiments/forward/rotation/mnist/pointnet/0.05/0.025.pgf}}
         \caption{$\sigma=0.05$, $||\mDelta||=0.025$}
     \end{subfigure}
     \begin{subfigure}[b]{0.3\textwidth}
         \centering
         \resizebox{\textwidth}{!}{\input{figures/experiments/forward/rotation/mnist/pointnet/0.05/0.05.pgf}}
         \caption{$\sigma=0.05$, $||\mDelta||=0.05$}
     \end{subfigure}
     \begin{subfigure}[b]{0.3\textwidth}
         \centering
         \resizebox{\textwidth}{!}{\input{figures/experiments/forward/rotation/mnist/pointnet/0.05/0.1.pgf}}
         \caption{$\sigma=0.05$, $||\mDelta||=0.1$}
     \end{subfigure}
     \begin{subfigure}[b]{0.3\textwidth}
         \centering
         \resizebox{\textwidth}{!}{\input{figures/experiments/forward/rotation/mnist/pointnet/0.1/0.05.pgf}}
         \caption{$\sigma=0.1$, $||\mDelta||=0.05$}
     \end{subfigure}
     \begin{subfigure}[b]{0.3\textwidth}
         \centering
         \resizebox{\textwidth}{!}{\input{figures/experiments/forward/rotation/mnist/pointnet/0.1/0.1.pgf}}
         \caption{$\sigma=0.1$, $||\mDelta||=0.1$}
     \end{subfigure}
     \begin{subfigure}[b]{0.3\textwidth}
         \centering
         \resizebox{\textwidth}{!}{\input{figures/experiments/forward/rotation/mnist/pointnet/0.1/0.2.pgf}}
         \caption{$\sigma=0.1$, $||\mDelta||=0.2$}
     \end{subfigure}
     \begin{subfigure}[b]{0.3\textwidth}
         \centering
         \resizebox{\textwidth}{!}{\input{figures/experiments/forward/rotation/mnist/pointnet/0.2/0.1.pgf}}
         \caption{$\sigma=0.2$, $||\mDelta||=0.1$}
     \end{subfigure}
     \begin{subfigure}[b]{0.3\textwidth}
         \centering
         \resizebox{\textwidth}{!}{\input{figures/experiments/forward/rotation/mnist/pointnet/0.2/0.2.pgf}}
        \caption{$\sigma=0.2$, $||\mDelta||=0.2$}
     \end{subfigure}
     \begin{subfigure}[b]{0.3\textwidth}
         \centering
         \resizebox{\textwidth}{!}{\input{figures/experiments/forward/rotation/mnist/pointnet/0.2/0.4.pgf}}
         \caption{$\sigma=0.2$, $||\mDelta||=0.4$}
     \end{subfigure}
     \caption{Comparison of tight gray-box, orbit-based gray-box and black-box certificates for MNIST with EnsPointNet and different $\sigma$,
     with respect to probabilistic certified accuracy. Perturbed inputs are generated by sampling perturbations with $||\mDelta||_2 \in \{\sigma \mathbin{/} 2, \sigma, 2 \sigma\}$ and rotating by angle $\theta$.
     Group $\mathit{SE}(2)$ refers to simultaneous rotation and translation invariance.
     The orbit-based certificate is a good approximation of the tight certificate.
     }
     \label{fig:appendix_experiments_forward_rotation_mnist}
\end{figure}

\cref{fig:appendix_experiments_forward_rotation_mnist} shows the results for $\sigma \in \{0.05, 0.1, 0.2\}$,  $||\mDelta||_2 \in \{\sigma \mathbin{/} 2, \sigma, 2 \sigma\}$ and $\theta \in [0^{\circ}, 90^{\circ}]$.
While the black-box baseline fails to certify robustness even for small rotation angles, both gray-box certificates effectively eliminate the rotations, i.e.\ are constant in $\theta$.
However, the tight gray-box certificate does not offer any noteable benefit beyond that. The probabilistic lower bound  never yields better probabilistic certified accuracy than the orbit-based certificate.

These results are consistent with our observartions about the tight certificate's parameter space:
All values of $\sigma$ that retain high accuracy are small, relative to the average norm of the test set ($10.67$).

\clearpage

\subsubsection{Perturbations with random elemental rotation components on ModelNet40}\label{appendix:experiments_forward_axis_rotation_modelnet}

Next, we repeat the same experiment on ModelNet40.
After sampling a perturbation $\mDelta$ of specified norm $||\mDelta||_2$, we randomly rotate $\mX' = \mX + \mDelta$ by angle $\theta$ around a random axis. We obtain this axis by  sampling from a three-dimensional isotropic normal distribution and normalizing the result.

We then compare the tight certificate for simultaneous invariance under rotation and translation to the orbit-based one and the black-box baseline.

\begin{figure}[H]
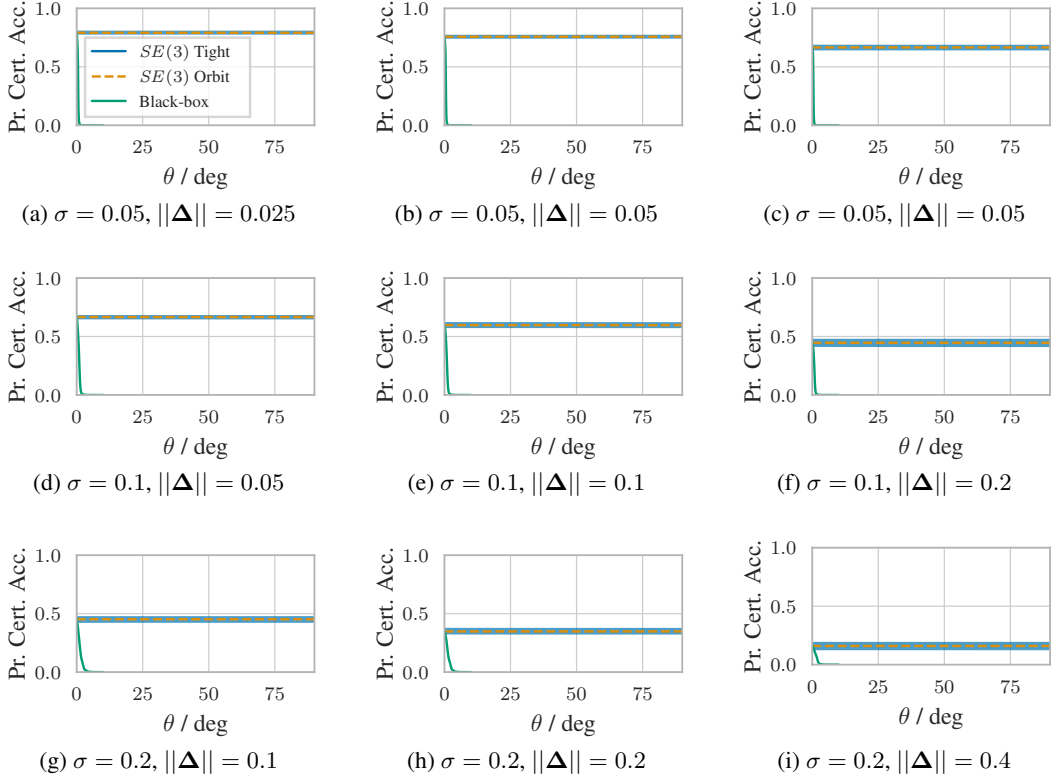

     \centering
     \begin{subfigure}[b]{0.3\textwidth}
         \centering
         \resizebox{\textwidth}{!}{\input{figures/experiments/forward/rotation/modelnet/pointnet/arbitrary/0.05/0.025.pgf}}
         \vskip-0.1cm
         \caption{$\sigma=0.05$, $||\mDelta||=0.025$}
     \end{subfigure}
     \hfill
     \begin{subfigure}[b]{0.3\textwidth}
         \centering
         \resizebox{\textwidth}{!}{\input{figures/experiments/forward/rotation/modelnet/pointnet/arbitrary/0.05/0.05.pgf}}
         \vskip-0.1cm
         \caption{$\sigma=0.05$, $||\mDelta||=0.05$}
     \end{subfigure}
     \hfill
     \begin{subfigure}[b]{0.3\textwidth}
         \centering
         \resizebox{\textwidth}{!}{\input{figures/experiments/forward/rotation/modelnet/pointnet/arbitrary/0.05/0.1.pgf}}
         \vskip-0.1cm
         \caption{$\sigma=0.05$, $||\mDelta||=0.05$}
     \end{subfigure}
     \vskip0.5cm
     \begin{subfigure}[b]{0.3\textwidth}
         \centering
         \resizebox{\textwidth}{!}{\input{figures/experiments/forward/rotation/modelnet/pointnet/arbitrary/0.1/0.05.pgf}}
         \vskip-0.1cm
         \caption{$\sigma=0.1$, $||\mDelta||=0.05$}
     \end{subfigure}
     \hfill
     \begin{subfigure}[b]{0.3\textwidth}
         \centering
         \resizebox{\textwidth}{!}{\input{figures/experiments/forward/rotation/modelnet/pointnet/arbitrary/0.1/0.1.pgf}}
         \vskip-0.1cm
         \caption{$\sigma=0.1$, $||\mDelta||=0.1$}
     \end{subfigure}
     \hfill
     \begin{subfigure}[b]{0.3\textwidth}
         \centering
         \resizebox{\textwidth}{!}{\input{figures/experiments/forward/rotation/modelnet/pointnet/arbitrary/0.1/0.2.pgf}}
         \vskip-0.1cm
         \caption{$\sigma=0.1$, $||\mDelta||=0.2$}
     \end{subfigure}
     \vskip0.5cm
     \begin{subfigure}[b]{0.3\textwidth}
         \centering
         \resizebox{\textwidth}{!}{\input{figures/experiments/forward/rotation/modelnet/pointnet/arbitrary/0.2/0.1.pgf}}
         \vskip-0.1cm
         \caption{$\sigma=0.2$, $||\mDelta||=0.1$}
     \end{subfigure}
     \hfill
     \begin{subfigure}[b]{0.3\textwidth}
         \centering
         \resizebox{\textwidth}{!}{\input{figures/experiments/forward/rotation/modelnet/pointnet/arbitrary/0.2/0.2.pgf}}
         \vskip-0.1cm
        \caption{$\sigma=0.2$, $||\mDelta||=0.2$}
     \end{subfigure}
     \hfill
     \begin{subfigure}[b]{0.3\textwidth}
         \centering
         \resizebox{\textwidth}{!}{\input{figures/experiments/forward/rotation/modelnet/pointnet/arbitrary/0.2/0.4.pgf}}
         \caption{$\sigma=0.2$, $||\mDelta||=0.4$}
     \end{subfigure}
     \caption{Comparison of tight gray-box, orbit-based gray-box and black-box certificates for ModelNet40 with EnsPointNet and different $\sigma$,
     w.r.t.\ probabilistic certified accuracy. Perturbed inputs are generated by sampling perturbations with $||\mDelta||_2 \in \{\sigma \mathbin{/} 2, \sigma, 2 \sigma\}$ and rotating around a random axis by angle $\theta$.
     Group $\mathit{SE}(3)$ refers to simultaneous rotation and translation invariance.
     The orbit-based certificate is a good approximation of the tight certificate.
     }
     \label{fig:appendix_experiments_forward_axis_rotation_modelnet_pointnet}
\end{figure}

\begin{figure}[H]
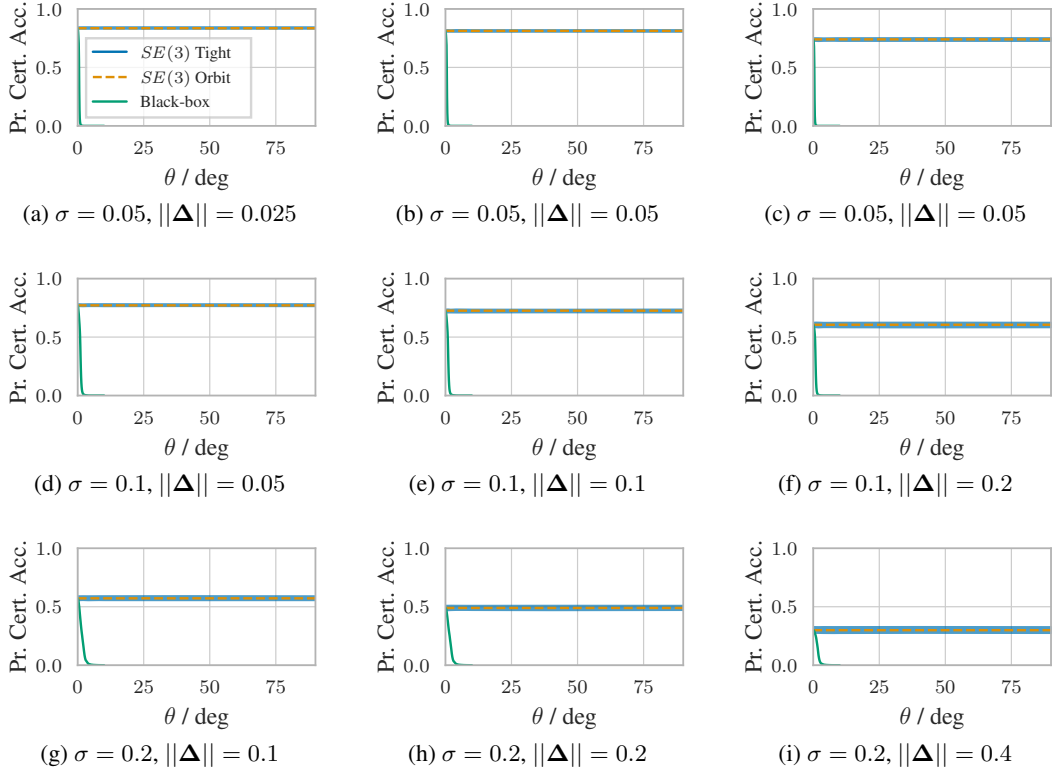

     \centering
     \begin{subfigure}[b]{0.3\textwidth}
         \centering
         \resizebox{\textwidth}{!}{\input{figures/experiments/forward/rotation/modelnet/dgcnn/arbitrary/0.05/0.025.pgf}}
         \vskip-0.1cm
         \caption{$\sigma=0.05$, $||\mDelta||=0.025$}
     \end{subfigure}
     \hfill
     \begin{subfigure}[b]{0.3\textwidth}
         \centering
         \resizebox{\textwidth}{!}{\input{figures/experiments/forward/rotation/modelnet/dgcnn/arbitrary/0.05/0.05.pgf}}
         \vskip-0.1cm
         \caption{$\sigma=0.05$, $||\mDelta||=0.05$}
     \end{subfigure}
     \hfill
     \begin{subfigure}[b]{0.3\textwidth}
         \centering
         \resizebox{\textwidth}{!}{\input{figures/experiments/forward/rotation/modelnet/dgcnn/arbitrary/0.05/0.1.pgf}}
         \vskip-0.1cm
         \caption{$\sigma=0.05$, $||\mDelta||=0.05$}
     \end{subfigure}
     \vskip0.5cm
     \begin{subfigure}[b]{0.3\textwidth}
         \centering
         \resizebox{\textwidth}{!}{\input{figures/experiments/forward/rotation/modelnet/dgcnn/arbitrary/0.1/0.05.pgf}}
         \vskip-0.1cm
         \caption{$\sigma=0.1$, $||\mDelta||=0.05$}
     \end{subfigure}
     \hfill
     \begin{subfigure}[b]{0.3\textwidth}
         \centering
         \resizebox{\textwidth}{!}{\input{figures/experiments/forward/rotation/modelnet/dgcnn/arbitrary/0.1/0.1.pgf}}
         \vskip-0.1cm
         \caption{$\sigma=0.1$, $||\mDelta||=0.1$}
     \end{subfigure}
     \hfill
     \begin{subfigure}[b]{0.3\textwidth}
         \centering
         \resizebox{\textwidth}{!}{\input{figures/experiments/forward/rotation/modelnet/dgcnn/arbitrary/0.1/0.2.pgf}}
         \vskip-0.1cm
         \caption{$\sigma=0.1$, $||\mDelta||=0.2$}
     \end{subfigure}
     \vskip0.5cm
     \begin{subfigure}[b]{0.3\textwidth}
         \centering
         \resizebox{\textwidth}{!}{\input{figures/experiments/forward/rotation/modelnet/dgcnn/arbitrary/0.2/0.1.pgf}}
         \vskip-0.1cm
         \caption{$\sigma=0.2$, $||\mDelta||=0.1$}
     \end{subfigure}
     \hfill
     \begin{subfigure}[b]{0.3\textwidth}
         \centering
         \resizebox{\textwidth}{!}{\input{figures/experiments/forward/rotation/modelnet/dgcnn/arbitrary/0.2/0.2.pgf}}
         \vskip-0.1cm
        \caption{$\sigma=0.2$, $||\mDelta||=0.2$}
     \end{subfigure}
     \hfill
     \begin{subfigure}[b]{0.3\textwidth}
         \centering
         \resizebox{\textwidth}{!}{\input{figures/experiments/forward/rotation/modelnet/dgcnn/arbitrary/0.2/0.4.pgf}}
         \vskip-0.1cm
         \caption{$\sigma=0.2$, $||\mDelta||=0.4$}
     \end{subfigure}
     \caption{Comparison of tight gray-box, orbit-based gray-box and black-box certificates for ModelNet40 with EnsDGCNN and different $\sigma$,
     w.r.t.\ probabilistic certified accuracy. Perturbed inputs are generated by sampling perturbations with $||\mDelta||_2 \in \{\sigma \mathbin{/} 2, \sigma, 2 \sigma\}$ and rotating around a random axis by angle $\theta$.
     Group $\mathit{SE}(3)$ refers to simultaneous rotation and translation invariance.
The orbit-based certificate is a good approximation of the tight certificate.
     }
     \label{fig:appendix_experiments_forward_axis_rotation_modelnet_dgcnn}
\end{figure}

\cref{fig:appendix_experiments_forward_axis_rotation_modelnet_pointnet,fig:appendix_experiments_forward_axis_rotation_modelnet_pointnet_attention,fig:appendix_experiments_forward_axis_rotation_modelnet_dgcnn} show the results for EnsPointNet, AttnPointNet and EnsDGCNN.
In all cases, both the tight- and orbit-based certificate are constant in $\theta$, but the tight certificate does not noticeably improve upon the orbit-based one w.r.t.\ probabilistic certified accuracy.

\clearpage

\begin{figure}[H]
     \centering
     \begin{subfigure}[b]{0.3\textwidth}
         \centering
         \resizebox{\textwidth}{!}{\input{figures/experiments/forward/rotation/modelnet/pointnet_attention/arbitrary/0.05/0.025.pgf}}
         \vskip-0.1cm
         \caption{$\sigma=0.05$, $||\mDelta||=0.025$}
     \end{subfigure}
     \hfill
     \begin{subfigure}[b]{0.3\textwidth}
         \centering
         \resizebox{\textwidth}{!}{\input{figures/experiments/forward/rotation/modelnet/pointnet_attention/arbitrary/0.05/0.05.pgf}}
         \vskip-0.1cm
         \caption{$\sigma=0.05$, $||\mDelta||=0.05$}
     \end{subfigure}
     \hfill
     \begin{subfigure}[b]{0.3\textwidth}
         \centering
         \resizebox{\textwidth}{!}{\input{figures/experiments/forward/rotation/modelnet/pointnet_attention/arbitrary/0.05/0.1.pgf}}
         \vskip-0.1cm
         \caption{$\sigma=0.05$, $||\mDelta||=0.05$}
     \end{subfigure}
     \vskip0.5cm
     \begin{subfigure}[b]{0.3\textwidth}
         \centering
         \resizebox{\textwidth}{!}{\input{figures/experiments/forward/rotation/modelnet/pointnet_attention/arbitrary/0.1/0.05.pgf}}
         \vskip-0.1cm
         \caption{$\sigma=0.1$, $||\mDelta||=0.05$}
     \end{subfigure}
     \hfill
     \begin{subfigure}[b]{0.3\textwidth}
         \centering
         \resizebox{\textwidth}{!}{\input{figures/experiments/forward/rotation/modelnet/pointnet_attention/arbitrary/0.1/0.1.pgf}}
         \vskip-0.1cm
         \caption{$\sigma=0.1$, $||\mDelta||=0.1$}
     \end{subfigure}
     \hfill
     \begin{subfigure}[b]{0.3\textwidth}
         \centering
         \resizebox{\textwidth}{!}{\input{figures/experiments/forward/rotation/modelnet/pointnet_attention/arbitrary/0.1/0.2.pgf}}
         \vskip-0.1cm
         \caption{$\sigma=0.1$, $||\mDelta||=0.2$}
     \end{subfigure}
     \vskip0.5cm
     \begin{subfigure}[b]{0.3\textwidth}
         \centering
         \resizebox{\textwidth}{!}{\input{figures/experiments/forward/rotation/modelnet/pointnet_attention/arbitrary/0.2/0.1.pgf}}
         \vskip-0.1cm
         \caption{$\sigma=0.2$, $||\mDelta||=0.1$}
     \end{subfigure}
     \hfill
     \begin{subfigure}[b]{0.3\textwidth}
         \centering
         \resizebox{\textwidth}{!}{\input{figures/experiments/forward/rotation/modelnet/pointnet_attention/arbitrary/0.2/0.2.pgf}}
         \vskip-0.1cm
        \caption{$\sigma=0.2$, $||\mDelta||=0.2$}
     \end{subfigure}
     \hfill
     \begin{subfigure}[b]{0.3\textwidth}
         \centering
         \resizebox{\textwidth}{!}{\input{figures/experiments/forward/rotation/modelnet/pointnet_attention/arbitrary/0.2/0.4.pgf}}
         \vskip-0.1cm
         \caption{$\sigma=0.2$, $||\mDelta||=0.4$}
     \end{subfigure}
     \caption{Comparison of tight gray-box, orbit-based gray-box and black-box certificates for ModelNet40 with AttnPointNet and different $\sigma$,
     w.r.t.\ probabilistic certified accuracy. Perturbed inputs are generated by sampling perturbations with $||\mDelta||_2 \in \{\sigma \mathbin{/} 2, \sigma, 2 \sigma\}$ and rotating around a random axis by angle $\theta$.
     Group $\mathit{SE}(3)$ refers to simultaneous rotation and translation invariance.
     The orbit-based certificate is a good approximation of the tight certificate.
     }
     \label{fig:appendix_experiments_forward_axis_rotation_modelnet_pointnet_attention}
\end{figure}

\clearpage

\clearpage

\clearpage

\section{Full experimental setup}\label{appendix:full_experimental_setup}
In the following, we provide a full specification of the used models,
data sets and training parameters (\cref{appendix:hyperparams},
randomized smoothing parameters (\cref{appendix:other_params}), as well as the used computational resources (\cref{appendix:compute})
and third-party assets (\cref{appendix:assets}).

The code and all configuration files needed for reproducing the experimental results will be made available at  \href{https://www.cs.cit.tum.de/daml/invariance-smoothing/}{https://www.cs.cit.tum.de/daml/invariance-smoothing}.

\subsection{Models, data and training parameters}\label{appendix:hyperparams}
\subsubsection{Models}
The experiments in~\cref{section:main_experiments_forward,appendix:experiments_forward} are performed with three models: EnsPointNet, EnsDGCNN and AttnPointNet, which are rotation and translation invariant versions of PointNet~\cite{Qi2017},  and the Dynamic Graph Convolutional Neural Network~\cite{Wang2019}.

\textbf{EnsPointNet}
is based on a standard PointNet architecture with an input T-Net but without a feature T-Net (we did not find the feature T-Net to improve the accuracy of the rotation invariant model).
The input T-Net uses three convolutional layers with kernel size $1 \times 1$ ($64$, $128$ and $1024$ filters) and three linear layers ($512$, $256$ and $9$ neurons).
The PointNet itself uses three convolutional layers with kernel size $1 \times 1$ ($64$, $128$ and $1024$ filters) and three linear layers ($512$, $256$ and $|\sY|$ neurons).
All layers, except the last one, use BatchNorm ($\epsilon=1e-05$, $\mathrm{momentum}=0.1$). The second linear layer uses dropout ($p=0.4$).
To achieve rotation and translation invariance, the input data is centered and the two (for $2D$ data) or three (for $3D$ data) principal components are computed.
Principal components are not unique. One has to account for order and sign ambiguities to ensure rotation invariance (see discussion in~\cite{Xiao2020,Li2021attn}).
If two or more eigenvalues are numerically close (relative tolerance $1e^{-5}$, absolute tolerance $1e^{-8})$, we iterate over all possible eigenvector signs and orders ($8 \cdot 6$), multiply the input data with the principal components, and pass it through the PointNet.
If the eigenvalues are distinct, we sort them in ascending order and iterate over all possible signs ($8$).
The $8$ or $48$ logit vectors are then averaged to obtain a prediction.

\textbf{EnsDGCNN}
 is based on a standard DGCNN architecture with an input spatial transform.
The spatial transform uses uses three convolutional layers with kernel size $1 \times 1$ ($64$, $128$ and $1024$ filters) and three linear layers ($512$, $256$ and $9$ neurons).
The DGCNN encodes the spatially transformed input with three DGC layers ($64$, $64$, $64$ and $128$ filters and $k=20$). The encoder output and residuals are concatenated and passed through a convolution with kernel size $1 \times 1$ ($1024$ filters), followed by max-pooling and three linear layers ($512$, $256$, $|\sY|$ neurons).
All layers use BatchNorm ($\epsilon=1e-05$, $\mathrm{momentum}=0.1$). The first two linear layers use dropout ($p=0.5$).
We use the same ensembling approach as for EnsPointNet to achieve rotation and translation invariance.

\textbf{AttnPointNet}
 combines PointNet with the attention-based mechanism for combining canonical poses proposed in~\cite{Xiao2020}.
It uses the same encoder as EnsPointNet.
After passing the different PCA-based canonical poses through the encoder, the hidden vectors are combined via a self-attention layer ($1024$ neurons each for query, key  and value transform) and then passed through the same decoder as in EnsPointNet.
To reduce memory usage, we only consider sign combinations that correspond to proper rotation matrices (see discussion in~\cite{Li2021attn}).

\subsubsection{Data}

\textbf{ModelNet40}~\cite{Wu2015} consists of $12311$ CAD models from $40$ categories, split into $9843$ training samples and $2468$ test samples.
We subdivide the original training set into $80\%$ train data and $20\%$ validation data. The same split is used for all experiments.
Each CAD model is then transformed into a 3D point cloud with $1024$ points using the same pre-processing steps as in~\cite{Qi2017}, i.e.\
randomly sampling from the mesh faces according to surface area and normalizing the resulting point cloud into the unit sphere.

\textbf{MNIST}~\cite{Lecun1998} consists of $70000$ handwritten digits, split into $60000$ training samples and $10000$ test samples.
We subdivide the original training set into $80\%$ train data and $20\%$ validation data. The same split is used for all experiments.
Each image is transformed into a 2D point with $1024$ points using the same pre-processing steps as in~\cite{Qi2017}, i.e.\ 
mapping all pixels with values greater than $128$ to x-y coordinates, then uniformly sub-sampling or padding with a randomly chosen point and finally normalizing into the unit sphere.
We additionally combine classes $6$ and $9$, because these digits cannot be expected to be differentiated by a rotation-invariant model.

\subsubsection{Training parameters}
All models are trained on samples from the smoothing distribution (i.e.\ the model that is evaluated and certified with $\sigma$ is trained on data augmented with Gaussian noise sampled from an isotropic normal distribution with standard deviation $\sigma$).
Per element of a training batch, we take exactly one sample from the smoothing distribution.
We do not use consistency regularization.

\textbf{EnsPointNet} is trained with cross entropy loss and  the TNet regularization from~\cite{Qi2017} (weight $0.0001$).
Training is performed for $200$ epochs with batch size $128$ using Adam ($\beta_1 = 0.9, \beta_2 = 0.99$, $\epsilon=1e-8$, $\mathrm{weight\_decay} = 1e^-4$).
The learning rate starts at $0.001$ and is multiplied with $0.7$ every $20$ epochs.
The input data is randomly scaled by a factor from $[0.8, 1.25]$
and then transformed via principal component analysis, using a single set of eigenvectors with randomized sign and order.

\textbf{EnsDGCNN} is trained in the same manner as EnsPointNet, but with batch size $32$, $400$ epochs and learning rate decay every $40$ epochs.

\textbf{AttnPointNet} is trained with batch size $64$, $400$ epochs and learning rate decay every $40$ epochs. The other parameter values are identical to those for EnsPointNet.
In order to train the self-attention layer, we do not transform the input with a single set of eigenvectors, but pass all possible eigenvectors corresponding to rotations through the network.

\subsection{Randomized smoothing parameters}\label{appendix:other_params}
We use $1000$ samples from the smoothing distribution to compute smoothed predictions. Abstentions, i.e.\ predictions for which we cannot guarantee that $p_{\mX,y^*} \geq 0.5$, are considered as incorrect.

For certification, we set the significance parameter $\alpha$ to $0.01$. i.e.\ each certificate holds with probability $99.9\%$.
We use $10000$ samples per confidence bound, i.e.\ $10000$ samples to bound $p_{\mX,y^*}$. For the tight gray-box certificates involving rotation invariance we additionally use $10000$ samples to bound the threshold $\kappa$ and $10000$ samples to bound the optimal value itself (for more details, see~\cref{appendix:monte_carlo_certification}).
We then use Holm-Bonferroni correction (\cite{Holm1979}) to ensure that all three confidence bounds hold simultaneously with probability $1- \alpha$.

Evaluating the tight certificates for $\mathit{SO}(3)$ and $\mathit{SE}(3)$ requires numerical integration over the two-dimensional parameter space $[-\frac{\pi}{2},\frac{\pi}{2}] \times [0, 2\pi]$ (see~\cref{appendix:tight_rotation_3d}).
We use Clenshaw-Curtis quadrature with degree $20$ in each dimension.

\subsection{Computational resources}\label{appendix:compute}
\textbf{Training and  Monte Carlo sampling of smoothed predictions} was performed on a single NVIDIA A100 GPU (\SI{40}{\giga\byte} VRAM) with an AMD EPYC 7543 CPU @ \SI{2.8}{\giga\hertz}. For EnsPointNet and AttnPointNet, \SI{16}{\giga\byte} RAM were allocated. For EnsDGCNN, which requires additional memory for computation of k-nearest-neighbor graphs, \SI{32}{\giga\byte} RAM were allocated.
The average time for training EnsPointNet on MNIST was \SI{42.8}{\minute}.
The average time for training EnsPointNet, AttnPoint and EnsDGCNN on ModelNet40 was \SI{20.2}{\minute}, \SI{114}{\minute} and \SI{158}{\minute}, respectively.
The average time for obtaining $11000$ samples from EnsPointNet on MNIST was \SI{1.1}{\second}.
The average time for obtaining $11000$ samples from EnsPointNet, AttnPoint and EnsDGCNN on ModelNet40 was \SI{1.59}{\second}, \SI{1.4}{\second} and \SI{24.91}{\second}, respectively.

\textbf{Computation of certificates} was performed on an Intel Xeon E5-2630 v4 CPU @ \SI{2.2}{\giga\hertz}, with \SI{16}{\giga\byte} RAM allocated to each experiment.
The average time for computing a tight certificate for 2D rotation invarance (i.e.\ computing threshold $\kappa$ and bounding the optimal value, using $10000$ samples each),
was \SI{0.05}{\second}.
The average time for computing a tight certificate for 3D rotation invariance was \SI{6.3}{\second}.
The increase in computational cost is caused by the fact that we can only evaluate the worst-case classifier via numerical integration.
The average time for computing the black-box randomized smoothing certificate and the tight certificate for translation invariance was in the sub-milisecond range and can thus not be accurately reported.

\subsection{Third-party assets}\label{appendix:assets}
Our work uses the publicly available MNIST~\cite{Lecun1998} and ModelNet40~\cite{Wu2015} datasets, as well as the quadpy quadrature library~\cite{quadpy}.
Our implementation additionally uses code from a publicly available implementation of PointNet and PointNet++~\cite{pytorch_pointnet}, as well as the author's reference implementation of DGCNN~\cite{Wang2019}. Both are available under MIT license.

\clearpage

\changelocaltocdepth{1}
\section{Group definitions}\label{appendix:group_definitions}
In the following, we define the different groups we consider in our work, each corresponding to a specific type of spatial invariance.
Recall from~\cref{section:background_invariance} that a group is a set $\sT$, associated with a binary operator $\cdot : \sT \times \sT \rightarrow \sT$
that is closed and associative under the operation, has an inverse $t^{-1}$ for each element $t \in \sT$ and features an identity element $e$.
Further recall that we associate each group with a group action $\circ : \sT \times \sR^{N \times D} \rightarrow \sR^{N \times D}$ that describes how elements of the group modify elements of the input space.

\subsection{Translation group \texorpdfstring{$\mathit{T(D)}$}{T(D)}}
\label{appendix:group_definition_translation}
The translation group in $D$ dimensions is the set of all $D$-dimensional continuous vectors, i.e.\ $\mathit{T}(D) = \sR^{D}$.
Two group elements $\vt, \vt' \in \mathit{T}(D)$ are combined via vector addition, i.e.\ $\vt \cdot \vt' = \vt + \vt'$.
A group element $\vt \in \mathit{T}(D)$ acts on an input $\mX \in \sR^{N \times D}$ via row-wise addition, i.e.\
$\vt \circ \mX = \mX + \ones_N t^T$ with all-ones vector $\ones_N \in \sR^{N}$.
The identity element $e$ is the all-zeros vector $\zeros_D$.

\subsection{Rotation group \texorpdfstring{$\mathit{SO}(D)$}{SO(D)}}
\label{appendix:group_definition_rotation}
The rotation group in $D$ dimensions, $\mathit{SO}(D)$ (short for \textit{special orthogonal}),
is the set of all $D$-dimensional rotation matrices, i.e.\
$\mathit{SO}(D) = \left\{\mR \in \sR^{D \times D} \mid \mR^T \mR = \eye_N \land \mathrm{det}(\mR) = 1\right\}$.
Two group elements $\mR, \mR' \in \mathit{SO}(D)$ are combined via matrix multiplication, i.e.\ $\mR \cdot \mR' = \mR \mR'$.
A group element $\mR \in \mathit{SO}(D)$ acts on an input $\mX \in \sR^{N \times D}$ via row-wise matrix-vector multiplication,  i.e.\
$\mR \circ \mX = \mX \mR^T$.
The identity element $e$ is the identity matrix $\eye_D$.

\subsection{Orthogonal group \texorpdfstring{$\mathit{O}(D)$}{O(D)}}
\label{appendix:group_definition_orthogonal}
The orthogonal group in $D$ dimensions, $\mathit{O}(D)$, is the set of all $D$-dimensional orthogonal matrices, i.e.\ 
$\mathit{O}(D) = \left\{\mA \in \sR^{D \times D} \mid \mR^T \mR = \eye_N\right\}$.
Note that $\mathit{O}(D) \supset \mathit{SO}(D)$.
Like the rotation group, group elements are combined via matrix multiplication, i.e.\ $\mA \cdot \mA' = \mA \mA'$ and act on an input $\mX \in \sA^{N \times D}$ via row-wise matrix-vector multiplication,  i.e.\
$\mA \circ \mX = \mX \mA^T$.
The identity element $e$ is the identity matrix $\eye_D$.

\subsection{Roto-translation group \texorpdfstring{$\mathit{SE}(D)$}{SE(D)}}
\label{appendix:group_definition_special_euclidean}
The roto-translation group in $D$ dimensions, $\mathit{SE}(D)$ (short for \textit{special Euclidean}),
is composed of all pairs of $D$-dimensional rotation matrices and translation vectors.
That is, $\mathit{SE}(D) = \mathit{SO}(D) \times \mathit{T}(D)$.
Two group elements $(\mR, \vt), (\mR', \vt') \in \mathit{SE}(D)$ are combined via a semidirect product, i.e.\
$(\mR, \vt) \cdot (\mR', \vt') = (\mR \mR', \mR \vt' + \vt)$.
A group element $(\mR, \vt) \in \mathit{SE}(D)$ acts on an input $\mX \in \sR^{N \times D}$ via row-wise matrix-vector multiplication,  followed by a row-wise addition i.e.\
$(\mR, \vt) \circ \mX = \mX \mR^T + \ones_N \vt^T$.
The identity element $e$ is $(\eye_D, \zeros_D)$.

\subsection{Euclidean group \texorpdfstring{$\mathit{E}(D)$}{E(D)}}\label{definition:euclidean_group}
\label{appendix:group_definition_euclidean}
The Euclidean group in $D$ dimensions, $\mathit{E}(D)$,  corresponds to the set of all distance-preserving functions in Euclidean space.
It is composed of all pairs of $D$-dimensional orthonormal matrices and translation vectors.
That is, $\mathit{E}(D) = \mathit{O}(D) \times \mathit{T}(D)$.
Note that $\mathit{E}(D) \supset \mathit{SE}(D)$.
Like the roto-translation group, group elements combined via a semidirect product, i.e.\
$(\mA, \vt) \cdot (\mA', \vt') = (\mA \mA', \mA \vt' + \vt)$
and act on an input $\mX \in \sR^{N \times D}$ via row-wise matrix-vector multiplication,  followed by a row-wise addition i.e.\
$(\mA, \vt) \circ \mX = \mX \mA^T + \ones_N \vt^T$.
The identity element $e$ is $(\eye_D, \zeros_D)$.

\subsection{Permutation group \texorpdfstring{$\mathit{S}(N)$}{S(N)}}\label{definition:permutation_group}
\label{appendix:group_definition_permutation}
The permutation group in $N$ dimensions, $\mathit{S}(N)$, is the set of all permutation matrices,
i.e.\ $\mathit{S}(N) = 
\left\{\mP \in \{0,1\}^{N \times N} \mid \mP^T \mP = \eye_N\right\}
$.
Two group elements $\mP, \mP' \in \mathit{S}(N)$ are combined via matrix multiplication, i.e.\ $\mP \cdot \mP' = \mP \mP'$.
A group element $\mP \in \mathit{S}(N)$ acts on an input $\mX \in \sR^{N \times D}$ via matrix multiplication,  i.e.\
$\mP \circ \mX = \mP \mX$.
Note that the group action permutes the rows and not the columns.
The identity element $e$ is the identity matrix $\eye_N$.
\changelocaltocdepth{2}

\clearpage

\section{Proof of Theorem 1}\label{appendix:invariance_preservation}
\invariancepreservation* 
Recall that
\begin{align*}
    &f(\mX) = \mathrm{argmax}_{y \in \sY} p_{\mX,y} \\
    \text{with } &p_{\mX,y} = \Pr_{\mZ \sim \mu_\mX} \left[g(\mZ) = y\right] \\
    \text{and } &\mu_\mX(\mZ) = \prod_{d=1}^D\mathcal{N}\left(\mZ_{:,d} \mid \mX_{:,d}, \sigma^2 \eye_N\right).
\end{align*}
We shall prove~\cref{theorem:invariance_preservation} by showing that $\forall t \in \sT: p_{\mX,y} = p_{(t \circ \mX),y}$, i.e.~the prediction probabilities are invariant.
To do so, we need the following simple lemma:
\begin{restatable}{lemma}{normalchangeofvariables}\label{lemma:normal_change_of_variables}
    Consider an arbitrary invertible matrix $\mA \in \sR^{N \times N}$, vectors $\vz,\vm \in \sR^N$ and scalar $\sigma > 0$. Then
    \begin{align*}
        \left\lvert\det\left(\mA^{-1}\right)\right\rvert \mathcal{N}\left(\mA^{-1}\vz \mid \vm, \sigma^2 \eye_N\right) = 
        \mathcal{N}\left(\vz \mid \mA \vm, \sigma^2 \mA \mA^T\right).
    \end{align*}
\end{restatable}
\begin{proof}
    It follows from the change of variables formula for densitiy functions that the l.h.s.\ term is the density function of random variable $\vz' = \mA \vz$ with $\vz \sim \mathcal{N}\left(\vm, \sigma^2 \ones_N\right)$.
    Due to the behavior of multivariate normal distributions under affine transformation, the r.h.s.\ term is also the density function of $\vz'$.
\end{proof}
We begin by proving~\cref{theorem:invariance_preservation} for $\sT \subseteq \mathit{E}(D)$, before considering $\sT  \subseteq \mathit{S}(N)$.
In the following, let $\hat{g}(\mZ) = \indicator\left[g(\mZ) = y^*\right]$ indicate whether base classifier $g$ classifies input $\mZ$ as $y^*$.

\textbf{Case 1:}
    Assume that $\sT \subseteq \mathit{E}(D)$ and consider an arbitrary $t \in \sT$.
    By definition of the Euclidean group and the associated group action (see~\cref{definition:euclidean_group}) 
    we must have $t = (\mA, \vb)$ and $t \circ \mX = \mX \mA^T + \ones_N \vb^T$
    for some orthogonal matrix $\mA \in \sR^{D \times D}$ and translation vector $\vb \in \sR^{N}$.
    By definition of the smoothing distribution, we have 
    \begin{align*}
        p_{(t \circ \mX),y} 
        & = 
        \int_{\sR^{N \times D}}\hat{g}(\mZ) \mu_{(\mX \mA^T + \ones_N \vb^T)}(\mZ) \ d \mZ
        \\
        & =
        \int_{\sR^{N \times D}}\hat{g}(\mZ) \prod_{n=1}^N \mathcal{N}\left(\mZ_n \mid \mA \mX_n + \vb, \sigma^2\eye_D\right) \ d \mZ
        \\
        & =
        \int_{\sR^{N \times D}}\hat{g}(\mZ) \prod_{n=1}^N \mathcal{N}\left(\mZ_n - \vb \mid \mA \mX_n, \sigma^2\eye_D\right) \ d \mZ,
    \end{align*}
    where in the last equality we have used that $\mathcal{N}(z \mid  m + b) = \mathcal{N}(z -b  \mid m)$.
    We can now perform two substitutions, $\mV = \mZ - \ones_N \vb^T$ and $\mW = \mV (\mA^{-1})^T$, to transform this term into an expectation w.r.t.\ the original smoothing distribution $\mu_{\mX}$:
    \begin{align*}
        & =
        \int_{\sR^{N \times D}}\hat{g}(\mV + \ones_N \vb^T) \prod_{n=1}^N \mathcal{N}\left(\mV_n \mid \mA \mX_n , \sigma^2\eye_D\right) \ d \mV
        \\
        & =
        \int_{\sR^{N \times D}}\hat{g}(\mV + \ones_N \vb^T) \prod_{n=1}^N \mathcal{N}\left(\mV_n \mid  \mA \mX_n , \sigma^2 \mA \mA^T\right) \ d \mV
        \\
        & =
        \int_{\sR^{N \times D}}\hat{g}(\mV + \ones_N \vb^T) \prod_{n=1}^N \mathcal{N}\left(\mA^{-1} \mV_n \mid  \mX_n , \sigma^2 \eye_N\right) \ d \mV
        \\
        & =
        \int_{\sR^{N \times D}}\hat{g}(\mW \mA^T + \ones_N \vb^T) \prod_{n=1}^N \mathcal{N}\left(\mW_n \mid  \mX_n , \sigma^2 \eye_N\right) \ d \mW.
    \end{align*}
    In the second step we have used the fact that $\mA$ is orthogonal, i.e.$\mA \mA^T = \eye_N$.
    In the third step, we have applied~\cref{lemma:normal_change_of_variables}.
    Note that, because $\left\lvert\mathrm{det}(\mA) \right\rvert = 1$, we did not have to change the volume element.
    Finally, we can use the fact that $g$ is invariant w.r.t.\ group $\sT$ to prove that
    \begin{align*}
        p_{(t \circ \mX),y}
        &= 
        \int_{\sR^{N \times D}}\hat{g}(\mA \mW + \ones_N \vb^T) \prod_{n=1}^N \mathcal{N}\left(\mW_n \mid  \mX_n , \sigma^2 \eye_N\right) \ d \mW
        \\
        & = 
        \int_{\sR^{N \times D}}\hat{g}(\mW) \prod_{n=1}^N \mathcal{N}\left(\mW_n \mid  \mX_n , \sigma^2 \eye_N\right) \ d \mW
        \\
        & = 
        p_{\mX,y}.
    \end{align*}

\textbf{Case 2:}
    Assume that $\sT \subseteq \mathit{S}(N)$ and consider an arbitrary $t \in \sT$.
    By definition of the permutation group and the associated group action (see~\cref{definition:permutation_group}) 
    we must have $t = \mP$ and $t \circ \mX = \mP \mX$
    for some orthogonal matrix $\mP \in \{0,1\}^{N \times N}$.
    The proof is virtually identical to that for the Euclidean group, except we now use the substitution $\mV = \mP \mZ$:
     \begin{align*}
        p_{(t \circ \mX),y} 
        & = 
        \int_{\sR^{N \times D}}\hat{g}(\mZ) \mu_{(\mP \mX)}(\mZ) \ d \mZ
        \\
        & =
        \int_{\sR^{N \times D}}\hat{g}(\mZ) \prod_{d=1}^D \mathcal{N}\left(\mZ_{:,d} \mid \mP \mX_{:,d}, \sigma^2\eye_N\right) \ d \mZ
        \\
        & =
        \int_{\sR^{N \times D}}\hat{g}(\mZ) \prod_{d=1}^D \mathcal{N}\left(\mZ_{:,d} \mid \mP \mX_{:,d}, \sigma^2 \mP \mP^T\right) \ d \mZ
        \\
        & =
        \int_{\sR^{N \times D}}\hat{g}(\mZ) \prod_{d=1}^D \mathcal{N}\left(\mP^{-1} \mZ_{:,d} \mid  \mX_{:,d}, \sigma^2 \eye_N\right) \ d \mZ
        \\
        & =
        \int_{\sR^{N \times D}}\hat{g}(\mP \mV) \prod_{d=1}^D \mathcal{N}\left( \mV_{:,d} \mid  \mX_{:,d}, \sigma^2 \eye_N\right) \ d \mV
        \\
        & =
        \int_{\sR^{N \times D}}\hat{g}(\mV) \prod_{d=1}^D \mathcal{N}\left( \mV_{:,d} \mid  \mX_{:,d}, \sigma^2 \eye_N\right) \ d \mV
        \\
        & = p_{\mX,y}.
    \end{align*}

\clearpage

\section{Orbit-based gray-box certificates}\label{appendix:postprocessing}
In the following, we first prove the soundness of our orbit-based approach to gray-box robustness certification.
We then present more explicit characterizations of the orbit-based certificates, which let us determine whether a specific perturbed input is part of the augmented certified region.

\subsection{Proof of Theorem 2}\label{appendix:postprocessing_main_proof}

\postprocessingmain* 
\begin{proof}
    Consider an arbitrary $\mX' \in \tilde{\sB}$.
    Due to the definition of $\tilde{\sB}$, there must be a $\mZ \in \sB$ and a $t \in \sT$ such that $\mX' = t \circ \mZ$.
    Since $t \in \sT$, we know that $f(t \circ \mZ) = f(\mZ)$.
    Since $\mZ \in \sB$, we know that $f(\mZ) = y^*$.
    By the transitive property, we have $f(\mX') = y^*$.
\end{proof}

\subsection{Explicit characterizations}
Recall that, for randomly smoothed models, a prediction $y^* = f(\mX)$ is certifiably robust within a Frobenius norm ball $\sB = \left\{\mZ \mid || \mZ - \mX ||_2 <  r \right\}$ with $r = \sigma \Phi^{-1}\left(p_{\mX,y^*}\right)$.
For this certificate, we can determine whether a specific perturbed input $\mX'$ is in augmented certified region $\tilde{\sB} =  \cup_{\mZ \in \sB} [\mZ]_\sT$ by finding a transformation that minimizes the Frobenius distance between $\mX'$ and $\mX$:

\begin{restatable}{corollary}{preprocessingminimization}\label{theorem:preprocessing_minimization}
    Let $f \in \sR^{N \times D} \rightarrow \sY$ be invariant under  group $\sT$.
    Let $y^* = f(\mX)$ be a prediction that is certifiably robust to a set of perturbed inputs $\sB=\left\{ \mZ \mid ||\mZ - \mX||_2 < r \right\}$, i.e.\ $\forall \mZ \in \sB: f(\mZ) = y^*$.
    If $\min_{t \in \sT} ||(t \circ \mX') - \mX||_2 < r$, then $f(\mX') = y^*$.
\end{restatable}
\begin{proof}
    Let $t^* = \mathrm{argmin}_{t \in \sT} || (t \circ \mX') - \mX||_2$,
    define $\mZ = t^* \circ \mX'$ 
    and assume that $||\mZ - \mX||_2 < r$.
    By definition of $\sB$, we have $\mZ \in \sB$.
    Because $\sT$ is a group, there must be an inverse element $(t^*)^{-1} \in \sT$
    with $(t^*)^{-1} \circ \mZ = \mX'$.
    Thus, by definition of orbits (see~\cref{definition:equivalence_class}), we have $\mX' \in [\mZ]_\sT$.
    It follows from~\cref{theorem:postprocessing_main} that $f(\mX') = y^*$.
\end{proof}
In the following, we discuss how to solve the optimization problem $\min_{\tau \in \sT} ||\tau\left(\mX'\right) - \mX||_2 < r$ for invariance under different groups $\sT$.
Before proceeding, remember that solving this optimization problem is not necessary for certifying robustness, i.e.~specifying a set of inputs $\tilde{\sB}$ such that $\forall \mX' \in \tilde{\sB}: f(\mX') = y^*$ (see~\cref{theorem:postprocessing_main}).
It is only a way of performing membership inference, i.e.\ determining whether $\mX' \in \tilde{\sB}$ for a specific $\mX' \in \sR^{N \times D}$ -- 
just like computing the Frobenius distance between  $\mX'$ and  $\mX$ can be used to determine whether $\mX'$ is part of the original certified region $\sB$.

\subsubsection{Translation invariance}
By definition of the translation group $\mathit{T}(D)$ and the associated group action (see~\cref{appendix:group_definition_translation}), we have $t \circ \mX' = \mX' + \ones_N \vb^T$ for some translation vector $\vb \in \sR^{D}$.
Thus, 
\begin{align*}
    \min_{\tau \in \sT} ||(\tau \circ \mX') - \mX||_2
    = 
    \min_{\vb \in \sR^D} ||(\mX' + \ones_N \vb^T) - \mX||_2
    = 
    ||\mDelta - \ones_N \overline{\mDelta}||_2,
\end{align*}
where $\mDelta = \mX' - \mX$ and $\overline{\mDelta} \in \sR^{1 \times D}$ are the column-wise averages.
The second equality can be shown by computing the gradients w.r.t.\ $\vb$ and setting them to zero.

\subsubsection{Rotation invariance}\label{appendix:preprocessing_rotation}
By definition of the rotation group $\mathit{SO}(D)$ and the associated group action (see~\cref{appendix:group_definition_rotation}), we have $t \circ \mX' = \mX' \mR^T$ for some rotation matrix $\mR$. Thus, 
\begin{align*}
    \min_{\tau \in \sT} ||(\tau \circ \mX') - \mX||_2
    = 
    \min_{\mR \in \mathit{SO}(D)} ||\mX' \mR^T - \mX||_2.
\end{align*}
This is a special case of the orthogonal Procrustes problem, which can be solved via singular value decomposition~\cite{Schonemann1966}.
The optimal rotation matrix $\mR^*$ is given by
$\mR^* = \mV \hat{S} \mU^T$, where $\mU \mS \mV = (\mX')^T \mX$ and
$\hat{S} = \mathrm{diag}\left(1,\dots, 1, \mathrm{sign}\left(\mathrm{det}\left(\mV \mU^T\right)\right)\right)$.

\subsubsection{Simultaneous rotation and reflection invariance}
By definition of the orthogonal group $\mathit{O}(D)$ and the associated group action (see~\cref{appendix:group_definition_orthogonal}), we have $t \circ \mX' = \mX' \mA^T$ for some orthogonal matrix $\mA$. Thus, 
\begin{align*}
    \min_{\tau \in \sT} ||(\tau \circ \mA') - \mX||_2
    = 
    \min_{\mA \in \mathit{O}(D)} ||\mX' \mA^T - \mX||_2.
\end{align*}
This is the orthogonal Procrustes problem, which can again be solved via singular value decomposition~\cite{Schonemann1966}.
The optimal orthogonal matrix $\mA^*$ is given by
$\mA^* = \mV  \mU^T$, where $\mU \mS \mV = (\mX')^T \mX$. Here, accounting for the sign of the determinant is not necessary.

\subsubsection{Simultaneous rotation and translation invariance}
By definition of the special Euclidean group $\mathit{SE}(D)$ and the associated group action (see~\cref{appendix:group_definition_special_euclidean}), we have $t \circ \mX' =  \mX' \mR^T + \ones_N \vb^T$ for some rotation matrix $\mR$ and translation vector $\vb \in \sR^{D}$. Thus, 
\begin{align*}
    \min_{\tau \in \sT} ||(\tau \circ \mX') - \mX||_2
    = 
    \min_{\mR \in \mathit{O}(D), \vc \in \sR^D} ||\mX' \mR^T + \ones_N \vc^T - \mX||_2.
\end{align*}
It can be shown that this is equivalent to centering $\mX'$ and $\mX$ and then solving the orthogonal Procrustes problem~\cite{Kabsch1976}, i.e.~
\begin{align*}
    \min_{\tau \in \sT} ||\tau\left(\mX'\right) - \mX||_2
    = 
    \min_{\mR \in \mathit{SO}(D)} ||\left(\mX' - \ones_N \overline{\mX'}\right) \mR^T - \left(\mX - \ones_N \overline{\mX} \right)||_2.
\end{align*}
As discussed in~\cref{appendix:preprocessing_rotation}, this problem can be solved via singular value decomposition.

\subsubsection{Permutation invariance}
By definition of the permutation group $\mathit{S}(N)$ and the associated group action (see~\cref{appendix:group_definition_permutation}), we have $t \circ \mX' = \mP \mX'$ for some permutation matrix $\mP$. Thus, 
\begin{align*}
    \min_{\tau \in \sT} ||(\tau \circ \mX') - \mX||_2
    = 
    \min_{\mP \in \mathit{S}(N)} ||\mP \mX' - \mX||_2
    = 
    \sqrt{\min_{\mP \in \mathit{S}(N)} ||\mP \mX' - \mX||_2^2}.
\end{align*}
The inner optimization problem is equivalent to finding an optimal matching in a bipartite graph with $2 \cdot N$ nodes  whose cost matrix is given by
$\emC_{n,m} = ||\mX'_n - \mX_m||_2^2$. This problem can be solved in polynomial time, for example via the Hungarian algorithm~\cite{Kuhn1955}.

\subsubsection{Simultaneous permutation, rotation and translation invariance}
By definition of the permutation group $\mathit{S}(N)$ and the special Euclidean group $\mathit{SE}(N)$, we have 
\begin{align*}
    \min_{\tau \in \sT} ||(\tau \circ \mX') - \mX||_2
    = 
    \min_{\mP \in \mathit{S}(N) \mR \in \mathit{O}(D), \vc \in \sR^D} ||\mP \left(\mX' \mR^T + \ones_N \vc^T\right) - \mX||_2.
\end{align*}
Different from the previously discussed problems, the above is a challenging optimization problem known as point cloud registration, which does not have an efficiently computable solution.
Approximate solutions to this problem are being actively studied (for a comprehensive survey, see~\cite{Huang2021}).
Note that, if some upper bound $\hat{m}$ with $\min_{\tau \in \sT} ||(\tau \circ \mX') - \mX||_2 < \hat{m}$ fulfills $\hat{m} < r$, then
$\min_{\tau \in \sT} ||(\tau \circ \mX') - \mX||_2 < r $.
Thus, any approximate solution to the point cloud registration problem can be used for certification.
If $\hat{m} < r$, we provably know that $\mX' \in \tilde{\sB}$ and thus $f(\mX') = y^*$.
However, there may be some $\mX' \in \tilde{\sB}$ with $\hat{m} > r$, which would be incorrectly declared as potential adversarial examples.
Thus, an explicit characterization based on approximate solutions to the point cloud registration problem is sound, but not optimal.

\clearpage

\section{Tight gray-box certificates}\label{appendix:tight_certificates}
In the following, we first prove~\cref{theorem:canonical_cert}, which lets us reduce the invariance-constrained optimization problem from~\cref{section:tight_certificates} to a problem that is only constrained by the classifier's clean prediction probability.
Then, we restate the Neyman-Pearson lemma from statistical testing, which provides an exact solution for the worst-case classifier, given its clean prediction probability.
After that, we use the two lemmeta in proving~\cref{theorem:main}, which summarizes tight certificates for translation, rotation and roto-translation invariance.
Next, we discuss the certificates for each type of invariance in more detail and derive the results presented in~\cref{section:tight_translation,section:rotation_2d_tight,section:tight_rotation_3d,section:tight_translation_rotation} from~\cref{theorem:main}.
We conclude by showing how to use Monte Carlo sampling to obtain narrow probabilistic bounds on the certificates involving rotation invariance  (\cref{appendix:monte_carlo_certification}).

\subsection{Proof of Lemma 1}\label{appendix:tight_equivalence_class_theorem}
Recall from~\cref{section:tight_certificates} that, in order to obtain tight gray-box certificates, we need to find the worst-case invariant classifier. That is, we need to solve $\min_{h \in \sH_\sT} \Pr_{\mZ \sim \mu_{\mX'}} \left[h(\mZ) = y^*\right]$, where $\sH_\sT$ is the set of all classifiers that are at least as likely as base classifier $g$ to predict class $p_{\mX,y^*}$ and have the same invariances.
We want to prove that the invariance constraint can be eliminated via a canonical map $\gamma(\mZ)$, which maps each input $\mZ \in \sR^{N \times D}$ to a distinct representative of its orbit $[\mZ]_\sT$.
\canonicalcert*
For the proof, recall
\definitionequivalenceclasses*
and
\canonicalmap* 
and
\begin{equation*}
    \sH_\sT = \left\{h : \sR^{N \times D} \rightarrow \sY \mid
  \Pr_{\mZ \sim \mu_\mX} \left[h(\mZ) = y\right]
   \geq
  p_{\mX,y^*}
  \land \forall \mZ, \forall \mZ' \in \left[\mZ\right]_\sT : h(\mZ) = h(\mZ')
\right\}.
\end{equation*}

We begin our proof by deriving two lemmeta.

The first lemma states that, if an input $\mZ$ is not the representative of its own orbit, then it cannot be the representative of any orbit.
\begin{lemma}\label{lemma:unique_representative}
    Let $\gamma$ be a canonical map for invariance under $\sT$. Then, for all $\mZ \in \sR^{N \times D}$
    \begin{equation*}
        \mZ \neq \gamma(\mZ) \implies \nexists \mZ'\in \sR^{N \times D} : \gamma(\mZ') = \mZ.
    \end{equation*}
\end{lemma}
\begin{proof}
    Proof by contraposition.
    Assume there was some $\mZ'$ with $\gamma(\mZ') = \mZ$.
    \cref{eq:canonical_map_1} from~\cref{definition:equivalence_class} would imply that $\mZ \in \left[\mZ'\right]_\sT$.
    \cref{eq:canonical_map_2} from~\cref{definition:equivalence_class} would then imply that $\gamma(\mZ') = \gamma(\mZ)$.
    By the transitive property, we would have $\mZ = \gamma(\mZ') = \gamma(\mZ)$.
\end{proof}

The second lemma allows us to replace all equality constraints $\forall \mZ' \in \left[\mZ\right]_\sT : h(\mZ) = h(\mZ')$ involving input $\mZ$ and element of its equivalence class with a single equality constraint $h(\mZ) = h(\gamma(\mZ))$.
\begin{lemma}\label{lemma:simplify_worstcase_set}
    Let $\gamma$ be a canonical map for invariance under $\sT$.
    Then
    \begin{equation*}
        \forall \mZ, \forall \mZ' \in \left[\mZ\right]_\sT : h(\mZ) = h(\mZ') 
        \iff
        \forall \mZ: h(\mZ) = h(\gamma(\mZ)).
    \end{equation*}
\end{lemma}
\begin{proof}
    \item[$\Rightarrow$]
    Consider an arbitrary $\mZ \in \sR^{N \times D}$.
    Due to ~\cref{eq:canonical_map_1} from~\cref{definition:equivalence_class}, we know that $\gamma(\mZ) \in \left[\mZ\right]_\sT$.
    Therefore, $\forall \mZ' \in \left[\mZ\right]_\sT : h(\mZ) = h(\mZ')$ implies $h(\mZ) = h(\gamma(\mZ))$.
    \item[$\Leftarrow$]
    Assume that $\forall \mZ: h(\mZ) = h(\gamma(\mZ))$.
    Consider  an arbitrary pair of   inputs $\mZ \in \sR^{N \times D}, \mZ' \in \left[\mZ\right]_\sT$.
    Due to our assumption, we know that $h(\mZ) = h(\gamma(\mZ))$.
    Due to~\cref{eq:canonical_map_2} from~\cref{definition:canonical_map}, we know that $\gamma(\mZ) = \gamma(\mZ')$ and
    thus $h(\gamma(\mZ)) = h(\gamma(\mZ'))$.
    Again, due to our assumption, we know that  $h(\gamma(\mZ')) = h(\mZ')$.
    By the transitive property, we have $h(\mZ)  = h(\mZ')$.
\end{proof}

We can now apply the two lemmata to prove~\cref{theorem:canonical_cert}.
\cref{lemma:simplify_worstcase_set} allows us to restate $\sH_\sT$ as follows:
\begin{equation*}
    \sH_\sT = \left\{h : \sR^{N \times D} \rightarrow \sY \mid
  \Pr_{\mZ \sim \mu_\mX} \left[h(\mZ) = y\right]
   \geq
  p_{\mX,y^*}
  \land \forall \mZ : h(\mZ) = h(\gamma(\mZ))
\right\}.
\end{equation*}
Thus, the optimization problem $\min_{h \in \sH_\sT} \Pr_{\mZ \sim \mu_{\mX'}} \left[h(\mZ) = y^*\right]$ can be written as
\begin{align*}
    \min_{h : \sR^{N \times D} \rightarrow \sY} & \Pr_{\mZ \sim \mu_{\mX'}} \left[h(\mZ) = y^*\right]
    \\
    \text{ s.t. }& \Pr_{\mZ \sim \mu_{\mX}} \left[h(\mZ) = y^*\right] \geq p_{\mX, y^*}
    \land
     \forall \mZ : h(\mZ) = h(\gamma(\mZ)).
\end{align*}
Since we already know from the second constraint that any feasible solution must fulfill $h(\mZ) = h(\gamma(\mZ))$, we may as well substitute $h(\gamma(\mZ))$ for $h(\mZ)$  before solving the problem:
\begin{align}\label{eq:proof_canonical_cert_1}
    \begin{split}
    \min_{h : \sR^{N \times D} \rightarrow \sY} & \Pr_{\mZ \sim \mu_{\mX'}} \left[h(\gamma(\mZ)) = y^*\right]
    \\
    \text{ s.t. }& \Pr_{\mZ \sim \mu_{\mX}} \left[h(\gamma(\mZ)) = y^*\right] \geq p_{\mX, y^*}
    \land
     \forall \mZ : h(\mZ) = h(\gamma(\mZ)).
    \end{split}
\end{align}
For each $\mZ \in \sR^{N \times D}$, we can now distinguish two cases:
If $\mZ = \gamma(\mZ)$, then the constraint $h(\mZ) = h(\gamma(\mZ))$ is trivially fulfilled and can thus be dropped.
If $\mZ \neq \gamma(\mZ)$, then~\cref{lemma:unique_representative} shows that $h(\mZ)$ does not appear in the objective function or first constraint of~\cref{eq:proof_canonical_cert_1}, because there is no other $\mZ'$ with $\gamma(\mZ') = \mZ$.
We can thus ignore the second constraint, solve the optimization problem
\begin{align*}
    \min_{h : \sR^{N \times D} \rightarrow \sY}  \Pr_{\mZ \sim \mu_{\mX'}} \left[h(\gamma(\mZ)) = y^*\right]
    \text{ s.t. } \Pr_{\mZ \sim \mu_{\mX}} \left[h(\gamma(\mZ)) = y^*\right] \geq p_{\mX, y^*}
\end{align*}
and then let $h(\mZ) \gets h(\gamma(\mZ))$ to obtain an optimal, feasible solution to~\cref{eq:proof_canonical_cert_1}.

\subsection{Neyman-Pearson Lemma}\label{appendix:neyman_pearson}
For our purposes, the Neyman-Pearson lemma~\cite{Neyman1933} can be formulated as follows:
\begin{restatable}[Neyman-Pearson lower bound]{lemma}{neymanpearsonlower}\label{lemma:neyman_pearson_lower}
    Let $\mu_{\mX'}$, $\mu_\mX$, be two continuous distributions over a measurable set $\sA$
    such that, for all $\kappa \in \sR_{+}$, the set $\left\{z \mid \frac{\mu_{\mX'}(z)}{\mu_{\mX}(z)} = \kappa\right\}$ has measure zero.
    Consider an arbitrary label set $\sY$, a specific class label $y \in \sY$ and scalar $p \in [0,1]$.
    Then
    \begin{align*}
        \left(
        \min_{h : \sA \rightarrow \sY}
        \Pr_{z \sim \mu_{\mX'}}\left[h(z) = y \right]
        \textrm{ s.t.\ }
        \Pr_{z \sim \mu_{\mX}}\left[h(z) = y \right] \geq p
        \right) 
        = 
         & \E_{z \sim \mu_{\mX'}}\left[h^*(z)\right] \\
        \text{with } h^*(z) = \indicator\left[\frac{\mu_{\mX'}(z)}{\mu_{\mX}(z)} \leq \kappa\right]
        \text{ and } \kappa \in \sR_+ \text{ such that } & \E_{z \sim \mu_{\mX}}\left[h^*(z)\right] = p.
    \end{align*}
\end{restatable}
Here, indicator function $h^*$ corresponds to a classifier that predicts class $y$ if and only if the likelihood ratio is below a specific threshold $\kappa$.
For an application of this variant of the Neyman-Pearson lemma to black-box robustness certification, as well as a discussion of its relation to most powerful hypothesis tests, see~\cite{Cohen2019}.
For various other formulations of the lemma, see~\cite{Johnson2016}.

One can use the same approach to obtain an upper bound on the probability of predicting a specific class.
This will be relevant for our discussion of multi-class certificates in~\cref{appendix:multiclass}.
\begin{restatable}[Neyman-Pearson upper bound]{lemma}{neymanpearsonupper}\label{lemma:neyman_pearson_upper}
    Let $\mu_{\mX'}$, $\mu_\mX$, be two continuous distributions over a measurable set $\sA$
    such that, for all $\kappa \in \sR_{+}$, the set $\left\{z \mid \frac{\mu_{\mX'}(z)}{\mu_{\mX}(z)} = \kappa\right\}$ has measure zero.
    Consider an arbitrary label set $\sY$, a specific class label $y \in \sY$ and scalar $p \in [0,1]$.
    Then
    \begin{align*}
        \left(
        \max_{h : \sA \rightarrow \sY}
        \Pr_{z \sim \mu_{\mX'}}\left[h(z) = y \right]
        \textrm{ s.t.\ }
        \Pr_{z \sim \mu_{\mX}}\left[h(z) = y \right] \leq p
        \right)
        = 
         & \E_{z \sim \mu_{\mX'}}\left[h^*(z)\right] \\
        \text{with } h^*(z) = \indicator\left[\frac{\mu_{\mX'}(z)}{\mu_{\mX}(z)} \geq \kappa\right]
        \text{ and } \kappa \in \sR_+ \text{ such that } & \E_{z \sim \mu_{\mX}}\left[h^*(z)\right] = p.
    \end{align*}
\end{restatable}

\subsection{Proof of Theorem 3}\label{appendix:tight_methodology}
\maincert*
We begin our proof by applying \cref{theorem:canonical_cert}, which lets us restate the l.h.s.\ optimization problem from~\cref{eq:main_result_objective} as 
\begin{equation}\label{eq:after_lemma_1}
        \min_{h : \sR^{N \times D} \rightarrow \sY}   \Pr_{\mZ \sim \mu_{\mX'}} \left[h(\gamma(\mZ)) = y^*\right]
        \text{s.t.} \Pr_{\mZ \sim \mu_{\mX}} \left[h(\gamma(\mZ)) = y^*\right] \geq p_{\mX, y^*}
\end{equation}
with canonical map $\gamma : \sR^{N \times D} \rightarrow \sR^{N \times D}$.
Note that~\cref{eq:after_lemma_1} has the same form as the optimization problem solved by the Neyman Pearson lemma (\cref{lemma:neyman_pearson_lower}), save for the canonical representation in the probability terms.

Our goal is to 1.) specify  a canonical map $\gamma$ for invariance w.r.t.\ group $\sT$ 2.) bring the two probability terms from~\cref{eq:after_lemma_1} into a form that does not depend $\gamma$, so that we can solve~\cref{eq:after_lemma_1} exactly via the Neyman-Pearson lemma.
To this end, we make the following proposition, which we shall later verify for the different considered groups $\sT$:
\begin{proposition}\label{proposition:reparameterization}
    Let $\sT \in \{\mathit{T}(D), \mathit{SO}(D), \mathit{SE}(D)\}$.
    For $\mathit{SO}(D)$ and $\mathit{SE}(D)$, let $D \in \{2,3\}$.
    Let 
    $e \in \sT$ be the group's identity element
    and 
    $\circ : \sT \times \sR^{N \times D} \rightarrow \sR^{N \times D}$ be the group action (see~\cref{appendix:group_definitions}).
    There exist a parameter space $\Omega \subseteq \sR^{K}$, a function $\tau : \Omega \rightarrow \sT$
    with $\tau(\zeros) = e$,
    as well as a parameter space $\Psi \subseteq \sR^{N \cdot D - K}$ 
    and a function $\xi : \psi \rightarrow \sR^{N \times D}$
    with $K \in \{1,\dots,N \cdot D - 1\}$ 
    such that
    \begin{equation}\label{eq:main_substitution}
        \lambda(\vomega, \vpsi) = \tau(\vomega) \circ \xi(\vpsi).
    \end{equation}
    is a differentiable, surjective function from $\Omega \times \Psi$ to $\sR^{N \times D}$,  injective almost everywhere and fulfills 
    \begin{equation}
    \label{eq:disentangling}
        \forall t \in \sT, \vomega \in \Omega, \vpsi \in \Psi , 
        \exists \vomega' \in \Omega: 
        \quad
        t \circ \lambda(\vomega, \vpsi)  = \lambda(\vomega', \vpsi).
    \end{equation}
\end{proposition}
That is, the input space $\sR^{N \times D}$ can be parameterized by a matrix $\xi(\vpsi) \in \sR^{N \times D}$ that is transformed by a group element $\tau(\vomega) \in \sT$.
As formalized in~\cref{eq:disentangling}, this alternative parameterization lets us neatly disentagle the effect of a group action,
which is a translation and/or rotation of the coordinate system,
from the (group-invariant) geometry of $\mZ$, i.e.\ $\xi(\vpsi)$.

\textbf{Canonical map.} Based on~\cref{proposition:reparameterization}, we can define the following canonical map $\gamma : \sR^{N \times D} \rightarrow \sR^{N \times D}$ for $\mZ = \lambda(\vomega, \vpsi)$:
\begin{equation}\label{eq:main_canonical_map}
    \gamma(\mZ) = (\tau(\vomega))^{-1} \circ \lambda(\vomega, \vpsi).
\end{equation}
Note that $(\tau(\vomega))^{-1} \in \sT$ is the inverse of group element $\tau(\vomega) \in \sT$ and not the inverse of the function $\tau$.
Further note that by~\cref{eq:main_substitution}, it holds for arbitrary $\mZ = \lambda(\vomega, \vpsi)$ that 
\begin{equation}\label{eq:canonical_map_zeros}
    \gamma(\mZ) = (\tau(\vomega))^{-1} \circ \tau(\vomega) \circ \xi(\vpsi) = e \circ \xi(\vpsi)
    = \tau(\zeros) \circ \xi(\vpsi) = \lambda(\mathbf{0}, \vpsi).
\end{equation}
We can verify that $\gamma$ is indeed a valid canonical map by testing the two criteria from~\cref{definition:canonical_map}.
Because $\gamma(\mZ)$ applies a group action, we have $\gamma(\mZ) \in [\mZ]_\sT$, where $[\mZ]_\sT$ is the orbit of $\mZ$ w.r.t.\ $\sT$.
Due to~\cref{eq:disentangling}, we have $\forall \mZ, \forall \mZ \in [\mZ]_\sT: \gamma(\mZ) = 
\gamma(\lambda(\vomega, \vz)) =
\lambda(\zeros, \vpsi) = \gamma(\lambda(\vomega', \vpsi)) =  \gamma(\mZ')$.

\textbf{Substitution. }
Next, we use~\cref{proposition:reparameterization} and the canonical map from~\cref{eq:main_canonical_map} to bring the probabilities from our optimization problem in~\cref{eq:after_lemma_1} into a form that is compatible with the Neyman-Pearson lemma.
To declutter the terms, we first introduce $\hat{h}(\mZ)$ as a shorthand for $\indicator\left[h(\mZ) = y^*\right]$:
\begin{equation*}
    \Pr_{\mZ \sim \mu_\mX}\left[h(\gamma(\mZ)) = y^*\right] = \E_{\mZ \sim \mu_\mX}\left[\hat{h}(\gamma(\mZ))\right].
\end{equation*}
We then perform the substitution $\mZ = \lambda(\vomega, \vpsi)$ with $\vomega \in \Omega \subseteq{\sR^K}$ and
$\vpsi \in \Psi \subseteq \sR^{N \cdot D - K}$:
\begin{align} \label{eq:substitution_start}
     \E_{\mZ \sim \mu_\mX}\left[\hat{h}(\gamma(\mZ))\right]
     = &
     \int_{\sR^{N \times D}} \hat{h}(\gamma(\mZ)) \cdot \mu_\mX(\mZ) \  d \mZ
     \\ \nonumber
     = &
     \int_{\Psi} \int_{\Omega} \hat{h}(\gamma(\lambda(\vomega, \vpsi))) \cdot  \mu_\mX(\lambda(\vomega, \vpsi))
     \cdot 
     \left\lvert \mathrm{det}(\mJ_\lambda(\vomega,\vpsi))\right\rvert
     \ d \vomega d \vpsi.
\end{align}
where $\mJ_\lambda$ is the Jacobian of $\lambda$ after vectorizing its domain and codomain.
Next, we apply the canonical map defined in~\cref{eq:main_canonical_map} to eliminate the group parameters $\vomega$ from the term involving classifier $\hat{h}$. This allows us to marginalize out $\vomega$:
\begin{align}
    \nonumber
    & \int_{\Psi} \int_{\Omega} \hat{h}(\gamma(\lambda(\vomega, \vpsi))) \cdot  \mu_\mX(\lambda(\vomega, \vpsi))
     \cdot 
     \left\lvert \mathrm{det}(\mJ_\lambda(\vomega,\vpsi))\right\rvert
     \ d \vomega d \vpsi \\
    \nonumber
     & = 
     \int_{\Psi} \int_{\Omega} \hat{h}(\lambda(\zeros, \vpsi)) \cdot  \mu_\mX(\lambda(\vomega, \vpsi))
     \cdot 
     \left\lvert \mathrm{det}(\mJ_\lambda(\vomega,\vpsi))\right\rvert
     \ d \vomega d \vpsi \\
     \nonumber
     & = 
     \int_{\Psi}  \hat{h}(\lambda(\zeros, \vpsi)) \int_{\Omega}  \mu_\mX(\lambda(\vomega, \vpsi))
     \cdot 
     \left\lvert \mathrm{det}(\mJ_\lambda(\vomega,\vpsi))\right\rvert
     \ d \vomega d \vpsi
     \\
     \label{eq:substiution_end}
     & \vcentcolon =
     \E_{\vpsi \sim \nu_\mX}\left[\hat{h}(\lambda(\zeros, \vpsi))\right]
     \\
     \nonumber
     &  =
     \Pr_{\vpsi \sim \nu_\mX}\left[h(\lambda(\zeros, \vpsi)) = y^*\right],
\end{align}
where $\nu_\mX(\vpsi) = \int_{\Omega}  \mu_\mX(\lambda(\vomega, \vpsi))
     \cdot 
     \left\lvert \mathrm{det}(\mJ_\lambda(\vomega,\vpsi))\right\rvert
     \ d \vomega $ is the marginal density of parameters $\vpsi$.
Finally, we can insert our result into~\cref{eq:after_lemma_1} to obtain the simplified optimization problem
\begin{equation}\label{eq:after_lemma_1_simplified}
\min_{h : \sR^{N \times D} \rightarrow \sY}   \Pr_{\vpsi \sim \nu_{\mX'}} \left[h(\lambda(\zeros,\vpsi)) = y^*\right]
        \text{s.t.} \Pr_{\vpsi \sim \nu_{\mX}} \left[h(\lambda(\zeros,\vpsi)) = y^*\right] \geq p_{\mX, y^*}.
\end{equation}

\textbf{Applying the Neyman-Pearson lemma.}
\cref{eq:after_lemma_1_simplified} is almost in the form required by the Neyman-Pearson lemma.
But, unlike in~\cref{lemma:neyman_pearson_lower}, we optimize over a function defined on $\sR^{N \times D}$ while only evaluating it on a subset of $\sR^{N \times D}$, namely $\lambda\left(\zeros, \Psi\right)$.
To resolve this mismatch, it is convenient to treat $h$ as a family of variables $(h_\mZ)$ indexed by $\sR^{N \times D}$.
As specified in~\cref{proposition:reparameterization}, $\lambda$ is injective almost everywhere, meaning that each tuple $(\zeros, \vpsi)$ indexes a distinct variable $h_{\lambda(\zeros, \vpsi)}$, save for sets of measure zero, which do not influence the objective and constraint in~\cref{eq:after_lemma_1_simplified}.
Therefore, we may equivalently optimize over a family of variables $(\tilde{h}_\vpsi)$ indexed by $\Psi$:
\begin{equation}\label{eq:after_lemma_1_lowdim}
\min_{\tilde{h} : \Psi \rightarrow \sY}   \Pr_{\vpsi \sim \nu_{\mX'}} \left[\tilde{h}(\vpsi) = y^*\right]
        \text{s.t.} \Pr_{\vpsi \sim \nu_{\mX}} \left[\tilde{h}(\vpsi) = y^*\right] \geq p_{\mX, y^*}.
\end{equation}
According to~\cref{lemma:neyman_pearson_lower}, the minimizer is a function that classifies $\vpsi$ as  $y^*$  iff 
$\tilde{h}^*(\vpsi) = 1$
with 
\begin{equation*}
    \tilde{h}^*(\vpsi) = \indicator\left[\frac{\nu_{\mX'}(\vpsi)}{\nu_{\mX}(\vpsi)} \leq \kappa \right] \\
        \text{ with } \kappa \in \sR_+ \text{ such that }  \E_{\vpsi \sim \nu_{\mX}}\left[\tilde{h}^*(\vpsi)\right] = p_{\mX, y^*}.
\end{equation*}
Consequently, the optimum of our original problem~\cref{eq:after_lemma_1_simplified} is given by any function $h^*(\mZ)$ with $h^*(\lambda(\zeros, \vpsi)) = \tilde{h}(\vpsi)$, i.e.\ we can use an arbitrary classifier for all parts of the domain that do not appear in the probability terms of~\cref{eq:after_lemma_1_simplified}.
We make the following proposition for our choice of worst-case classifier $h^*(\mZ)$:
\begin{proposition}\label{proposition:haar_kernel}
    Consider marginal density $\nu_\mX(\vpsi) = \int_{\Omega}  \mu_\mX(\lambda(\vomega, \vpsi))
     \cdot 
     \left\lvert \mathrm{det}(\mJ_\lambda(\vomega,\vpsi))\right\rvert
     \ d \vomega $.
    Let $\eta$ be a right Haar measure on group $\sT \in \{\mathit{T}(D), \mathit{SO}(D), \mathit{SE}(D)\}$.
    For $\mathit{SO}(D)$ and $\mathit{SE}(D)$, let $D \in \{2,3\}$. 
    Define the group-averaged kernel 
    \begin{equation}
    \beta_{\mX}(\mZ) = \int_{t \in \sT} \exp\left(
        \left\langle t \circ \mZ, \mX \right\rangle_\mathrm{F} \mathbin{/} \sigma^2
        \right) \ d \eta(t)
    \end{equation}
    Then,
    \begin{equation*}
        \frac{\nu_{\mX'}(\vpsi)}{\nu_{\mX}(\vpsi)}
        \propto 
        \frac{\beta_{\mX'}(\lambda(\vzero, \vpsi))}{\beta_{\mX}(\lambda(\vzero, \vpsi))},
    \end{equation*}
    where $\propto$ absorbs factors that are constant in $\vpsi$. 
\end{proposition}
Assuming~\cref{proposition:haar_kernel} holds, we have by transitivity of the previous equalities that
\begin{equation}\label{eq:solution_lowdim_objective}
\min_{h \in \sH_\sT} \Pr_{\mZ \sim \mu_{\mX'}} \left[h(\mZ) = y^*\right]
=
\E_{\vpsi \sim \nu_{\mX'}}\left[h^*(\lambda(\zeros,\vpsi))\right]
\end{equation}
with 
\begin{equation}\label{eq:solution_lowdim_constraint}
        h^*(\mZ) = \indicator\left[\frac{\beta_{\mX'}(\mZ)}{\beta_{\mX}(\mZ)} \leq \kappa\right]
        \text{ and } \kappa \in \sR_+ \text{ such that } \E_{\vpsi \sim \nu_\mX}\left[h^*(\lambda(\zeros,\vpsi))\right] = p_{\mX,y^*}.
\end{equation}

\textbf{Resubstitution.}
Next, we need to transform the expectations w.r.t.\ marginal distribution $\nu_\mX$ over $\Psi \subseteq \sR^{N \cdot D - K}$ back into an expectation w.r.t.\ our original smoothing distribution $\mu_{\mX}$ over $\sR^{N \times D}$.
Applying the steps from~\cref{eq:substitution_start} to~\cref{eq:substiution_end} in reverse order shows that
\begin{equation}\label{eq:resubstitution}
\E_{\vpsi \sim \nu_{\mX}}\left[h^*(\lambda(\zeros,\vpsi))\right] = \E_{\mZ \sim \mu_{\mX}}\left[h^*(\gamma(\mZ))\right].
\end{equation}

\textbf{Exploiting invariance.}
Finally, we use the fact that our worst-case classifier $h^*$ is invariant under group $\sT$ to eliminate the canonical map $\gamma$ from the expectation.
Recall from~\cref{eq:main_canonical_map} that, for any $\mZ = \lambda(\vomega, \vpsi)$, we defined $\gamma(\mZ) = (\tau(\vomega))^{-1} \circ \mZ$, i.e.\ the canonical map lets a group element act on $\mZ$.
Since $\eta$ is a right Haar measure, we have
\begin{align*}
    \beta_\mX((\tau(\vomega))^{-1} \circ \mZ)
    & =
    \int_{t \in \sT} \exp\left(
        \left\langle t \circ (\tau(\vomega))^{-1} \circ \mZ, \mX \right\rangle_\mathrm{F} \mathbin{/} \sigma^2
        \right) \ d \eta(t) \\
    &= 
    \int_{t \in \sT} \exp\left(
        \left\langle (t \cdot (\tau(\vomega))^{-1}) \circ \mZ, \mX \right\rangle_\mathrm{F} \mathbin{/} \sigma^2
        \right) \ d \eta(t)
    \\
    &=
    \int_{t \in \sT} \exp\left(
        \left\langle u \circ \mZ, \mX \right\rangle_\mathrm{F} \mathbin{/} \sigma^2
        \right) \ d \eta(u)
    \\ 
    & = \beta_\mX(\mZ),
\end{align*}
where the second equality follows from the fact that $\circ$ is a group action
The third equality holds because $\eta$ is a right Haar measure, meaning we can make the substitution $u = t \cdot (\tau(\vomega))^{-1})$ without having to change the measure.
Thus, we have 
\begin{align*}
    h^*(\gamma(\mZ)) = \indicator\left[\frac{\beta_{\mX'}(\gamma(\mZ))}{\beta_\mX(\gamma(\mZ))} \leq \kappa \right] = 
    \indicator\left[\frac{\beta_{\mX'}(\mZ)}{\beta_\mX(\mZ)} \leq \kappa\right]
    = h^*(\mZ).
\end{align*}
Combined with~\cref{eq:resubstitution} and \cref{eq:solution_lowdim_objective,eq:solution_lowdim_constraint}, this proves that the optimal value of the original variance-constrained problem is
\begin{equation}
\min_{h \in \sH_\sT} \Pr_{\mZ \sim \mu_{\mX'}} \left[h(\mZ) = y^*\right]
=
\E_{\mZ \sim \mu_{\mX'}} \left[h^*(\mZ)\right]
\end{equation}
with 
\begin{equation*}
        h^*(\mZ) = \indicator\left[\frac{\beta_{\mX'}(\mZ)}{\beta_{\mX}(\mZ)} \leq \kappa\right]
        \text{ and } \kappa \in \sR_+ \text{ such that } \E_{\mZ \sim \mu_{\mX}} \left[h^*(\mZ)\right] = p_{\mX, y^*} = p_{\mX,y^*}.
\end{equation*}
The last thing we need to do in order to conclude our proof is verify that~\cref{proposition:reparameterization,proposition:haar_kernel} hold for each of the considered groups $\sT$.

\subsubsection{Verifying Propositions 1 and 2 for translation invariance}\label{section:verification_translation}
For $\sT=  \mathit{T}(D)$, we define  $\lambda(\vomega, \vpsi) = \tau(\vomega) \circ \xi(\vpsi)$ with $\tau : \sR^{D} \rightarrow \mathit{T}(D)$, $\xi : \sR^{(N-1) D} \rightarrow \sR^{N \times D}$ and
\begin{align*}
    \tau(\vomega) &= \vomega \\
    \xi(\vpsi) &=
    \begin{bmatrix}
        \zeros_D^T \\
        \mathrm{vec}^{-1}(\vpsi)
    \end{bmatrix},
\end{align*}
where $\mathrm{vec}^{-1} : \sR^{(N-1) D} \rightarrow \sR^{(N-1) \times D}$ reshapes an input vector into a matrix.
Due to the definition of group action $\circ$ from~\cref{appendix:group_definition_translation}, we have
\begin{equation*}
    \lambda(\vomega, \vpsi) = \begin{bmatrix}
        \zeros_D^T \\
        \mathrm{vec}^{-1}(\vpsi)
    \end{bmatrix}
    + \ones_N \vomega^T,
\end{equation*}
with all-ones vector $\ones_N \in \sR^{N}$.
In other words: We represent each element of $\sR^{N \times D}$ as a matrix whose first row is zero, translated by a vector $\vomega$.

The function $\lambda$ is evidently a differentiable bijection with inverse
\begin{equation*}
    \lambda^{-1}(\mZ) = \left(\mZ_1, \mathrm{vec}\left(\mZ_{2:} - \ones_D (\mZ_1)^T\right)\right),
\end{equation*}
meaning it is surjective and injective.
Furthermore, we have $\tau(\zeros_D) = \zeros_D$, with $\zeros_D$ being the identity element of $\mathit{T}(D)$.
Lastly, we have for all $\vt \in \mathit{T}(D)$, $\vomega \in \sR^{D}$, $\vpsi \in \sR^{(N-1)D}$ that
\begin{equation*}
    \vt \circ \lambda(\vomega, \vpsi) = \lambda(\vt + \vomega, \vpsi).
\end{equation*}
Thus, all conditions from~\cref{proposition:reparameterization} are fulfilled.

To verify~\cref{proposition:haar_kernel}, we need to show that 
\begin{equation*}
        \frac{\nu_{\mX'}(\vpsi)}{\nu_{\mX}(\vpsi)}
        \propto 
        \frac{\beta_{\mX'}(\lambda(\vzero, \vpsi))}{\beta_{\mX}(\lambda(\vzero, \vpsi))},
\end{equation*}
with marginal density $\nu_\mX(\vpsi) = \int_{\sR^{D}}  \mu_\mX(\lambda(\vomega, \vpsi))\cdot 
     \left\lvert \mathrm{det}(\mJ_\lambda(\vomega,\vpsi))\right\rvert
     \ d \vomega $
and  Haar integral 
$\beta_{\mX}(\mZ) = \int_{t \in \sR^{D}} \exp\left(
        \left\langle t \circ \mZ, \mX \right\rangle_\mathrm{F} \mathbin{/} \sigma^2
        \right) \ d \eta(t)$, where $\eta$ is an arbitrary right Haar measure on translation group $\mathit{T}(D) = \sR^D$.
Firstly, we see that $\left\lvert \mathrm{det}(\mJ_\lambda(\vomega,\vpsi))\right\rvert = 1$, since we are only performing translations.
Thus,
\begin{align*}
    \nu_\mX(\vpsi) 
    & = \int_{\sR^{D}}  \mu_\mX(\lambda(\vomega, \vpsi)) \ d \vomega  
    \\
    & \propto 
    \int_{\sR^D} \prod_{n=1}^N \exp\left(-\frac{1}{2\sigma^2} (\lambda(\vomega, \vpsi)_n - \mX_n)^T \lambda(\vomega, \vpsi)_n - \mX_n\right)  \ d \vomega
    \\
    & \propto
    \int_{\sR^D} \prod_{n=1}^N \exp\left(\frac{1}{\sigma^2} (\lambda(\vomega, \vpsi)_n)^T \lambda(\vomega, \vpsi)_n\right)  \ d \vomega
    \\
    & =
    \int_{\sR^D} \exp\left(
        \left\langle \lambda(\vomega, \vpsi), \mX \right\rangle_\mathrm{F} \mathbin{/} \sigma^2
        \right) \ d \vomega
    \\
    & = 
    \int_{\sR^D} \exp\left(
        \left\langle \vomega \circ \lambda(\zeros_D, \vpsi), \mX \right\rangle_\mathrm{F} \mathbin{/} \sigma^2
        \right) \ d \vomega,
\end{align*}
where $\propto$ absorbs factors that are constant in $\vomega$.
In the above equalities we have first inserted the definition of our isotropic matrix normal smoothing distribution, then removed constant terms, then expressed the product of exponential functions more compactly using the Frobenius inner product and finally used that, by definition, $\lambda(\vomega, \vpsi) = \tau(\vomega) \circ \xi(\vpsi) = \vomega \circ \lambda(\zeros_D, \vpsi)$.
Finally, we note that the Lebesgue measure is translation-invariant, i.e. $\int_{\sR^{D}} h(\vomega + \vc) \ d \vomega = \int_{\sR^{D}} h(\vomega) \ d \vomega$ for arbitrary functions $h$, meaning it is a Haar measure of the translation group.
Since the translation group is a Lie group and thus locally compact,  the Haar measure is unique up to a multiplicative constant~\cite{Cartan1940,Alfsen1963}.
Thus, \cref{proposition:haar_kernel} holds.

\subsubsection{Verifying Propositions 1 and 2 for rotation invariance in 2D}\label{section:verification_rot_2d}
In this section, let
\begin{equation}
    \mR(\omega) = \begin{bmatrix}\cos(\omega) & - \sin(\omega) \\ \sin(\omega) & \cos(\omega) \end{bmatrix}
\end{equation}
be the matrix that rotates counter-clockwise by angle $\omega$. Note that 
$\mathit{SO}(2) = \{\mR(\omega) \mid \omega \in [0, 2\pi]\}$.
To verify our propositions for $\sT=  \mathit{SO}(2)$, we define  $\lambda(\vomega, \vpsi) = \tau(\omega) \circ \xi(\vpsi)$ with $\tau : [0,2\pi] \rightarrow \mathit{SO}(2)$ and $\xi : \sR_+ \times \sR^{2(N-1)} \rightarrow \sR^{N \times 2}$ (note that the first argument is non-negative) with 
\begin{align*}
    \tau(\omega) & = \mR(\omega)
    \\
    \xi(\vpsi) &=
    \begin{bmatrix}
        \psi_1 & 0 \\
       \multicolumn{2}{c}{ \mathrm{vec}^{-1}(\vpsi_{2:})}
    \end{bmatrix},
\end{align*}
where $\mathrm{vec}^{-1} : \sR^{2(N-1)} \rightarrow \sR^{(N-1) \times 2}$ reshapes an input vector into a matrix.
Due to the definition of group action $\circ$ from~\cref{appendix:group_definition_rotation}, we have
\begin{equation*}
    \lambda(\omega, \vpsi) =
        \begin{bmatrix}
        \psi_1 & 0 \\
       \multicolumn{2}{c}{ \mathrm{vec}^{-1}(\vpsi_{2:})}
    \end{bmatrix} \mR(\omega)^T
\end{equation*}
In other words: We represent each element of $\sR^{N \times 2}$ as a matrix whose first row is aligned with the $x$-axis, rotated counter-clockwise by an angle $\omega$.

The function $\lambda$ is evidently surjective. Any $\mZ  \in \sR^{N \times 2}$ can be represented via
\begin{equation*}
    \mZ = \begin{bmatrix}
        ||\mZ||_2 & 0 \\
        \multicolumn{2}{c}{\mZ_{2:} \mR(- \omega^*)^T}
    \end{bmatrix}
    \mR(\omega^*)^T
\end{equation*}
with $\omega^* = \mathrm{arctan2}(Z_{1,2}, Z_{1,1})$.
It is also injective, save for the set $\{\mZ \in \sR^{N \times 2} \mid \mZ_1 = \zeros\}$, which has measure zero.
The first row $\mZ_1$ uniquely defines $\psi_1 = ||\mZ_1||_2$ and $\omega = \mathrm{arctan2}(Z_{1,2}, Z_{1,1})$, because polar coordinates for non-zero vectors are unique.
The parameter vector $\vz_{2:}$ must fulfill
\begin{equation*}
    \mathrm{vec}^{-1}(\vpsi_{2:}) \mR(\omega) = \mZ_{2:},
\end{equation*}
which has a unique solution because rotation matrix $\mR(\omega)$ is invertible.
In addition to surjectivity and injectivity almost everywhere, $\lambda$ is differentiable and we have $\tau(0) = \eye_N$, with identity matrix $\eye_N$ being the identity element of $\mathit{SO}(2)$. Furthermore, we have for all $\mR(\alpha)$, $\vomega$, $\vpsi$ that
\begin{equation*}
    \mR(\alpha) \circ \lambda(\omega, \vpsi) = \lambda(\alpha + \omega, \vpsi).
\end{equation*}
Thus, all conditions from~\cref{proposition:reparameterization} are fulfilled.

To verify~\cref{proposition:haar_kernel}, we need to show that 
\begin{equation*}
        \frac{\nu_{\mX'}(\vpsi)}{\nu_{\mX}(\vpsi)}
        \propto 
        \frac{\beta_{\mX'}(\lambda(\vzero, \vpsi))}{\beta_{\mX}(\lambda(\vzero, \vpsi))},
\end{equation*}
with marginal density $\nu_\mX(\vpsi) = \int_{[0, 2\pi]}  \mu_\mX(\lambda(\vomega, \vpsi))\cdot 
     \left\lvert \mathrm{det}(\mJ_\lambda(\vomega,\vpsi))\right\rvert
     \ d \vomega $
and  Haar integral 
$\beta_{\mX}(\mZ) = \int_{\mR \in \mathit{SO}(2)} \exp\left(
        \left\langle t \circ \mZ, \mX \right\rangle_\mathrm{F} \mathbin{/} \sigma^2
        \right) \ d \eta(t)$.
We begin by calculating the Jacobian 
\begin{align*}
    \mJ_\lambda(\omega,\vpsi) &= 
    \begin{bmatrix}
        \frac{\partial \mathrm{vec}(\lambda)}{\partial \omega }
        &
        \frac{\partial \mathrm{vec}(\lambda)}{\partial \psi_1 }
        &
        \frac{\partial \mathrm{vec}(\lambda)}{\partial \psi_2 }
        &
        \dots
        &
        \frac{\partial \mathrm{vec}(\lambda)}{\partial \psi_{2N - 1} }
    \end{bmatrix}
    \\
    &=
    \begin{bmatrix}
         - \psi_1 \sin(\omega) & \cos(\omega) & \zeros\\
         \psi_1 \cos(\omega) & \sin(\omega) & \zeros \\
         \va & \zeros & \mB
    \end{bmatrix},
\end{align*}
with some vector $\va \in \sR^{2(N-1)}$ and block-diagonal matrix $\mB \in \sR^{2(N-1) \times 2 (N-1)}$ with 
\begin{equation*}
    \mB = \begin{bmatrix}
        \mR(\omega) & & \zeros \\
        & \ddots & \\
        \zeros & & \mR(\omega)
    \end{bmatrix}.
\end{equation*}
Due to the block structure, we have
\begin{equation*}
    \left\lvert \mathrm{det}(\mJ_\lambda(\omega,\vpsi)) \right\lvert
    = \left \lvert 
        \mathrm{det}\left(
        \begin{bmatrix}
            - \psi_1 \sin(\omega) & \cos(\omega) \\
            \psi_1 \cos(\omega) & \sin(\omega)\\
        \end{bmatrix}
        \right)
    \right \rvert 
    \prod_{n=2}^N  \lvert \mathrm{det}(\mR(\omega))\rvert
    = |\psi_1|.
\end{equation*}
Thus, our marginal density is
\begin{align*}
    \nu_\mX(\vpsi) 
    & = \int_{[0,2\pi]} |\psi_1| \cdot \mu_\mX(\lambda(\omega, \vpsi)) \ d \vomega  
    \\
    & \propto 
    \int_{[0,2\pi]} \prod_{n=1}^N \exp\left(-\frac{1}{2\sigma^2} (\lambda(\omega, \vpsi)_n - \mX_n)^T \lambda(\omega, \vpsi)_n - \mX_n\right)  \ d \omega
    \\
    & \propto
    \int_{[0,2\pi]} \prod_{n=1}^N \exp\left(\frac{1}{\sigma^2} (\lambda(\omega, \vpsi)_n)^T \lambda(\omega, \vpsi)_n\right)  \ d \omega
    \\
    & =
    \int_{[0,2\pi]} \exp\left(
        \left\langle \lambda(\omega, \vpsi), \mX \right\rangle_\mathrm{F} \mathbin{/} \sigma^2
        \right) \ d \omega
    \\
    & = 
    \int_{[0,2\pi]} \exp\left(
        \left\langle \mR(\omega) \circ \lambda(\zeros_D, \vpsi), \mX \right\rangle_\mathrm{F} \mathbin{/} \sigma^2
        \right) \ d \omega,
\end{align*}
where $\propto$ absorbs factors that are constant in $\omega$.
Because the composition of two rotations corresponds to a translation of rotation angles, the translation-invariant Lebesgue measure is an invariant measure for group $\mathit{SO}(2)$ (in this angle-based parameterization): 
$\int_{[0,2\pi]} h(\mR(\omega) \cdot \mR(\omega')) \ d \omega = \int_{[0,2\pi]} h(\mR(\omega + \omega')) \ d \omega =  \int_{[0,2\pi]} h(\mR(\omega)) \ d \omega$ for arbitrary functions $h$. Since $\mathit{SO}(2)$ is a Lie group and thus locally compact,  the Haar measure is unique up to a multiplicative constant~\cite{Cartan1940,Alfsen1963}.
Thus, \cref{proposition:haar_kernel} holds.

\subsubsection{Verifying Propositions 1 and 2 for rotation invariance in 3D}\label{section:verification_rot_3d}
In this section, let
\begin{equation*}
    \mR(\vomega) = 
    \begin{bmatrix}
        \cos(\omega_1) & -\sin(\omega_1) & 0\\
        \sin(\omega_1) & \cos(\omega_1) & 0\\
        0 & 0 & 1
    \end{bmatrix}
    \cdot
    \begin{bmatrix}
        \cos(\omega_2) & 0 & \sin(\omega_2) \\
        0 & 1 & 0 \\
        - \sin(\omega_2) & 0 & \cos(\omega_2)
    \end{bmatrix}
    \cdot
    \begin{bmatrix}
        1 & 0 & 0\\
        0 & \cos(\omega_3) & -\sin(\omega_3)\\
        0 & \sin(\omega_3) & \cos(\omega_3)
    \end{bmatrix}
\end{equation*}
be the matrix that performs an \textit{intrinsic} rotation around the $z$-axis by angle $\omega_1$, followed by a rotation around the new $y$-axis by angle $\omega_2$ and then a rotation around the new $x$-axis by angle $\omega_3$.

Note that $\mathit{SO}(3) = \{\mR(\vomega) \mid \vomega \in \Omega\}$ with $\Omega = [0, 2\pi] \times [-\frac{\pi}{2}, \frac{\pi}{2}] \times [0, 2\pi]$.
To verify our propositions for $\sT=  \mathit{SO}(3)$, we define  $\lambda(\vomega, \vpsi) = \tau(\omega) \circ \xi(\vpsi)$ with $\tau : \Omega  \rightarrow \mathit{SO}(3)$ and $\xi : \sR_+ \times \sR \times \sR_+ \times \sR^{3(N-2)} \rightarrow \sR^{N \times 3}$ (note that the first and third argument are non-negative) with 
\begin{align*}
    \tau(\vomega) & = \mR(\vomega)
    \\
    \xi(\vpsi) &=
    \begin{bmatrix}
        \psi_1 & 0 & 0\\
        \psi_2 & \psi_3 & 0 \\
       \multicolumn{3}{c}{ \mathrm{vec}^{-1}(\vpsi_{4:})}
    \end{bmatrix},
\end{align*}
where $\mathrm{vec}^{-1} : \sR^{3(N-1)} \rightarrow \sR^{(N-1) \times 3}$ reshapes an input vector into a matrix.
Due to the definition of group action $\circ$ from~\cref{appendix:group_definition_rotation}, we have
\begin{equation*}
    \lambda(\omega, \vpsi) =
        \begin{bmatrix}
        \psi_1 & 0 & 0\\
        \psi_2 & \psi_3 & 0 \\
       \multicolumn{3}{c}{ \mathrm{vec}^{-1}(\vpsi_{4:})}
    \end{bmatrix} \mR(\vomega)^T
\end{equation*}
In other words: We represent each element of $\sR^{N \times 3}$ as a matrix whose first row is aligned with the $x$-axis,
and whose second row is in the first or second quadrant of the $x$-$y$-plane (because $\psi_3$ is non-negative) 
which is then intrinsically rotated by $z$-$y$-$x$ angles $\omega_1, \omega_2, \omega_3$.

The function $\lambda(\vomega, \vpsi)$ is injective and surjective, save for the set $\{\mZ \in \sR^{N \times 3} \mid \mZ_1 = \zeros \lor \exists c \in \sR : \mZ_2 = c \cdot \mZ_1\}$, which has measure zero.
That is, for any $\mZ$ outside this set, $\mZ = \lambda(\vomega, \vpsi)$ has a unique solution.
Firstly, $(\psi_1, \omega_1, \omega_2)$ are spherical coordinates of $\mZ_1$, which are unique for $\mZ_1 \neq \zeros$.
Secondly, the unique angles $\omega_1$ and $\omega_2$ constrain the $x$-axis after the intrinsic rotation to be co-linear with $\mZ_1$.
Thus, and because we assume that $\mZ_1$ and $\mZ_2$ are not co-linear, there must be unique angle $\omega_3$ for rotation around the $x$-axis that ensures that $\mZ_2$ is in the first or second quadrant of the new $x$-$y$-plane.
Finally, with the rotation angles $\vomega$ and parameter $\psi_1$ fixed, the remaining parameter values are determined by
\begin{equation*}
        \begin{bmatrix}
        \psi_2 & \psi_3 & 0 \\
       \multicolumn{3}{c}{ \mathrm{vec}^{-1}(\vpsi_{4:})}
    \end{bmatrix} \mR(\vomega)^T = \mZ_{2:}
    \iff
        \begin{bmatrix}
        \psi_2 & \psi_3 & 0 \\
       \multicolumn{3}{c}{ \mathrm{vec}^{-1}(\vpsi_{4:})}
    \end{bmatrix} = \mZ_{2:} (\mR(\vomega)^{-1})^T.
\end{equation*}
The function is also surjective on the entirety of $\sR^{N \times 3}$.
If $\mZ_1 = \zeros$, we have $\psi_1 = 0$ and $\omega_1, \omega_2$ can be chosen arbitrarily.
If $\exists c : \mZ_2 = c \cdot \mZ_1$, then $\omega_3$ can be chosen arbitrarily.
The remaining parameters can be chosen using the procedure described above.
Finally, the function is differentiable, we have $\tau(\zeros) = \eye_N$ and $\tau$ is a surjective function into $\mathit{SO}(3)$, meaning 
\begin{equation*}
    \tau(\vomega') \circ \lambda(\vomega, \vpsi) = (\mR(\vomega') \cdot \mR(\vomega)) \circ \xi(\vv) = \lambda(\vomega'', \vpsi) 
\end{equation*}
for some $\omega'' \in \Omega$. Thus, all criteria from~\cref{proposition:reparameterization} are fulfilled.

Like in previous sections, we verify~\cref{proposition:haar_kernel} by first calculating the Jacobian of $\lambda$.
In the following, we use the shorthands $s_i = \sin(\omega_i)$ and  $c_i = \cos(\omega_i)$. We have 
\begin{align*}
    \mJ_\lambda(\omega,\vpsi) &= 
    \begin{bmatrix}
        \frac{\partial \mathrm{vec}(\lambda)}{\partial \psi_1 }
        &
        \frac{\partial \mathrm{vec}(\lambda)}{\partial \omega_1 }
        &
        \frac{\partial \mathrm{vec}(\lambda)}{\partial \omega_2 }
        &
        \frac{\partial \mathrm{vec}(\lambda)}{\partial \omega_3 }
        &
        \frac{\partial \mathrm{vec}(\lambda)}{\partial \psi_2 }
        &
        \dots
        &
        \frac{\partial \mathrm{vec}(\lambda)}{\partial \psi_{(3 (N-1))} }
    \end{bmatrix}
    \\
    &=
    \begin{bmatrix}
         \mA & \zeros & \zeros\\
         \mD & \mB & \zeros\\
         \mE & \mF & \mC
         \end{bmatrix},
\end{align*}
with 
\begin{align*}
    \mA = & \begin{bmatrix}c_1 c_2 & - \psi_1  c_2 s_1 & -\psi_1 c_1 s_2 \\
        c_2 s_1 & \psi_1 c_1 c_2 &  -\psi_1 s_1 s_2\\
        - s_2 & 0 & -\psi_1 c_2
    \end{bmatrix},
    \\
    \mB = & \begin{bmatrix}
        \psi_3 \left(c_1 c_\gamma s_2 + s_1 s_\gamma\right) & c_1 c_2 & c_1 s_2 s_\gamma - c_\gamma s_1\\
        \psi_3 \left(- c_1 s_\gamma + c_\gamma s_1 s_2 \right) & c_2 s_1 & c_1 c_\gamma + s_1 s_2 s_\gamma \\
        \psi_3 \left(c_2 c_\gamma\right) & - s_2  & c_2 s_\gamma,
    \end{bmatrix}
    \\
    \mC = & \begin{bmatrix}
        \mR(\vomega) & & \zeros \\
        & \ddots & \\
        \zeros & & \mR(\vomega)
    \end{bmatrix}.
\end{align*}
Due to the block structure and because $\omega_2 \in [-\frac{\pi}{2}, \frac{\pi}{2}]$ and $\vpsi_3 \in \sR_+$ 
, we have
\begin{align*}
\left\lvert \mathrm{det}\left(\mJ_\lambda(\omega,\vpsi)\right) \right\rvert
    & =
\mathrm{det}\left(\mA\right) \cdot
\mathrm{det}\left(\mB\right) \cdot
\mathrm{det}\left(\mC\right)
\\
& = \lvert \psi_1^2 \cos(\omega_2)\rvert \cdot \lvert \psi_3 \rvert   \cdot \lvert1 \rvert \\
& = \psi_1^2  \cdot \psi_3  \cdot \cos(\omega_2).
\end{align*}
Thus, our marginal density $\nu_\mX(\vpsi)$ is
\begin{align*}
    \nu_\mX(\vpsi) 
    & = \int_\Omega \psi_1^2  \cdot \psi_3  \cdot \cos(\omega_2) \cdot \mu_\mX(\lambda(\vomega, \vpsi)) \ d \vomega  
    \\
    & \propto 
    \int_\Omega \cos(\omega_2) \cdot \prod_{n=1}^N   \exp\left(-\frac{1}{2\sigma^2} (\lambda(\vomega, \vpsi)_n - \mX_n)^T \lambda(\vomega, \vpsi)_n - \mX_n\right)  \ d \vomega
    \\
    & \propto
    \int_\Omega \cos(\omega_2) \cdot \prod_{n=1}^N \exp\left(\frac{1}{\sigma^2} (\lambda(\vomega, \vpsi)_n)^T \lambda(\vomega, \vpsi)_n\right)  \ d \vomega
    \\
    & =
    \int_\Omega \cos(\omega_2) \cdot \exp\left(
        \left\langle \lambda(\vomega, \vpsi), \mX \right\rangle_\mathrm{F} \mathbin{/} \sigma^2
        \right) \ d \vomega
    \\
    & = 
    \int_\Omega \cos(\omega_2) \cdot \exp\left(
        \left\langle \mR(\vomega) \circ \lambda(\zeros_D, \vpsi), \mX \right\rangle_\mathrm{F} \mathbin{/} \sigma^2
        \right) \ d \vomega.
\end{align*}
This is, up to a multiplicative constant, the unique Haar integral for this angle-based parameterization of $\mathit{SO}(D)$ (see, for instance~\cite[Chapter~1]{Naimark1964}).\footnote{Note that they have a factor $\sin$ instead of $\cos$, because they parameterize $\mathit{SO}(D)$ via $z$-$x$-$z$ Euler angles. The proof for intrinsic $z$-$y$-$x$ rotation is analogous.}
Thus,~\cref{proposition:haar_kernel} holds.

\subsubsection{Verifying Propositions 1 and 2 for roto-translation invariance in 2D and 3D}\label{appendix:verification_rot_trans}
Finally, we prove that the propositions hold for $\sT = \mathit{SE}(D)$ with $D \in \{2,3\}$,
which amounts to combining the results from the previous sections.
In the following, let $\tau_\mathrm{rot} : \Omega_\mathrm{rot} \rightarrow \mathit{SO}(D)$
with $\tau_\mathrm{rot}(\vomega) = \mR(\vomega)$ 
be the the parameterization of $\mathit{SO}(D)$ defined in~\cref{section:verification_rot_2d} or~\cref{section:verification_rot_3d}.
Further let $\xi_\mathrm{rot} : \Psi_\mathrm{rot} \rightarrow \sR^{(N-1) \times D}$ be the same function as in~\cref{section:verification_rot_2d} or~\cref{section:verification_rot_3d}, but for matrices with $N-1$ instead of $N$ rows.

We begin by defining $\tau : \Phi_\mathrm{rot} \times \sR^{D} : \rightarrow \mathit{SE}(D)$ and
$\xi : \Psi_\mathrm{rot} \rightarrow \sR^{N \times D}$ as follows:
\begin{align*}
    \tau(\vomega, \vb) & = (\tau_\mathrm{rot}(\vomega), \vb) = (\mR(\vomega), \vb)
    \\
    \xi(\vpsi) &=
    \begin{bmatrix}
        \zeros_D^T \\
        \xi(\vpsi)
    \end{bmatrix},
\end{align*}
Due to the definition of group action $\circ$ from~\cref{appendix:group_definition_special_euclidean}, we have
\begin{equation*}
    \lambda((\vomega, \vb), \vpsi) = \tau(\vomega, \vb) \circ \xi(\vpsi) = 
        \begin{bmatrix}
        \zeros_D^T \\
        \xi(\vpsi)
    \end{bmatrix}\mR(\vomega)^T + \ones_N \vb^T.
\end{equation*}
In other words: We represent each element of $\sR^{N \times D}$ as a matrix whose first row is zero and whose second row is aligned with the $x$-axis which is then rotated and finally translated.
For $D = 3$ we additionally constrain the third row to be in the first or second quadrant of the $x$-$y$ plane (for more details, see~\cref{section:verification_rot_3d}).

The function $\lambda((\vomega,\vb), \vpsi)$ is surjective and injective, save for
\begin{itemize}
    \item $\{\mZ \in \sR^{N \times 2} \mid \mZ_2 - \mZ_1 = 0 \}$ (if $D = 2$),
    \item $\{\mZ \in \sR^{N \times 3} \mid \mZ_2 - \mZ_1 = 0 \lor \exists c \in \sR :  \mZ_2 - \mZ_1 = c (\mZ_3 - \mZ_1) \}$ (if $D = 3$),
\end{itemize}
which have measure zero.
That is, given an $\mZ \in \sR^{N \times D}$ the solution to  $\mZ = \lambda((\vomega, \vb), \vpsi)$ is unique.
Evidently, we must have $\vb = \mZ_1$.
With this parameter fixed, the remaining parameters can be found by solving the equation 
$\mZ_{2:} - \ones_{N-1} \mZ_1^T = \tau_\mathrm{rot}(\vomega) \circ \xi_\mathrm{rot}(\vpsi)$, whose unique solution we discussed in~\cref{section:verification_rot_2d} and~\cref{section:verification_rot_3d}.
The function is also surjective on the entirety of $\sR^{N \times D}$ for the reasons presented in~\cref{section:verification_rot_2d} and~\cref{section:verification_rot_3d}.
Further note that the function  $\lambda((\vomega,\vb),\vpsi)$ is differentiable and that 
$\tau(\zeros,\zeros) = (\eye_D, \zeros_D)$, which is the identity element of $\mathit{SE}(D)$ (see~\cref{appendix:group_definition_special_euclidean}). Finally, we see that 
\begin{align*}
    (\mR(\vomega'),\vb') \circ \lambda((\vomega, \vb),\vpsi) & = 
    ((\mR(\vomega'),\vb') \cdot (\mR(\vomega), \vb) \circ \xi(\vpsi)
    \\
    & =
    (\mR(\vomega')\mR(\vomega), \mR(\vomega) \vb + \vb') \circ \xi(\vpsi)
    \\
    & = 
    \lambda((\vomega'', \vb''),\vpsi)
\end{align*}
for some $\vomega'' \in \Omega, \vb'' \in \sR^{D}$.
Thus, all criteria from~\cref{proposition:reparameterization} are fulfilled.

Next, we verify~\cref{proposition:haar_kernel} by again showing that the marginal density ${\nu_\mX(\vpsi)}$ is a Haar integral.
The function $\lambda$ is a composition of two functions:
The function $\lambda_\mathrm{rot}$ from the previous sections 
and a translation.
Thus, the Jacobian determinant is the product of the two corresponding Jacobians. As discussed in~\cref{section:verification_translation}, the Jacobian determinant for translation is $1$.
Let $\mJ_{\lambda_\mathrm{rot}}(\vomega,\vpsi)$ be the Jacobian of $\lambda_\mathrm{rot}$.
Then
\begin{align*}
\nu_\mX(\vpsi) 
    & = \int_{\Omega_\mathrm{rot}} 
    \left\lvert \mathrm{det}\left(\mJ_{\lambda_\mathrm{rot}}(\vomega,\vpsi)\right)\right\rvert \cdot  \int_{\sR^{D}}\mu_\mX(\lambda((\vomega,\vb), \vpsi)) \ d \vb  \ d \vomega  
    \\
    & \propto
    \int_{\Omega_\mathrm{rot}}  \left\lvert \mathrm{det}\left(\mJ_{\lambda_\mathrm{rot}}(\vomega,\vpsi)\right)\right\rvert \cdot \int_{\sR^{D}} \exp\left(
        \left\langle \lambda((\vomega,\vb), \vpsi), \mX \right\rangle_\mathrm{F} \mathbin{/} \sigma^2
        \right) \ d \vb d \ \vomega  
    \\
    & = 
    \int_{\Omega_\mathrm{rot}}  \left\lvert \mathrm{det}\left(\mJ_{\lambda_\mathrm{rot}}(\vomega,\vpsi)\right)\right\rvert \cdot  \int_{\sR^{D}} \exp\left(
        \left\langle (\mR(\vomega), \vb) \circ \lambda((\zeros_D, \zeros_D), \vpsi), \mX \right\rangle_\mathrm{F} \mathbin{/} \sigma^2
        \right) \ d \vb \ d \vomega  ,
\end{align*}
where $\propto$ absorbs factors that are constant in $\vomega$ and $\vb$.
This is a Haar integral, because the inner and outer integral are Haar integrals for $\mathit{T}(D)$ and $\mathit{SO}(D)$, respectively.
Let us verify this by considering an arbitrary $(\mR(\vomega'), \vb') \in \mathit{SE}(D)$.
To avoid clutter, define the shorthand $f(\mZ) = \exp\left(\left\langle\mZ, \mX \right\rangle_\mathrm{F} \mathbin{/} \sigma^2 \right)$.
We have 
\begin{align*}
   & \int_{\Omega_\mathrm{rot}}  \left\lvert \mathrm{det}\left(\mJ_{\lambda_\mathrm{rot}}(\vomega,\vpsi)\right)\right\rvert \cdot  \int_{\sR^{D}} f\left(
         ((\mR(\vomega), \vb) \cdot (\mR(\vomega'), \vb')) \circ \lambda((\zeros_D, zeros_D), \vpsi)
        \right) \ d \vb \ d \vomega
    \\
    = &
    \int_{\Omega_\mathrm{rot}}  \left\lvert \mathrm{det}\left(\mJ_{\lambda_\mathrm{rot}}(\vomega,\vpsi)\right)\right\rvert \cdot \int_{\sR^{D}} f\left(
        \lambda(\zeros_D, \vpsi) (\mR(\omega)\mR(\vomega')) + \ones_N \left( \mR(\omega) \vb' + \vb\right)^T
        \right) \ d \vb \ d \vomega
    \\
    = &
    \int_{\Omega_\mathrm{rot}}  \left\lvert \mathrm{det}\left(\mJ_{\lambda_\mathrm{rot}}(\vomega,\vpsi)\right)\right\rvert \cdot \int_{\sR^{D}} f\left(
        \lambda(\zeros_D, \vpsi) (\mR(\omega)\mR(\vomega')) + \ones_N \vb^T
        \right) \ d \vb \ d \vomega
    \\
    = &
    \int_{\Omega_\mathrm{rot}}  \left\lvert \mathrm{det}\left(\mJ_{\lambda_\mathrm{rot}}(\vomega,\vpsi)\right)\right\rvert \cdot \int_{\sR^{D}} f\left(
        \lambda(\zeros_D, \vpsi) \mR(\omega) + \ones_N \vb^T
        \right) \ d \vb \ d \vomega
    \\
    = & \int_{\Omega_\mathrm{rot}}  \left\lvert \mathrm{det}\left(\mJ_{\lambda_\mathrm{rot}}(\vomega,\vpsi)\right)\right\rvert \cdot  \int_{\sR^{D}} f\left(
        (\mR(\vomega), \vb) \circ \lambda((\zeros_D, \zeros_D), \vpsi)
        \right) \ d \vb \ d \vomega.
\end{align*}
Here, we have first applied the definition of the group action and group operator from~\cref{appendix:group_definition_special_euclidean} and then used the fact that we are integrating over Haar measures for $\mathit{T}(D)$ and then $\mathit{SO}(D)$.
Since $\mathit{SE}(D)$ is a Lie group and thus locally compact,  the Haar measure is unique up to a multiplicative constant~\cite{Cartan1940,Alfsen1963}.
This confirms that~\cref{proposition:haar_kernel} holds.

\clearpage

\changelocaltocdepth{3}
\subsection{Group-specific certificates}
In this section, we prove the results for specific invariances from~\cref{section:tight_translation,section:rotation_2d_tight,section:tight_rotation_3d,section:tight_translation_rotation}, i.e.\ translation invariance (\cref{appendix:tight_translation}), rotation invariance in 2D  (\cref{appendix:tight_rotation_2d} and 3D (\cref{appendix:tight_rotation_3d}, as well as roto-translation invariance in 2D and 3D (\cref{appendix:tight_roto_translation}).

\subsubsection{Translation invariance}\label{appendix:tight_translation}
\translationcert* 
During our proof of~\cref{theorem:main} in~\cref{appendix:tight_methodology,section:verification_translation}, we have shown that
\begin{equation}\label{eq:translation_intro_1}
\min_{h \in \sH_\sT} \Pr_{\mZ \sim \mu_{\mX'}} \left[h^*(\mZ) = y^*\right]
=
\E_{\vpsi \sim \nu_{\mX'}}\left[\tilde{h}^*(\vpsi)\right]
\end{equation}
with 
\begin{equation}\label{eq:translation_intro_2}
        \tilde{h}^*(\vpsi) = \indicator\left[\frac{\nu_{\mX'}(\vpsi)}{\nu_{\mX}(\vpsi)} \leq \kappa\right]
        \text{ and } \kappa \in \sR_+ \text{ such that } \E_{\vpsi \sim \nu_\mX}\left[\tilde{h}^*(\vpsi)\right] = p_{\mX,y^*}
\end{equation}
with $\vpsi \in \sR^{(N-1) \cdot D}$ and marginal distribution
    \begin{align*}
    \nu_\mX(\vpsi) & = 
    \int_{\sR^{D}}
    \mu_\mX \left(
    \begin{bmatrix}
        \zeros_D^T \\
        \mathrm{vec}^{-1}(\vpsi)
    \end{bmatrix}
    + \ones_N \vomega^T
    \right)
    \ d \vomega
    \\ &=
    \int_{\sR^{D}}
    \prod_{d=1}^D
    \mathcal{N} \left(
    \begin{bmatrix}
        0 \\
        \mathrm{vec}^{-1}(\vpsi)_{:, d}
    \end{bmatrix}
    + \ones_N \vomega_d
    \mid
    \mX_{:,d}, \sigma^2 \eye_N
    \right)
    \ d \vomega,
\end{align*}
where we have inserted the definition of our isotropic matrix normal smoothing distribution for the second equality.
Using the matrix
$\mA = \begin{bmatrix}
        1 & \zeros_{N-1}^T\\
        -\ones_{N-1} & \eye_{N-1}
\end{bmatrix}
 \in \sR^{N \times N}$
with inverse
$\mA^{-1} = \begin{bmatrix}
    1 & \zeros_{N-1}^T \\
    \ones_{N-1} & \eye_{N-1}
\end{bmatrix}$ 
and all-zeros vector $\zeros_{N-1} \in \sR^{N-1}$, all-ones vector $\ones_{N-1} \in \sR^{N-1}$ and identity matrix $\eye_{N-1}$,
the marginal density can equivalently be written as
\begin{align*}
    \nu_\mX(\vpsi) & = 
    \int_{\sR^{D}}
    \prod_{d=1}^D
    \mathcal{N} \left(
    \mA^{-1}
    \begin{bmatrix}
        \omega_d \\
         \mathrm{vec}^{-1}(\vpsi)_{:, d}
    \end{bmatrix}
    \mid
    \mX_{:,d}, \sigma^2 \eye_N
    \right)
    \ d \vomega,
    \\
    & =
    \int_{\sR^{D}}
    \prod_{d=1}^D
    \frac{1}{|\mathrm{det}(\mA)}
    \mathcal{N} \left(
    \begin{bmatrix}
        \omega_d \\
         \mathrm{vec}^{-1}(\vpsi)_{:, d}
    \end{bmatrix}
    \mid
    \mA
    \mX_{:,d}, \sigma^2 \mA \mA^T
    \right)
    \ d \vomega,
    \\
    & =
    \int_{\sR^{D}}
    \prod_{d=1}^D
    \mathcal{N} \left(
    \begin{bmatrix}
        \omega_d \\
         \mathrm{vec}^{-1}(\vpsi)_{:, d}
    \end{bmatrix}
    \mid
    \mA
    \mX_{:,d}, \sigma^2 \mA \mA^T
    \right)
    \ d \vomega,
\end{align*}
where the second equality follows from the change of variable formula for densities (see also~\cref{lemma:normal_change_of_variables}) 
and the third equality is due to $\mathrm{det}(\mA) = 1$. 
Evidently, we are marginalizing out the first dimension of each of the $D$ densities.
For normal distributions, this is equivalent to dropping the first row of the mean  as well as the first row and column of the covariance matrix, i.e.\
\begin{align*}
    \nu_\mX(\vpsi) &= 
    \prod_{d=1}^D
    \mathcal{N} \left(
         \mathrm{vec}^{-1}(\vpsi)_{:, d}
    \mid
    \mA_{2:}
    \mX_{:,d}, \sigma^2 \mA_{2:} (\mA_{2:})^T
    \right)
    \\
    & =
    \mathcal{N}\left(
        \vpsi \mid \vm_\mX, \mSigma
    \right)
\end{align*}
with \begin{align*}
        \vm_\mX &= \mathrm{vec}\left(\mA_{2:} \mX\right),\\
        \mSigma &= \begin{bmatrix}
        \mB & \zeros & \hdots & \zeros \\
        \zeros & \mB & \hdots & \zeros \\
        \vdots & \vdots & \ddots & \zeros \\
        \zeros & \zeros & \hdots & \mB
    \end{bmatrix}
    \end{align*}
and $\mB = \sigma^2 \mA_{2:} \left(\mA_{2:}\right)^T$.

Inserting back into~\cref{eq:translation_intro_1,eq:translation_intro_2}, we see that the optimal value of our variance-constrained optimization problem $\min_{h \in \sH_\sT} \Pr_{\mZ \sim \mu_{\mX'}} \left[h^*(\mZ) = y^*\right]$ equals the optimal value of the black-box certification problem for anisotropic normal smoothing distribution $\mathcal{N}\left( \vm_\mX, \mSigma \right)$. This optimal value has been derived in prior work (see~\cite{Eiras2021} and Appendix A of~\cite{Fischer2020}) and is
    \begin{align*}
        \Phi\left(\Phi^{-1}\left(p_{\mX, y^*}\right) -
        \sqrt{
        \left(\vm_{\mX'} - \vm_{\mX}\right)^T \mSigma^{-1} \left(\vm_{\mX'} - \vm_{\mX}\right)
        }\right)
    \end{align*}
    We can conclude our proof by showing that
    \begin{align}
        \sqrt{
        \left(\vm_{\mX'} - \vm_{\mX}\right)^T \mSigma^{-1} \left(\vm_{\mX'} - \vm_{\mX}\right)
        }
        = \frac{1}{\sigma} \left\lvert\left\lvert \mDelta - \ones_N \overline{\mDelta}\right\rvert\right\rvert_2
    \end{align}

    To this end, we need the following result:
    \begin{lemma}\label{lemma:inner_product_mean}
    Consider an arbitrary vector $\vx \in \sR^D$ and let $\overline{\vx} = \frac{1}{D}\sum_{d=1}^D \evx_d$ be its average.
    Then
    \begin{align*}
        \vx^T \left(\vx - \ones_D \overline{\vx}\right)
        = 
        \left(\vx - \ones_D \overline{\vx}\right)^T \left(\vx - \ones_D \overline{\vx}\right).
    \end{align*}
\end{lemma}
\begin{proof}
    We substract the left-hand side from the right-hand side
    \begin{align*}
        &\vx^T \left(\vx - \ones_D \overline{\vx}\right)
        = 
        \left(\vx - \ones_D \overline{\vx}\right)^T \left(\vx - \ones_D \overline{\vx}\right) \\
        \iff
        &
        0 = \left(- \ones_D \overline{\vx}\right)^T \left(\vx - \ones_D \overline{\vx}\right)\\
        \iff
        &
        0 = \overline{\vx}\sum_{d=1}^D \overline{\vx} - \overline{\vx}\sum_{d=1}^D \evx_d  \\
        \iff
        &
        0 = \overline{\vx}D \overline{\vx} - \overline{\vx} D \overline{\vx},
    \end{align*}
    where the last equality follows from the fact that $\sum_{d=1}^D \evx_d = D \overline{\vx}$.
\end{proof}
    
    Now, we can proceed by using  the fact that the inverse of a block-diagonal matrix is also a block-diagonal matrix and inserting the definitions of $\vm_{\mX'}$, $\vm_{\mX}$, $\mSigma$ and  $\mB$ to show that:
    \begin{align}
        &
        \left(\vm_{\mX'} - \vm_{\mX}\right)^T \mSigma^{-1} \left(\vm_{\mX'} - \vm_{\mX}\right)
        \\
        = &
        \left(\mathrm{vec}\left(\mA_{2:} \mX'\right) - \mathrm{vec}\left(\mA_{2:} \mX\right)\right)^T \mSigma^{-1} \left(\mathrm{vec}\left(\mA_{2:} \mX'\right) - \mathrm{vec}\left(\mA_{2:} \mX\right)\right)
\\
        = &
        \sum_{d=1}^D
        \left(\mA_{2:} \mX'_{:,d} - \mA_{2:} \mX_{:,d}\right)^T \mB^{-1} \left(\mA_{2:} \mX'_{:,d} - \mA_{2:} \mX_{:,d}\right)
    \\
        = &
        \sum_{d=1}^D
        \left(\mA_{2:} \mX'_{:,d} - \mA_{2:} \mX_{:,d}\right)^T
        \frac{1}{\sigma^2} \left(\mA_{2:} \left(\mA_{2:}\right)^T\right)^{-1}
        \left(\mA_{2:} \mX'_{:,d} - \mA_{2:} \mX_{:,d}\right)
 \\
        = &
        \frac{1}{\sigma^2}
        \sum_{d=1}^D
        \left(\mA_{2:} \mDelta_{:,d}\right)^T
         \left(\mA_{2:} \left(\mA_{2:}\right)^T\right)^{-1}
        \left(\mA_{2:}\mDelta_{:,d}\right),
         \\
        = &\label{eq:proof_translation_5}
        \frac{1}{\sigma^2}
        \sum_{d=1}^D
        \left(\mDelta_{:,d}\right)^T
         \left(\mA_{2:}\right)^T \left(\mA_{2:} \left(\mA_{2:}\right)^T\right)^{-1} \mA_{2:}
        \mDelta_{:,d},
    \end{align}
    where for the second to last last equality we have used that $\mX' = \mX + \mDelta$.
    
    Our next step is to compute $\left(\mA_{2:}\right)^T \left(\mA_{2:} \left(\mA_{2:}\right)^T\right)^{-1} \mA_{2:}$.
    Recall that $\mA_{2:} \in \sR^{(N-1) \times D}$ and
    $\mA_{2:} =
    \begin{bmatrix}
        -\ones_{N-1} & \eye_{N-1}
    \end{bmatrix}
    $. Matrix multiplication shows that
    \begin{align*}
    \mA_{2:} \left(\mA_{2:}\right)^T = 
        \begin{bmatrix}
        -\ones_{N-1} & \eye_{N-1}
    \end{bmatrix}
    \begin{bmatrix}
        -\ones_{N-1} & \eye_{N-1}
    \end{bmatrix}^T
    = 
    \ones_{N-1,N-1} + \eye_{N-1}
    =
    \begin{bmatrix}
        2 & 1 & \hdots & 1 \\
        1 & 2 & \hdots & 1 \\
        \vdots & \vdots & \ddots & 1 \\
        1 & 1 & \hdots & 2
    \end{bmatrix}.
    \end{align*}
    The matrix is sufficiently simple to be inverted by inspection:
    \begin{equation*}
        \left(\mA_{2:} \left(\mA_{2:}\right)^T\right)^{-1}
        = 
        \eye_{N-1} - \frac{1}{N} \ones_{N-1,N-1}
        =
        \begin{bmatrix}
        1 - \frac{1}{N} & - \frac{1}{N} & \hdots & - \frac{1}{N} \\
        - \frac{1}{N} & 1 -  \frac{1}{N} & \hdots & - \frac{1}{N} \\
        \vdots & \vdots & \ddots &   \frac{1}{N}\\
        - \frac{1}{N} & - \frac{1}{N} & \hdots & 1- \frac{1}{N}
    \end{bmatrix}.
    \end{equation*}
    Finally, matrix multiplication shows that
    \begin{align}\label{eq:proof_translation_6}
        \left(\mA_{2:}\right)^T \left(\mA_{2:} \left(\mA_{2:}\right)^T\right)^{-1} \mA_{2:}
        = 
        \eye_{N} - \frac{1}{N} \ones_{N,N}
        =
        \begin{bmatrix}
        1 - \frac{1}{N} & - \frac{1}{N} & \hdots & - \frac{1}{N} \\
        - \frac{1}{N} & 1 -  \frac{1}{N} & \hdots & - \frac{1}{N} \\
        \vdots & \vdots & \ddots &   \frac{1}{N}\\
        - \frac{1}{N} & - \frac{1}{N} & \hdots & 1- \frac{1}{N}
        \end{bmatrix}.
    \end{align}
    Note that the matrix in~\cref{eq:proof_translation_6} transforms a vector by subtracting its average from all entries.
    Inserting into~\cref{eq:proof_translation_5} shows that
    \begin{align*}
        &\left(\vm_{\mX'} - \vm_{\mX}\right)^T \mSigma^{-1} \left(\vm_{\mX'} - \vm_{\mX}\right)\\
         =&
         \frac{1}{\sigma^2}
         \sum_{d=1}^D (\mDelta_{:,d})^T \left(\mDelta_{:,d} - \ones_D \overline{\mDelta_{:,d}}\right)\\
         =&
         \frac{1}{\sigma^2}
         \sum_{d=1}^D \left(\mDelta_{:,d} - \ones_D \overline{\mDelta_{:,d}}\right)^T \left(\mDelta_{:,d} - \ones_D \overline{\mDelta_{:,d}}\right)\\
         = &
         \frac{1}{\sigma^2}
         \sum_{n=1}^N\sum_{d=1}^D \left(\mDelta - \ones_N\overline{\mDelta}\right)_{n,d}^2,
    \end{align*}
    where $\overline{\mDelta_{:,d}} \in \sR$ is the average of column $\mDelta_{:,d}$,  $\overline{\mDelta} \in \sR^{1 \times D}$ are the column-wise averages of matrix $\mDelta$, the second equality is due to~\cref{lemma:inner_product_mean} and the third equality uses the definition of inner products.

Taking the square root and using the definition of the Frobenius norm yields our desired result:
\begin{equation*}
    \sqrt{
        \left(\vm_{\mX'} - \vm_{\mX}\right)^T \mSigma^{-1} \left(\vm_{\mX'} - \vm_{\mX}\right)
        }
        =
        \sqrt{\frac{1}{\sigma^2}
         \sum_{n=1}^N\sum_{d=1}^D \left(\mDelta - \ones_N\overline{\mDelta}\right)_{n,d}^2}
        = \frac{1}{\sigma}\left\lvert\left\lvert \mDelta - \ones_N \overline{\mDelta}\right\rvert\right\rvert_2.
\end{equation*}

\subsubsection{Rotation invariance in 2D}\label{appendix:tight_rotation_2d}
\begin{customthm}{6}\label{theorem:2d_rotation_cert_complete}
    Let $g : \sR^{N \times 2} \rightarrow \sY$ be invariant under $\sT = \mathit{SO}(2)$ and $\sH_\sT$ be defined as in~\cref{eq:classifier_set_invariant}.
    Define the indicator function $\rho : \sR^4 \rightarrow \{0,1\}$ with
    \begin{align*}
        \rho(\vq) & =
        \indicator \left[\mathcal{I}_0\left(\sqrt{\evq_1^2 + \evq_2^2}\right) \mathbin{/} 
            \mathcal{I}_0\left(\sqrt{\evq_3^2 + \evq_4^2}\right) \leq \kappa\right],
       \\
        & \text{ with } \kappa \in \sR \text{ such that } 
        \E_{\vq \sim \mathcal{N}\left(
            \vm^{(2)},
            \mSigma
        \right)} \left[\rho(\vq)\right] = p_{\mX, y^*}.
    \end{align*}
    Then
    \begin{equation*}
        \min_{h \in \sH_\sT} \Pr_{\mZ \sim \mu_{\mX'}} \left[h(\mZ) = y^*\right]
        =
        \E_{\vq \sim \mathcal{N}\left(
            \vm^{(1)},
            \mSigma
        \right)} \left[\rho(\vq) \right],
    \end{equation*}
    where
    \begin{align*}
    \vm^{(1)} =
    \frac{1}{\sigma^2}
    \begin{bmatrix}
        2 \epsilon_1 + ||\mX||_2^2 + ||\mDelta||_2^2 \\
        0 \\
        \epsilon_1 + ||\mX||_2^2 \\
        \epsilon_2
    \end{bmatrix},
    \quad
    \vm^{(2)} =
    \frac{1}{\sigma^2}
    \begin{bmatrix}
        \epsilon_1 + ||\mX||_2^2 \\
        - \epsilon_2 \\
        ||\mX||_2^2 \\
        0
    \end{bmatrix}, \\
    \mSigma = \frac{1}{\sigma^2} \begin{bmatrix}
        2 \epsilon_1 + ||\mX||_2^2 + ||\mDelta||_2^2 & 0 & \epsilon_1 + ||\mX||_2^2 & \epsilon_2 \\
        0 & 2 \epsilon_1 + ||\mX||_2^2 + ||\mDelta||_2^2 & - \epsilon_2 & \epsilon_1 + ||\mX||_2^2 \\
        \epsilon_1 + ||\mX||_2^2 & - \epsilon_2 & ||\mX||_2^2 & 0 \\
        \epsilon_2 & \epsilon_1 + ||\mX||_2^2 & 0 & ||\mX||_2^2.
    \end{bmatrix},
    \end{align*}
    with 
    clean data norm $||\mX||_2$, perturbation norm $||\mDelta||_2$ and parameters
    $\epsilon_1 = \left\langle \mX, \mDelta \right\rangle_\mathrm{F}$, 
    $\epsilon_2 = \langle \mX \mR\left(-\nicefrac{\pi}{2}\right)^T, \mDelta \rangle_\mathrm{F}$.
\end{customthm}

We know from~\cref{theorem:main} and our derivations in~\cref{section:verification_rot_2d} that
\begin{equation}\label{eq:rotation_2d_intro_1}
\min_{h \in \sH_\sT} \Pr_{\mZ \sim \mu_{\mX'}} \left[h(\mZ) = y^*\right]
=
\E_{\mZ \sim \mu_{\mX'}} \left[h^*(\mZ)\right],
\end{equation}
with 
\begin{align}
        \label{eq:rotation_2d_intro_2}
        h^*(\mZ) &= \indicator\left[\frac{\beta_{\mX'}(\mZ)}{\beta_{\mX}(\mZ)} \leq \kappa\right]
        \text{ and } \kappa \in \sR_+ \text{ such that } \E_{\mZ \sim \mu_{\mX}} \left[h^*(\mZ)\right] = p_{\mX, y^*}, \\
        \nonumber
    \beta_\mX(\mZ) &= \int_{[0,2\pi]} \exp\left(
        \left\langle  \mZ \mR(\omega)^T, \mX \right\rangle_\mathrm{F} \mathbin{/} \sigma^2
        \right) \ d \omega, \\
        \nonumber
    \mR(\omega) &= \begin{bmatrix}\cos(\omega) & - \sin(\omega) \\ \sin(\omega) & \cos(\omega) \end{bmatrix}.
\end{align}
To prove~\cref{theorem:2d_rotation_cert_complete}, we shall first calculate the integral $\beta_\mX(\mZ)$ characterizing our worst-case classifier and then use the fact that it applies an affine transformation to our random input data sampled from $\mu_\mX$ and $\mu_{\mX'}$.

We begin by factoring out $\cos(\omega)$ and $\sin(\omega)$ from the exponent:
\begin{align*}
        \left\langle  \mZ \mR(\omega)^T, \mX \right\rangle_\mathrm{F}
        & = 
        \sum_{n=1}^N (\mR(\omega) \mZ_n) (\mX_n)^T
        \\
        &=
        \sum_{n=1}^N \cos(\omega) \mZ_{n,1} \mX_{n,1} - \sin(\omega) \mZ_{n,2} \mX_{n,1}
        + \sin(\omega) \mZ_{n,1} \mX_{n,2} + \cos(\omega) \mZ_{n,2} \mX_{n, 2}
        \\
        & = 
            \cos(\omega) 
            \begin{bmatrix}
                \mX_{:,1}
                \\
                \mX_{:,2}
            \end{bmatrix}^T
            \begin{bmatrix}
                \mZ_{:,1}
                \\
                \mZ_{:,2}
            \end{bmatrix}
            + 
            \sin(\omega)
            \begin{bmatrix}
                \mX_{:,2}
                \\
                - \mX_{:,1}
            \end{bmatrix}^T
            \begin{bmatrix}
                \mZ_{:,1}
                \\
                \mZ_{:,2}
            \end{bmatrix}
        \\
        & = 
        \cos(\omega) \cdot \mathrm{vec}(\mX)^T \mathrm{vec}(\mZ)
        +
        \sin(\omega) \cdot \mathrm{vec}\left(\mX \mR\left(- \pi \mathop{/} 2 \right)^T\right)^T \mathrm{vec}(\mZ).
\end{align*}
We can then show that
\begin{align*}
    & \beta_\mX(\mZ)
    \\
    = &
    \int_{[0,2\pi]}
    \exp\left(
        \frac{1}{\sigma^2}
        \left(
        \cos(\omega) \cdot \mathrm{vec}(\mX)^T \mathrm{vec}(\mZ)
        +
        \sin(\omega) \cdot \mathrm{vec}\left(\mX \mR\left(- \pi \mathop{/} 2 \right)^T\right)^T \mathrm{vec}(\mZ)
        \right)
        \right) \ d \omega
    \\
    = &
    \int_{[0,2\pi]}
    \exp\left(
        \cos(\omega)
        \cdot 
        \sqrt{
        \left(\mathrm{vec}(\mX)^T \mathrm{vec}(\mZ) \mathop{/} \sigma^2\right)^2
        +
        \left(\mathrm{vec}\left(\mX \mR\left(- \pi \mathop{/} 2 \right)^T \right)^T \mathrm{vec}(\mZ) \mathop{/} \sigma^2  \right)^2
        }
        \right) \ d \omega
    \\
    = &
    2 \pi \cdot 
    \mathcal{I}_0\left(\sqrt{
        \left(\mathrm{vec}(\mX)^T \mathrm{vec}(\mZ) \mathop{/} \sigma^2\right)^2
        +
        \left(\mathrm{vec}\left(\mX \mR\left(- \pi \mathop{/} 2 \right)^T \right)^T \mathrm{vec}(\mZ) \mathop{/} \sigma^2  \right)^2
        }\right),
\end{align*}
where the second equality follows from the fact that 
    $\cos(\omega) \eta_1 + \sin(\omega) \eta_2 =
    \cos(\omega + \alpha) \sqrt{\eta_1^2 + \eta_2^2}
    $ 
    for some $\alpha \in [0,2 \pi]$ (see also Eq.\ 29 of \cite{Cooper2020})
and the third equality is due to the integral representation of the 
modified Bessel function of the first kind with order $0$ (see~Eq.\ 10.32.1\ of~\cite{NIST}):
    \begin{align*}
        \mathcal{I}_0(z) = \frac{1}{\pi} 
        \int\limits_{[0,\pi]}
        \exp\left( \cos(\alpha) z  \right)
        \ d \alpha.
\end{align*}

We can now insert this expression for the worst-case classifier into the expectations w.r.t.\ clean and perturbed smoothing distributions $\mu_\mX, \mu_\mX'$
from~\cref{eq:rotation_2d_intro_1,eq:rotation_2d_intro_2}
. Note that, by definition, we have $\mathrm{vec}(\mZ) \sim \mathcal{N}(\mathrm{vec}(\mX), \sigma^2 \cdot \eye_{2N})$ for $\mZ \sim \mu_\mX$.
Thus:
\begin{align*}
    \E_{\mZ \sim \mu_{\mX'}} \left[h^*(\mZ)\right] = 
    \E_{\vq \sim \mathcal{N}(\vm^{(1)}, \mSigma)}
    \left[
    \indicator \left[\mathcal{I}_0\left(\sqrt{\evq_1^2 + \evq_2^2}\right) \mathbin{/} 
            \mathcal{I}_0\left(\sqrt{\evq_3^2 + \evq_4^2}\right) \leq \kappa\right]
    \right],
    \\
    \E_{\mZ \sim \mu_{\mX}} \left[h^*(\mZ)\right] = 
    \E_{\vq \sim \mathcal{N}(\vm^{(2)}, \mSigma)}
    \left[
    \indicator \left[\mathcal{I}_0\left(\sqrt{\evq_1^2 + \evq_2^2}\right) \mathbin{/} 
            \mathcal{I}_0\left(\sqrt{\evq_3^2 + \evq_4^2}\right) \leq \kappa\right]
    \right],
\end{align*}
with means $\vm^{(1)} = \mW \mathrm{vec}(\mX')$, $\vm^{(2)} = \mW \mathrm{vec}(\mX)$ and covariance matrix $\mSigma = \sigma^2 \mW \mW^T$, where
\begin{equation*}
    \mW = \frac{1}{\sigma^2} \cdot \begin{bmatrix}
        \mathrm{vec}(\mX')^T\\
        \mathrm{vec}\left(\mX' \mR\left(- \pi \mathop{/} 2 \right)^T\right)\\
        \mathrm{vec}(\mX)^T\\
        \mathrm{vec}\left(\mX \mR\left(- \pi \mathop{/} 2 \right)^T\right)
    \end{bmatrix}.
\end{equation*}
Calculating the matrix-vector and matrix-matrix products yields the values from~\cref{theorem:2d_rotation_cert_complete},
thus proving that the certificate is valid.

\subsubsection{Rotation invariance in 3D}\label{appendix:tight_rotation_3d}
As discussed in~\cref{section:tight_rotation_3d}, we can apply~\cref{theorem:main} to $\sT = \mathit{SO}(3)$, but do not have a closed-form expression for the Haar integral $\beta_\mX(\mZ)$ characterizing our worst-case classifier.
We can however evaluate it using numerical integration.
In the following, we first show how we can reduce the number of integration variables to facilitate numerical integration.
We then show that, similar to the 2D case, the Monte Carlo certification procedure presented in~\cref{section:rotation_2d_tight} only requires sampling from a $16$-dimensional normal distribution, which allows us to obtain tight bounds via a large number of samples at little computational cost.

\textbf{Reducing the number of integration variables.} 
Recall from our proof of~\cref{theorem:main} in~\cref{appendix:tight_methodology,section:verification_rot_3d} that
\begin{equation}
    \beta_\mX(\mZ) = 
    \int_\Omega \cos(\omega_2) \cdot \exp\left(
        \left\langle \mZ \mR(\vomega)^T, \mX \right\rangle_\mathrm{F} \mathbin{/} \sigma^2
    \right) \ d \vomega,
\end{equation}
with $\Omega = [0,2 \pi] \times [-\frac{\pi}{2}, \frac{\pi}{2}] \times [0, 2 \pi]$ and rotation matrix
\begin{align*}
    \mR(\vomega) & = 
    \begin{bmatrix}
        \cos(\omega_1) & -\sin(\omega_1) & 0\\
        \sin(\omega_1) & \cos(\omega_1) & 0\\
        0 & 0 & 1
    \end{bmatrix}
    \cdot
    \begin{bmatrix}
        \cos(\omega_2) & 0 & \sin(\omega_2) \\
        0 & 1 & 0 \\
        - \sin(\omega_2) & 0 & \cos(\omega_2)
    \end{bmatrix}
    \cdot
    \begin{bmatrix}
        1 & 0 & 0\\
        0 & \cos(\omega_3) & -\sin(\omega_3)\\
        0 & \sin(\omega_3) & \cos(\omega_3)
    \end{bmatrix}
    \\
    & = 
    \begin{bmatrix}
        \cos(\omega_1) & -\sin(\omega_1) & 0\\
        \sin(\omega_1) & \cos(\omega_1) & 0\\
        0 & 0 & 1
    \end{bmatrix}
    \cdot
    \begin{bmatrix}
        \cos(\omega_2) & \sin(\omega_2) \sin(\omega_3) & \cos(\omega_3) \sin(\omega_2) \\
        0 & \cos(\omega_3) & - \sin(\omega_3) \\
        - \sin(\omega_2) & \cos(\omega_2) \sin(\omega_3) & \cos(\omega_2) \cos(\omega_3)
    \end{bmatrix}
    \\
    & \vcentcolon= 
    \begin{bmatrix}
        \cos(\omega_1) & -\sin(\omega_1) & 0\\
        \sin(\omega_1) & \cos(\omega_1) & 0\\
        0 & 0 & 1
    \end{bmatrix}
    \cdot
    \tilde{\mR}(\vomega_{2:}),
\end{align*}
with $\tilde{\mR}(\vomega_{2:})$ being the matrix that rotates around the $y$- and $z$-axis by angles $\omega_2$ and $\omega_3$, respectively.
While we cannot evaluate this term analytically, we can calculate the inner integral over $\omega_1$ in order to reduce it to a double integral.
We begin by factoring out $\cos(\omega_1)$ and $\sin(\omega_1)$ from the exponent:
\begin{align*}
    & \left\langle \mZ \mR(\vomega)^T, \mX \right\rangle_\mathrm{F}  \mathbin{/} \sigma^2
    \\
    = &
    \sum_{n=1}^N (\mX_n)^T \mR(\vomega) \mZ_n  \mathbin{/} \sigma^2
    \\
    \begin{split}
    = &
    \cos(\omega_1) \left(
    \sum_{n=1}^N
        \mX_{n,1} \left(\tilde{\mR}(\vomega_{2:}) \mZ_n\right)_1
        + \mX_{n,2}\left(\tilde{\mR}(\vomega_{2:}) \mZ_n\right)_2
    \right)  \mathbin{/} \sigma^2
    \\
    & +
    \sin(\omega_1) \left(
    \sum_{n=1}^N
        \mX_{n,2} \left(\tilde{\mR}(\vomega_{2:}) \mZ_n\right)_1
        - \mX_{n,1}\left(\tilde{\mR}(\vomega_{2:}) \mZ_n\right)_2
    \right)  \mathbin{/} \sigma^2
    \\
    & +
    \sum_{n=1}^N
        \mX_{n,3} \left(\tilde{\mR}(\vomega_{2:}) \mZ_n\right)_3  \mathbin{/} \sigma^2.
    \end{split}
    \\
    =& 
    \cos(\omega_1) \cdot \chi_1(\vomega_{2:}, \mX^T \mZ) +  \sin(\omega_1) \cdot \chi_2(\vomega_{2:}, \mX^T \mZ) + \chi_3(\vomega_{2:}, \mX^T \mZ),
\end{align*}
where $\chi_1,\chi_2,\chi_3$ are shorthands we introduce to avoid clutter in the following derivations.
Due to the entries of $\tilde{\mR}(\vomega_2:)$, the shorthand $\chi_1 : \sR^2 \times \sR^{3 \times 3} \rightarrow \sR$ is defined as
\begin{align*}
   & \chi_1(\vomega_{2:}, \mX^T \mZ))
   \\
   \begin{split}
    \vcentcolon= & 
     \frac{1}{\sigma^2} \sum_{n=1}^N
        \Bigl(
        \mX_{n,1} \left(
            \cos(\omega_2) \mZ_{n,1}
            +
            \sin(\omega_2) \sin(\omega_3) \mZ_{n,2}
            +
            \cos(\omega_3) \sin(\omega_2) \mZ_{n,3}
        \right)
        \\
        & + \mX_{n,2}\left(
            \cos(\omega_3) \mZ_{n,2}
            - \sin(\omega_3) \mZ_{n,3}
        \right)
        \Bigr)
    \end{split}
    \\
    \begin{split}
    = & 
        \cos(\omega_2) (\mX_{:,1})^T \mZ_{:, 1}
        + 
        \sin(\omega_2) \sin(\omega_3) (\mX_{:,1})^T \mZ_{:,2}
        +
        \cos(\omega_3) \sin(\omega_2) (\mX_{:,1})^T \mZ_{:,3} 
        \\
        & +
        \cos(\omega_3) (\mX_{:,2})^T \mZ_{:, 2}
        - \sin(\omega_3)  (\mX_{:,2})^T \mZ_{:, 3}
    \end{split}
    \\
    \begin{split}
    = & 
        \cos(\omega_2) (\mX^T \mZ)_{1,1}
        + 
        \sin(\omega_2) \sin(\omega_3) (\mX^T \mZ)_{1,2}
        +
        \cos(\omega_3) \sin(\omega_2)(\mX^T \mZ)_{1,3}
        \\
        & +
        \cos(\omega_3) (\mX^T \mZ)_{2,2}
        - \sin(\omega_3)  (\mX^T \mZ)_{2,3}
    \end{split}
\end{align*}
Similarly, the shorthand $\chi_2 : \sR^2 \times \sR^{3 \times 3} \rightarrow \sR$ is defined as
\begin{align*}
    & \chi_2(\vomega_{2:,}, \mX^T \mZ) 
    \\
    \begin{split}
    \vcentcolon= & 
        \cos(\omega_2) (\mX^T \mZ)_{2,1}
        + 
        \sin(\omega_2) \sin(\omega_3) (\mX^T \mZ)_{2,2}
        +
        \cos(\omega_3) \sin(\omega_2)(\mX^T \mZ)_{2,3}
        \\
        & -
        \cos(\omega_3) (\mX^T \mZ)_{1,2}
        + \sin(\omega_3)  (\mX^T \mZ)_{1,3}.
    \end{split}
\end{align*}
The shorthand $\chi_3 : \sR^2 \times \sR^{3 \times 3} \rightarrow \sR$ is defined as
\begin{align*}
    & \chi_3(\vomega_{2:,}, \mX^T \mZ) 
    \\
    \begin{split}
    \vcentcolon= & 
        -\sin(\omega_2) (\mX^T \mZ)_{3,1}
        + 
       \cos(\omega_2) \sin(\omega_3) (\mX^T \mZ)_{3,2}
        +
        \cos(\omega_2) \cos(\omega_3) (\mX^T \mZ)_{3,3}
    \end{split}.
\end{align*}

Just like in~\cref{appendix:tight_rotation_2d}, we can use the integral formula for the modified Bessel function of the first kind and order zero to eliminate the integral w.r.t.\ $\omega_1$:
\begin{align*}
        & \beta_\mX(\mZ)
        \\
        = & 
        \int_{\Omega} \cos(\omega_2) \cdot \exp\left(\chi_3(\vomega_{2:}, \mX^T \mZ)\right)
        \cdot 
        \exp \left(
            \cos(\omega_1) \cdot \chi_1(\vomega_{2:}, \mX, \mZ) +  \sin(\omega_1) \cdot \chi_2(\vomega_{2:},\mX^T \mZ)
        \right) \ d \vomega
        \\
        = &
        2 \pi \cdot 
        \int_{[-\frac{\pi}{2}, \frac{\pi}{2}] \times [0, 2 \pi]}
        \cos(\omega_2) \cdot \exp\left(\chi_3(\vomega_{2:}, \mX^T \mZ))\right)
        \cdot
        \mathcal{I}_0 \left(
            \sqrt{
                \chi_1(\vomega_{2:}, \mX, \mZ)^2 +  \chi_2(\vomega_{2:}, \mX, \mZ)^2
            }
        \right) \ d \vomega_{2:}
        \\
        \vcentcolon= &
        \hat{\beta}(\mX^T \mZ).
\end{align*}
We introduce $\hat{\beta}(\mX^T \mZ)$ to avoid clutter in the next equations and to highlight that this function characterizing our worst-case classifier only depends on $\mX^T \mZ \in \sR^{3 \times 3}$.

\textbf{Efficient Monte Carlo certification.}
Just like in~\cref{appendix:tight_rotation_2d}, we can now use the fact that our worst-case classifier only depends on a small number of variables that are the result of linearly transforming our randomized inputs $\mZ \sim \mu_\mX$, namely the 9 entires of $\mX^T \mZ$.
Let $\zeta : \sR^{8} \rightarrow \sR^{3 \times 3}$ with \begin{equation*}
    \zeta(\vq) =
    \begin{bmatrix}
        q_1 & q_3 & q_6 \\
        0 & q_4 & q_7 \\
        q_2 & q_5 & q_8 \\
    \end{bmatrix}
\end{equation*}
be a function that zero-pads its $8$-dimensional input vector before devectorizing into shape $3 \times 3$.
Since $\mathrm{vec}(\mZ) \sim \mathcal{N}\left(\mathrm{vec}(\mX), \sigma^2 \eye_{3N}\right)$ if $\mZ \sim \mu_\mX$, we have by definition of $\hat{\beta}$ that 
\begin{equation*}
    \E_{\mZ \sim \mu_\mX} \left[
        \frac{
            \beta_{\mX'}(\mZ)
        }
        {
            \beta_{\mX}(\mZ)
        }
        \leq \kappa
    \right]
    = 
    \E_{\vq \sim \mathcal{N}(\vm_\mX, \mSigma)} \left[
        \frac{
            \hat{\beta}(\zeta(\vq_{1:8}))
        }
        {
            \hat{\beta}(\zeta(\vq_{9:16}))
        }
        \leq \kappa
    \right],
\end{equation*}
with $\vm_\mX = \mW \mathrm{vec}(\mX)$, $\mSigma = \sigma^2 \mW \mW^T$ and
\begin{equation*}
    \mW = \begin{bmatrix}
        \mX'_{:, 1} & \mX'_{:, 3} & \zeros & \zeros & \mX_{:, 1} & \mX_{:, 3} & \zeros & \zeros \\
        \zeros & \zeros & \mX' & \zeros & \zeros & \zeros & \mX & \zeros \\
        \zeros & \zeros & \zeros & \mX' & \zeros & \zeros & \zeros & \mX 
    \end{bmatrix}^T.
\end{equation*}

Thus, the expectation that provides the optimal value of our variance-constrained optimization problem can be probabilistically bounded via Monte Carlo sampling from a $16$-dimensional normal distribution, regardless of the data dimensionality.

\subsubsection{Roto-translation invariance in 2D and 3D}\label{appendix:tight_roto_translation}
For the roto-granslation group $\mathit{SE}(D)$, we prove that evaluating the tight certificate is equivalent to evaluating the tight certificate for rotation group $\mathit{SO}(D)$ after centering the clean and perturbed data $\mX, \mX'$ by substracting their averages.
\begin{theorem}\label{theorem:roto_translation}
Let $\tilde{\mX} = \mX - \ones_N \overline{\mX}$ 
with column-wise averages $\overline{\mX}, \overline{\mX'} \in \sR^{1 \times D}$.
Further let $\eta$ be a right Haar measure on $\mathit{SO}(3)$ and $\mu_\mX$ our isotropic matrix normal smoothing distribution.
Let 
\begin{equation*}    \beta_\mX^{\mathit{SE}(D)} =
    \int_{\mathit{SE}(D)} \int_{\sR^{D}} \exp\left(
        \left\langle  \mZ \mR^T + \ones_N \vb^T , \mX \right\rangle_\mathrm{F} \mathbin{/} \sigma^2
        \right) \ d \vb \ d \eta(\mR)
\end{equation*}
be the Haar integral characterizing the worst-case classifier for roto-translation invariance (see also~\cref{appendix:verification_rot_trans}) and
\begin{equation*}    \beta_\mX^{\mathit{SO}(D)} =
    \int_{\mathit{SO}(D)} \int_{\sR^{D}} \exp\left(
        \left\langle  \mZ \mR^T , \mX \right\rangle_\mathrm{F} \mathbin{/} \sigma^2
        \right) \ d \eta(\mR)
\end{equation*}
be the Haar integral characterizing the worst-case classifier for rotation invariance (see also~\cref{section:verification_rot_2d,section:verification_rot_3d}).
Then, for all $\mV \in \sR^{N \times D}$ and $\kappa \in \sR_+$, 
\begin{equation*}
    \label{eq:roto_translation_main}
    \E_{\mZ \sim \mu_\mV}\left[
    \indicator\left[
    \frac{
        \beta_{\mX'}^{\mathit{SE}(D)}(\mZ)
    }{
        \beta_\mX^{\mathit{SE}(D)}(\mZ)
    }
    \leq \kappa
    \right]
    \right]
    =
    \E_{\mZ \sim \mu_{\tilde{\mV}}}\left[
    \indicator\left[
    \frac{
        \beta_{\tilde{\mX'}}^{\mathit{SO}(D)}(\mZ)
    }{
        \beta_{\tilde{\mX}}^{\mathit{SO}(D)}(\mZ)
    }
    \leq \kappa
    \right]
    \right].
\end{equation*}
\end{theorem}
Note that these (with $\mV = \mX$ and $\mV = \mX'$) are exactly the expectations determining the optimal value of our variance-constrained optimization problem (see~\cref{theorem:main}).

To prove this theorem, we first simplify the integrand of $\beta_\mX^{\mathit{SE}(D)}(\mZ)$ by moving the rotation matrix into the second argument of the Frobenius inner product (note that, since $\mR$ is a rotation matrix, we have $\mR^{-1} = \mR^T$):
\begin{align*}
        \left\langle  \mZ \mR^T + \ones_N \vb^T , \mX \right\rangle_\mathrm{F}
        = &
        \left\langle  \mZ \mR^T  + \ones_N \vb^T \mR \mR^T , \mX \right\rangle_\mathrm{F}
        \\
        = & 
        \left\langle  \left( \mZ   + \ones_N \vb^T \mR \right) \mR^T , \mX \right\rangle_\mathrm{F}
        \\
        = & 
        \left\langle   \mZ   + \ones_N \vb^T \mR  , \mX \mR \right\rangle_\mathrm{F}
        \\
        = & 
        \left\langle   \mZ   + \ones_N (\mR^T \vb)^T  , \mX \mR \right\rangle_\mathrm{F}
\end{align*}

Next, we bring our integral into the same form as in our derivation of the certificate for translation invariance (see~\cref{appendix:tight_translation}), so that we can easily integrate out the translation vector $\vb$.
First, we eliminate the rotation matrix $\mR^T$ via the substitution $\vc = \mR^T \vb$:
\begin{align*}
\beta_\mX^{\mathit{SE}(D)}(\mZ)
    &= 
    \int_{\mathit{SO}(D)} \int_{\sR^{D}} \exp\left(
        \left\langle   \mZ   + \ones_N (\mR^T \vb)^T  , \mX \mR \right\rangle_\mathrm{F} \mathbin{/} \sigma^2
        \right) \ d \vb \ d \eta(\mR)
    \\
    & =
    \int_{\mathit{SO}(D)} \int_{\sR^{D}} \exp\left(
        \left\langle   \mZ   + \ones_N \vc^T  , \mX \mR \right\rangle_\mathrm{F} \mathbin{/} \sigma^2
        \right) \ d \vc \ d \eta(\mR)
    \\
\end{align*}
For the next step, let
$\mA = \begin{bmatrix}
        1 & \zeros_{N-1}^T\\
        -\ones_{N-1} & \eye_{N-1}
\end{bmatrix}
 \in \sR^{N \times N}$
with inverse
$\mA^{-1} = \begin{bmatrix}
    1 & \zeros_{N-1}^T \\
    \ones_{N-1} & \eye_{N-1}
\end{bmatrix}$ 
and all-zeros vector $\zeros_{N-1} \in \sR^{N-1}$, all-ones vector $\ones_{N-1} \in \sR^{N-1}$ and identity matrix $\eye_{N-1}$.
Multiplying a vector or matrix with $\mA$ is equivalent to subtracting the first row from all other rows.
We introduce the redundant factor $\mA^{-1} \mA$ and make the substitution $\vu = \vc + \mZ_{1}$ to show that
\begin{align*}
    \beta_\mX^{\mathit{SE}(D)}(\mZ)
    &= 
    \int_{\mathit{SO}(D)} \int_{\sR^{D}} \exp\left(
        \left\langle \mA^{-1} \mA  \left(  \mZ   + \ones_N \vc^T  \right), \mX \mR \right\rangle_\mathrm{F} \mathbin{/} \sigma^2
        \right) \ d \vc \ d \eta(\mR)
    \\
    & =
    \int_{\mathit{SO}(D)} \int_{\sR^{D}} \exp\left(
        \left\langle \mA^{-1} 
        \begin{bmatrix}
            \mZ_{1}^T + \vc^T \\
            (\mZ_{2} - \mZ_{1})^T
            \\
            \dots
            \\
            (\mZ_{N} - \mZ_{1})^T
        \end{bmatrix}
        , \mX \mR \right\rangle_\mathrm{F} \mathbin{/} \sigma^2
        \right) \ d \vc \ d \eta(\mR)
    \\
    & =
    \int_{\mathit{SO}(D)} \int_{\sR^{D}} \exp\left(
        \left\langle \mA^{-1} 
        \begin{bmatrix}
            \vu^T \\
            (\mZ_{2} - \mZ_{1})^T
            \\
            \dots
            \\
            (\mZ_{N} - \mZ_{1})^T
        \end{bmatrix}
        , \mX \mR \right\rangle_\mathrm{F} \mathbin{/} \sigma^2
        \right) \ d \vu \ d \eta(\mR)
    \\
    & \propto
    \int_{\mathit{SO}(D)} \int_{\sR^{D}}
    \prod_{d=1}^D
    \mathcal{N}\left(\mA^{-1} \begin{bmatrix}
            \vu_d \\
            \mZ_{2,d} - \mZ_{1,d}
            \\
            \dots
            \\
            \mZ_{N,d} - \mZ_{1,d}
        \end{bmatrix}
        \mid
        \mX \mR_{:, d},
        \sigma^2 \eye_N
    \right)
    \ d \vu \ d \eta(\mR).
\end{align*}
Next, we use the behavior of multivariate normal densities under affine transformation (see also~\cref{lemma:normal_change_of_variables}), as well as the fact that marginalizing out one dimension of a multivariate normal density is equivalent to dropping the correspond row of the mean and corresponding row and column of the adjacency matrix, to show that
\begin{align*}
    \beta_\mX^{\mathit{SE}(D)}(\mZ)
    & \propto
    \int_{\mathit{SO}(D)} \int_{\sR^{D}}
    \prod_{d=1}^D
    \mathcal{N}\left( \begin{bmatrix}
            \vu_d \\
            \mZ_{2,d} - \mZ_{1,d}
            \\
            \dots
            \\
            \mZ_{N,d} - \mZ_{1,d}
        \end{bmatrix}
        \mid
        \mA \mX \mR_{:, d},
        \sigma^2 \mA \mA^T
    \right)
    \ d \vu \ d \eta(\mR)
    \\
    &= 
    \int_{\mathit{SO}(D)}
    \prod_{d=1}^D
    \mathcal{N}\left( \begin{bmatrix}
            \mZ_{2,d} - \mZ_{1,d}
            \\
            \dots
            \\
            \mZ_{N,d} - \mZ_{1,d}
        \end{bmatrix}
        \mid
        \mA_{2:} \mX \mR_{:, d},
        \sigma^2 \mA_{2:} (\mA_{2:})^T
    \right)
    \ d \eta(\mR)
    \\
    &= 
    \int_{\mathit{SO}(D)} 
    \prod_{d=1}^D
    \mathcal{N}\left( \mA_{2:} \mZ_{:, d}
        \mid
        \mA_{2:} \mX \mR_{:, d},
        \sigma^2 \mA_{2:} (\mA_{2:})^T
    \right)
    \ d \eta(\mR)
    \\
    & \propto
    \int_{\mathit{SO}(D)}
    \prod_{d=1}^D
    \exp\left(
        \frac{1}{\sigma^2}
            \mZ_{:,d}^T (\mA_{2:})^T \left(\mA_{2:} (\mA_{2:})^T\right)^{-1} \mA_{2:} \mX \mR_{:, d}
        \right)
        \ d \eta(\mR)
    \\
\end{align*}
Finally, reusing the fact that $(\mA_{2:})^T \left(\mA_{2:} (\mA_{2:})^T\right)^{-1} \mA_{2:} \mX = \tilde{\mX}$ with $\tilde{\mX} = \mX - \ones_N \overline{\mX}$ from~\cref{appendix:tight_translation} proves that
\begin{align*}
    \beta_\mX^{\mathit{SE}(D)}(\mZ)  & \propto
     \int_{\mathit{SO}(D)}
    \prod_{d=1}^D
    \exp\left(
        \frac{1}{\sigma^2}
            \mZ_{:,d}^T \tilde{\mX} \mR_{:, d}
        \right)
        \ d \eta(\mR)
        \\
       &  = 
         \int_{\mathit{SO}(D)}
         \exp\left(
            \left\langle
            \mZ \mR^T, \tilde{\mX} 
            \right\rangle
            \mathbin{/} \sigma^2
        \right)
        \ d \eta(\mR).
\end{align*}
The last term is the Haar integral $\beta_{\tilde{\mX}}^{\mathit{SO}(D)}(\mZ)$ characterizing the worst-case classifier for $\mathit{SO}(D)$, after centering $\mX$. This means that we have 
\begin{equation*}
    \frac{
    \beta_{\mX'}^{\mathit{SE}(D)}(\mZ)
    }{
    \beta_\mX^{\mathit{SE}(D)}(\mZ)
    }
    = 
    \frac{
    \beta_{\tilde{\mX'}}^{\mathit{SO}(D)}(\mZ)
    }{
    \beta_{\tilde{\mX}}^{\mathit{SO}(D)}(\mZ)
    }.
\end{equation*}
Note that, when using $\propto$ in the previous steps, we have only removed constant factors that appear in the enumerator and denominator and thus cancel out.

The last thing we need to do is verify that not only the classifiers, but also their expectations under $\mV$ and $\tilde{\mV}$ are identical (see~\cref{eq:roto_translation_main}).
We use the fact that the worst-case classifier only depends on a low-dimensional linear projection of our random input variable $\mZ$ (similar to our derivations for Monte Carlo evaluation in~\cref{appendix:tight_rotation_2d,appendix:tight_rotation_3d}):
\begin{align*}
    \beta_\mX^{\mathit{SE}(D)}(\mZ)  & \propto
     \int_{\mathit{SO}(D)}
    \prod_{d=1}^D
    \exp\left(
        \frac{1}{\sigma^2}
            \mZ_{:,d}^T \tilde{\mX} \mR_{:, d}
        \right)
        \ d \eta(\mR)
        \\
    & =
    \int_{\mathit{SO}(D)}
    \prod_{d=1}^D
    \exp\left(
        \frac{1}{\sigma^2}
            (\tilde{\mX}^T \mZ_{:,d})^T  \mR_{:, d}
        \right)
        \ d \eta(\mR)
        \\
    & =
    \int_{\mathit{SO}(D)}
    \exp\left(
        \left\langle
            \tilde{\mX}^T \mZ, \mR
        \right\rangle
        \mathbin{/} \sigma^2
        \right)
        \ d \eta(\mR)
        \\
\end{align*}

By definition of our smoothing distribution, we have $\mathrm{vec}(\mZ) \sim \mathcal{N}(\mathrm{vec}(\mV) \mid \mathrm{vec}(\mV), \sigma^2 \eye_{D\cdot N})$ for $\mZ \sim \mu_\mV$.
Because our worst-case classifier for $\mathit{SE}(D)$ performs a linear transformation of $\mathrm{vec}(\mZ)$, we have

\begin{align*}
    & \E_{\mZ \sim \mu_\mV}\left[
    \indicator\left[
    \frac{
        \beta_{\mX'}^{\mathit{SE}(D)}(\mZ)
    }{
        \beta_\mX^{\mathit{SE}(D)}(\mZ)
    }
    \leq \kappa
    \right]
    \right]
    \\
    =& 
    \E_{\mZ \sim \mu_\mV}\left[
    \indicator\left[
    \frac{
    \int_{\mathit{SO}(D)}
    \exp\left(
        \left\langle
            \tilde{\mX'}^T \mZ, \mR
        \right\rangle
        \mathbin{/} \sigma^2
        \right)
        \ d \eta(\mR)
    }{
        \int_{\mathit{SO}(D)}
    \exp\left(
        \left\langle
            \tilde{\mX}^T \mZ, \mR
        \right\rangle
        \mathbin{/} \sigma^2
        \right)
        \ d \eta(\mR)
    }
    \leq \kappa
    \right]
    \right]
    \\
    =& 
   \E_{\vq \sim \mathcal{N}(\vm_\mV, \mSigma)}\left[
    \indicator\left[
    \frac{
    \int_{\mathit{SO}(D)}
    \exp\left(
        \left\langle
            \mathrm{vec}^{-1}(\vq_{1:9}), \mR
        \right\rangle
        \mathbin{/} \sigma^2
        \right)
        \ d \eta(\mR)
    }{
        \int_{\mathit{SO}(D)}
    \exp\left(
        \left\langle
             \mathrm{vec}^{-1}(\vq_{10:18}), \mR
        \right\rangle
        \mathbin{/} \sigma^2
        \right)
        \ d \eta(\mR)
    }
    \leq \kappa
    \right]
    \right],
\end{align*}
with $\vm_\mV = \mW \mathrm{vec}(\mV)$ and $\mSigma = \mW \mW^T$, where
\begin{equation*}
    \mW = \begin{bmatrix}
        \tilde{\mX'} &\zeros & \zeros  & \tilde{\mX} & \zeros & \zeros \\
        \zeros & \tilde{\mX'} & \zeros  & \zeros & \tilde{\mX} & \zeros \\
        \zeros & \zeros & \tilde{\mX'} & \zeros & \zeros & \tilde{\mX} \\
    \end{bmatrix}^T.
\end{equation*}
Similarly, we have for $\mathit{SO}(3)$ (after centering $\mV$), that
\begin{align*}
    & \E_{\mZ \sim \mu_{\tilde{\mV}}}\left[
    \indicator\left[
    \frac{
        \beta_{\tilde{\mX'}}^{\mathit{SO}(D)}(\mZ)
    }{
        \beta_{\tilde{\mX}}^{\mathit{SO}(D)}(\mZ)
    }
    \leq \kappa
    \right]
    \right]
    \\
    =& 
   \E_{\vq \sim \mathcal{N}(\vm_{\tilde{\mV}}, \mSigma)}\left[
    \indicator\left[
    \frac{
    \int_{\mathit{SO}(D)}
    \exp\left(
        \left\langle
            \mathrm{vec}^{-1}(\vq_{1:9}), \mR
        \right\rangle
        \mathbin{/} \sigma^2
        \right)
        \ d \eta(\mR)
    }{
        \int_{\mathit{SO}(D)}
    \exp\left(
        \left\langle
             \mathrm{vec}^{-1}(\vq_{10:18}), \mR
        \right\rangle
        \mathbin{/} \sigma^2
        \right)
        \ d \eta(\mR)
    }
    \leq \kappa
    \right]
    \right],
\end{align*}
with $\vm_{\tilde{\mV}} = \mW \mathrm{vec}(\tilde{\mV})$ and $\mSigma = \mW \mW^T$.
Finally, due to the fact (see~\cref{lemma:inner_product_mean}) that calculating an inner product between a centered and an uncentered vector
(here: columns of $\tilde{\mX}$ and $\mV$)
is equivalent to centering both vectors before calculating the inner product
(here: columns of $\tilde{\mX}$ and $\tilde{\mV}$),
we have $\vm_\mV = \vm_{\tilde{\mV}}$. Thus, both expectations are equal, which concludes our proof.

\clearpage

\changelocaltocdepth{2}
\subsection{Monte Carlo certification}\label{appendix:monte_carlo_certification}
All of the discussed tight certificates, save for the one for translation invariance, are of the form
\begin{align*}
    \min_{h \in \sH_\sT}\Pr_{\mZ \sim \mu_{\mX'}}\left[h(\mZ) = y^* \right]] = &\Pr_{V \sim \upsilon^{(1)}} \left[\rho(V) \leq \kappa\right] \\
    \text{with } \kappa \in \sR \text{ such that } &\Pr_{V \sim \upsilon^{(2)}} \left[\rho(V) \leq \kappa\right] = p_{\mX,y^*} \\
    \text{and } &p_{\mX,y^*} \vcentcolon= \Pr_{\mZ \sim \mu_\mX}\left[g(\mZ) = y^*\right],
\end{align*}
where $g: \sR^{N \times D} \rightarrow \sY$ is our base classifier, $\mu_\mX, \mu_{\mX'}$ are the  clean and perturbed smoothing distribution,
$\upsilon^{(1)}$ and $\upsilon^{(2)}$ are distributions over some set $\sS$, and $\rho : \sS \rightarrow \sR$ is some arbitrary scalar-valued function.

As discussed in~\cref{section:rotation_2d_tight}, we propose to compute a narrow probabilistic lower  bound on $\min_{h \in \sH_\sT}\Pr_{\mZ \sim \mu_{\mX'}}\left[h(\mZ) = y^* \right]]$ by combining three confidence bounds.

Before computing these bounds, we have to inspect the monotonicity of our certificate.
Evidently, $\Pr_{V \sim \upsilon^{(1)}} \left[\rho(V) \leq \kappa\right]$ is monotonically decreasing in $\kappa$.
Furthermore, $\kappa$ is monotonically decreasing in $p_{\mX,y^*}$: If $p_{\mX,y^*}$ decreases, then a smaller $\kappa$ is sufficient for fulfilling the constraint $\Pr_{V \sim \upsilon^{(2)}} \left[\rho(V) \leq \kappa \right]
= p_{\mX,y^*}
$.
Thus, we can compute a lower bound on our certificate by
\begin{enumerate}
    \item lower-bounding $p_{\mX,y^*}$, i.e.\ $\underline{p_{\mX,y^*}} \leq p_{\mX,y^*}$,
    \item lower-bounding the $\kappa \in \sR$ that fulfills $\Pr_{V \sim \upsilon^{(2)}} \left[\rho(V) \leq \kappa\right] = \underline{p_{\mX,y^*}}$, i.e. $\underline{\kappa} \leq \kappa$
    \item and then lower-bounding $\Pr_{V \sim \upsilon^{(1)}} \left[\rho(V) \leq \underline{\kappa}\right]$.
\end{enumerate}

The random variable $\indicator\left[g(\mZ) = y^*\right]$ is a Bernoulli random variable.
Just like other randomized smoothing methods (e.g.~\cite{Cohen2019}), we can lower-bound $p_{\mX,y^*}$ by evaluating $g$ on $N_1$ samples from $\mu_\mX$ and computing a  binomial proportion confidence bound, such as the the Clopper-Pearson confidence bound~\cite{Clopper1934}.

Lower-bounding $\kappa$ requires computing a lower bound on $F^{-1}(\underline{p_{\mX,y^*}})$, where $F^{-1}$ is the quantile function of $\rho(V)$ with $V \sim \upsilon^{(2)}$.
A non-parametric lower confidence bound on this quantile can be constructed by evaluating $\rho(V)$ on $N_2$ samples from $\upsilon^{(2)}$ and returning the largest order statistic $R^{(n)}$ such that $\left[\rho(V) \leq R^{(n)}\right] \leq \underline{p_{\mX,y^*}}$ holds with high probability (see~Section 5.2.1.\ of~\cite{Hahn1991} and~\cref{algorithm:prob_cert} below).

The random variable $\left[\rho(V) \leq \underline{\kappa}\right]$ is another  Bernoulli random variable and can thus be lower-bounded by evaluating $\rho(V)$ on $N_3$ samples from $\mu_{\mX'}$ and computing a  binomial proportion confidence bound.

We want all three confidence bounds to simultaneously hold with high probability $1 - \alpha$.
We ensure this by using Holm-Bonferroni~\cite{Holm1979} correction to account for the multiple comparisons problem.\footnote{Holm-Bonferroni correction has already been used in the context of randomized smoothing, see~\cite{Fischer2021}.}
In our case, this corresponds to computing the first bound with significance $\alpha$, the second one with significance $\nicefrac{\alpha}{2}$ and the last one with significance $\nicefrac{\alpha}{3}$.

\cref{algorithm:prob_cert} summarizes our certification procedure. $\textsc{LowerCounfBound}$ refers to the Clopper-Pearson lower confidence bound.  
$\textsc{BinPValue}(n, N_2, \geq, \underline{p_{\mX,y^*}})$ refers to the $p$-value of a Binomial test with the null-hypothesis that the success probability is greater or equal $\underline{p_{\mX,y^*}}$.
Note that only the first confidence bound requires evaluating the base classifier $g$. For the other two confidence bounds a large number of samples can be evaluated at little computational cost.

\clearpage

\begin{algorithm}[H]
\caption{Monte Carlo certification procedure}
\label{algorithm:prob_cert}
\begin{algorithmic}
\Function{ProbCertify}{$y^*$, $g$, $\mu_{\mX}$, $\upsilon^{(1)}$, $\upsilon^{(2)}$, $N_1$, $N_2$, $N_3$, $\alpha$}
  \State $\mZ^{(1)}, \dots, \mZ^{(N_1)} \gets$ \Call{Sample}{$\mu_{\mX}$, $N_1$}
  \State $\mathrm{count}_1 \gets \sum_{n=1}^{N_1} \indicator\left[g(\mZ^{(n)}) = y^*\right]$
  \State $\underline{p_{\mX,y^*}} \gets$ \Call{LowerConfBound}{$\mathrm{count}_1$, $N_1$, $1 - \alpha$} \Comment{First bound}
  \State $n^* \gets \max \left\{n \mid \Call{BinPValue}{n, N_2, \geq, \underline{p_{\mX,y^*}}} < \frac{\alpha}{2}\right\}$
  \State $V^{(1)}, \dots, V^{(N_2)} \gets$ \Call{Sample}{$\upsilon^{(2)}$, $N_2$}
  \State $R^{(1)}, \dots, R^{(N_2)} \gets$ \Call{SortAscending}{$\rho(V^{(1)}), \dots, \rho(V^{(N_2)})$}
  \State $\underline{\kappa} \gets R^{(n^*)}$ \Comment{Second bound}
  \State $V^{(1)}, \dots, V^{(N_3)} \gets$ \Call{Sample}{$\upsilon^{(1)}$, $N_3$}
  \State $\mathrm{count}_2 \gets \sum_{n=1}^{N_3} \indicator\left[\rho(V) \leq \underline{\kappa} \right]$
  \State \Return \Call{LowerConfBound}{$\mathrm{count}_2$, $N_3$, $1 - \frac{\alpha}{3}$} \Comment{Third bound}
\EndFunction
\end{algorithmic}
\end{algorithm}

\null \vfill


\clearpage

\section{Proof of Theorem 5}\label{appendix:rotation_non_tightness}
Next, we prove that the post-processing-based certificate for rotation-invariance ( $\sT = \mathit{SO}(D)$ ) is not tight, which we formalized in~\cref{section:rotation_2d_tight} as follows:
\rotationnontightness*
The right-hand side term is the optimal value of the black-box optimization problem evaluated for perturbed input $\mX'\mR^T$, i.e.\
\begin{align*}
    \min_{h \in \sH}   \Pr_{\mZ \sim \mu_{\mX'\mR^T}} \left[h(\mZ) = y^*\right],
\end{align*}
with $\smash{\sH = \left\{h : \sR^{N \times D} \rightarrow \sY \mid
  \Pr_{\mZ \sim \mu_\mX} \left[h(\mZ) = y^*\right]
   \geq
  p_{\mX,y^*}
\right\}}$.
The left-hand side term is the optimal value of the gray-box optimization problem evaluated for perturbed input $\mX'$.
Note that $\sH_\sT$ is a set of rotation invariant classifiers. In~\cref{appendix:invariance_preservation} we have proven that rotation of the smoothing distribution's mean does not have an effect on the prediction probabilities of such classifiers, i.e.\
\begin{align*}
    \min_{h \in \sH_\sT} \Pr_{\mZ \sim \mu_{\mX'}} \left[h(\mZ) = y^*\right]
    =
    \min_{h \in \sH_\sT} \Pr_{\mZ \sim \mu_{\mX'\mR^T}} \left[h(\mZ) = y^*\right].
\end{align*}
Thus, we can prove~\cref{theorem:rotation_non_tightness} by proving the following, more general statement:
\begin{lemma}\label{lemma:non_tightness_auxil_1}
Let $g : \sR^{N \times D} \rightarrow \sY$ be invariant under $\sT = \mathit{SO}(D)$ and $\sH_\sT$ be defined as in~\cref{eq:classifier_set_invariant}.
    Further assume that $\mX' \neq \mX$ and that $p_{\mX,y^*} \in (0,1)$. Then:
    \begin{align}\label{eq:non_tightness_graybox_blackbox}
       \min_{h \in \sH_\sT} \Pr_{\mZ \sim \mu_{\mX'}} \left[h(\mZ) = y^*\right] 
       > \min_{h \in \sH} \Pr_{\mZ \sim \mu_{\mX'}} \left[h(\mZ) = y^*\right] 
    \end{align}
\end{lemma}
We can do so by showing that no optimal solution to the r.h.s.\ black-box problem from~\cref{eq:non_tightness_graybox_blackbox} is a feasible solution to the l.h.s.\ gray-box problem.
More formally, let $\sH^* \subseteq \sH$ be the set of all classifiers minimizing the r.h.s.\ black box problem.
We have $\sH^{*} \subseteq \sH$ and $\sH_\sT \subseteq \sH$. By showing that $\sH^* \cap \sH_\sT = \emptyset$, we prove that $\sH_\sT \subseteq \sH \setminus \sH^*$, i.e.\ all rotation invariant classifiers from $\sH_\sT$ yield strictly larger optimal values.

Recall that \textit{an} optimal solution to the r.h.s.\ black-box problem in~\cref{eq:non_tightness_graybox_blackbox} is given by the Neyman-Pearson lemma~\cite{Neyman1933}.
Later work on hypothesis testing has shown that the Neyman-Pearson lemma is not only a sufficient, but a necessary condition for optimality\cite{Dantzig1951}:
Every most powerful test must fulfill the likelihood ratio inequalities, save for sets of zero measure.
For our formulation of the Neyman-Pearson lemma (see~\cref{lemma:neyman_pearson_lower}), this means that any classifier $h$ that is an optimal solution to the r.h.s.\ black-box problem from~\cref{eq:non_tightness_graybox_blackbox} must fulfill
\begin{align*}
    h(\mZ) \in \begin{cases}
        \{y^*\} & \text{if } \frac{\mu_{\mX'}(\mZ)}{\mu_{\mX}(\mZ)} < \kappa \\
        \sY \setminus \{y^*\} & \text{if } \frac{\mu_{\mX'}(\mZ)}{\mu_{\mX}(\mZ)} > \kappa
    \end{cases}
\end{align*}
for some $\kappa \in \sR$, save for sets of zero measure.
\citet{Cohen2019} have shown that if $\mX' \neq \mX$ and $p \in (0,1)$ and $\mu_\mX, \mu_{\mX'}$ are isotropic Gaussian distributions, then these likelihood ratio inequalities correspond to a linear decision boundary, i.e. any optimal solution most fulfill\
\begin{align}\label{eq:behave_like_linear_1}
    h(\mZ) \in \begin{cases}
        \{y^*\} & \text{if } \left\langle \mW, \mZ \right\rangle_\mathrm{F} + b < 0 \\
        \sY \setminus \{y^*\} & \text{if } \left\langle \mW, \mZ \right\rangle_\mathrm{F} + b > 0
    \end{cases}
\end{align}
for some $b \in \sR$ and $\mW \in \sR^{N \times D}$ with $\vw \neq \zeros_{N,D}$, save for sets of zero measure.
Here, $\left\langle\mA, \mB \right\rangle_\mathrm{F} = \mathrm{vec}(\mW)^T\mathrm{vec}(\mZ)$ is the Frobenius inner product.

Therefore, we can prove~\cref{lemma:non_tightness_auxil_1} by showing that classifiers complying with~\cref{eq:behave_like_linear_1} (save for sets of zero measure) are not rotation invariant, i.e.\ they are not feasible solutions to the gray-box optimization problem.

We first do this for the case that $D$ is even and then for the case that $ D$ is odd.

\begin{lemma}\label{lemma:non_tightness_auxil_2}
Assume that $D = 2k$ for some $k \in \sN$.
Let $h: \sR^{N \times D} \rightarrow \sY$ be a classifier that fulfills
\begin{align}\label{eq:behave_like_linear_2}
   h(\mZ) \in \begin{cases}
        \{y^*\} & \text{if } \left\langle \mW, \mZ \right\rangle_\mathrm{F} + b < 0 \\
        \sY \setminus \{y^*\} & \text{if } \left\langle \mW, \mZ \right\rangle_\mathrm{F} + b > 0
    \end{cases}
\end{align}
for some $b \in \sR$ and $\vw \in \sR^{N \times D}$ with $\vw \neq \zeros_{N\cdot D}$, save for sets of zero measure.
Then, there is an input $\hat{\mZ}$ and a rotation matrix $\mR \in \mathit{SO}(D)$ such that 
$h(\hat{\mZ}) \neq h(\hat{\mZ} \mR^T)$.
\begin{proof}
    \textbf{Case 1.} Assume that $b \geq 0$.

    Consider the set $\sS = \left\{\mZ \mid  \left\langle \mW, \mZ \right\rangle_\mathrm{F} < b \right\}$.
    By construction, we have $\forall \mZ \in \sS: \left\langle \mW, \mZ \right\rangle_\mathrm{F} + b < 0$
    and $\forall \mZ \in \sS: \left\langle \mW, -\mZ \right\rangle_\mathrm{F} + b > 0$.

    There must be at least one $\hat{\mZ} \in \sS$ with $h(\hat{\mZ}) = y^{*}$ and $h(\hat{\mZ}) \neq y^*$.
    Otherwise, all points $\mZ \in \sS$ would need to fulfill $h(\mZ) \neq y^*$ or $h(-\mZ) = y^*$, in which case we would have sets of non-zero measure violating the likelihood ratio inequalities from~\cref{eq:behave_like_linear_2}.

    Note that $-\hat{\mZ} = \hat{\mZ} (- \eye_N)^T$ and $- \eye_N \in \mathit{SO}(D)$, because $- \eye_N$ is orthonormal and $\mathrm{det}(-\eye_N) = 1$, due to $D$ being even.
    Thus, we have found an input $\hat{\mZ}$ and a rotation matrix $\mR \in \mathit{SO}(D)$ such that 
$h(\mZ) \neq h(\mZ \mR^T)$.
    
    \textbf{Case 2.} Assume that $b < 0$.
    This case follows analogously by constructing the set 
    $\sS = \left\{\mZ \mid  \left\langle \mW, \mZ \right\rangle_\mathrm{F} > b \right\}$ and considering a point
    $\hat{\mZ} \in \sS$ with $h(\hat{\mZ}) \neq y^{*}$ and $h(-\hat{\mZ}) = y^{*}$
\end{proof}
\end{lemma}

Now, we consider the case that $D$ is odd. Here, $- \eye_N \notin \mathit{SO}(D)$, because $\mathrm{det}(- \eye_N) = -1$. Therefore, we will need to use slightly more complicated constructions.
\begin{lemma}\label{lemma:non_tightness_auxil_3}
Assume that $D = 2k - 1$ for some $k \in \sN$.
Let $h: \sR^{N \times D} \rightarrow \sY$ be a classifier that fulfills
\begin{align}\label{eq:behave_like_linear_3}
   h(\mZ) \in \begin{cases}
        \{y^*\} & \text{if } \left\langle \mW, \mZ \right\rangle_\mathrm{F} + b < 0 \\
        \sY \setminus \{y^*\} & \text{if } \left\langle \mW, \mZ \right\rangle_\mathrm{F} + b > 0
    \end{cases}
\end{align}
for some $b \in \sR$ and $\vw \in \sR^{N \times D}$ with $\vw \neq \zeros_{N\cdot D}$, save for sets of zero measure.
Then, there is an input $\hat{\mZ}$ and a rotation matrix $\mR \in \mathit{SO}(D)$ such that 
$h(\hat{\mZ}) \neq h(\hat{\mZ} \mR^T)$.
\end{lemma}
\begin{proof}
    In the following, let $\mA = 
    \begin{bmatrix}
        1 & \zeros_{D}^T \\
        \zeros_{D-1} & - \eye_{D-1}
    \end{bmatrix}
    $.
    We have $\mA \in \mathit{SO}(D)$, because $\mA$ is orthonormal and $\mathrm{det}(\mA) = 1$, due to $D$ being odd.

     \textbf{Case 1.} Assume that $b \geq 0$. 
     
     We know that $\mW \neq \zeros_{N , D}$. Without loss of generality, assume that $\mW_{:, 2:} \neq \zeros_{N , (D-1)}$, i.e.\ at least one of the the last $D-1$ columns is non-zero.
     
     Consider the set $\sS = \left\{\mZ \mid \mW_{:,1}^T \mZ_{:,1} \in [0,1] \land \left\langle \mW_{:, 2:}, \mZ_{:, 2:} \right\rangle_\mathrm{F} < b - 1 \right\}$.
    By construction, we have $\forall \mZ \in \sS: \left\langle \mW, \mZ \right\rangle_\mathrm{F} + b < 0$
    and $\forall \mZ \in \sS: \left\langle \mW,  \mZ \mA^T \right\rangle_\mathrm{F} + b > 0$.
    
    There must be at least one $\hat{\mZ} \in \sS$ with $h(\hat{\mZ}) = y^{*}$ and $h(\hat{\mZ}\mA^T) \neq y^*$.
    Otherwise, all points $\mZ \in \sS$ would need to fulfill $h(\mZ) \neq y^*$ or $h(\mZ \mA^T) = y^*$, in which case we would have sets of non-zero measure violating \cref{eq:behave_like_linear_2}.

    \textbf{Case 2.} Assume that $b < 0$. 
     This case follows analogously by constructing the set 
    $\sS = \left\{\mZ \mid \mW_{:,1}^T \mZ_{:,1} \in [-1,0] \land \left\langle \mW_{:, 2:}, \mZ_{:, 2:} \right\rangle_\mathrm{F} > b + 1 \right\}$ and considering a point
    $\hat{\mZ} \in \sS$ with $h(\hat{\mZ}) \neq y^{*}$ and $h(\mA\hat{\mZ}) = y^{*}$.
\end{proof}

\cref{lemma:non_tightness_auxil_2,lemma:non_tightness_auxil_3} combined prove~\cref{lemma:non_tightness_auxil_1}, which in turn proves~\cref{theorem:rotation_non_tightness}.

\clearpage

\section{Multi-class certificates}\label{appendix:multiclass}

In our discussions in~\cref{section:background_randomized_smoothing,section:postprocessing,section:tight_certificates},
we have only considered binary certificates\footnote{Not to be confused with certificates that are limited to binary classifiers.}.
That is, prediction $y^*$ of smoothed classifier is certifiably robust to a perturbed input $\mX'$
if the probability of base classifier $g$ predicting $y^*$ under perturbed smoothing distribution $\mu_{\mX'}$ is greater than the probability of all other classes combined. That is, $p_{\mX',y^*} > \frac{1}{2}$. 
We have then, for different invariances, computed lower bounds $\underline{p_{\mX',y^*}} \leq p_{\mX',y^*}$.
If $\underline{p_{\mX',y^*}} > \frac{1}{2}$, then $p_{\mX',y^*} > \frac{1}{2}$ and the prediction is certifiably robust.

Alternatively, one can certify robustness by showing that $y^*$ remains more likely than the second most likely class under the perturbed smoothing distribution~\cite{Cohen2019}, i.e.
$p_{\mX',y^*} > \max_{y \neq y^*} p_{\mX',y}$. To this end, one can compute a lower bound $\underline{p_{\mX',y^*}} \leq p_{\mX',y^*}$ and an upper-bound 
$\overline{\max_{y \neq y^*} p_{\mX',y}} \geq \max_{y \neq y^*} p_{\mX',y}$.
If $\underline{p_{\mX',y^*}} > \overline{\max_{y \neq y^*} p_{\mX',y}}$, then robustness  is guaranteed.
Due to the monotonicity of randomized smoothing certificates w.r.t.\ to the prediction probabilities, this can be achieved
by showing that
$\underline{p_{\mX',y^*}} > \overline{p_{\mX',y'}}$, where $y' = \mathrm{argmax}_{y \neq y^*} p_{\mX,y}$ is the second most likely class under the clean smoothing distribution.
This second most likely class $y'$ can be found via a binomial test, see~\cite{Cohen2019,Hung2019}.

The lower bound $\underline{p_{\mX',y^*}}$ can be obtained using the same formulae discussed throughout the main text (e.g.\ $\underline{p_{\mX',y^*}}= \Phi\left(\Phi^{-1}(p_{\mX,y'}) - \frac{||\Delta||_2}{\sigma}\right)$ for black-box randomized smoothing).
The upper bound $\overline{p_{\mX',y'}}$ can be found via the Neyman-Pearson upper bound discussed in~\cref{appendix:neyman_pearson}:
\neymanpearsonupper*
Note that it is identical to the lower bound, save for replacing "$\leq$" with "$\geq$" and vice-versa.

The resulting black-box upper bound is $\overline{p_{\mX',y'}} = \Phi\left(\Phi^{-1}(p_{\mX,y'}) + \frac{||\Delta||_2}{\sigma}\right)$.
It is identical to the black-box lower bound, save for replacing "$-$" with "$+$".
Substituting both bounds into $\underline{p_{\mX',y^*}} > \overline{\max_{y \neq y^*} p_{\mX',y}}$ and applying the inverse-normal CDF $\Phi^{-1}$ to both sides shows that the classifier is certifiably robust if $||\mDelta|| \leq \frac{\sigma}{2} \left(\Phi^{-1}(p_{\mX',y^*}) -  \Phi^{-1}(p_{\mX',y'}) \right)$~\cite{Cohen2019}.

Evidently, the orbit-based versions of the multi-class certificates are identical to their binary counterparts, except that the radius of the underlying black-box certificate changes.

For the tight gray-box approach, upper bounds can be derived by going through the same derivations as in~\cref{appendix:tight_certificates}, but using the Neyman-Pearson upper bounds from~\cref{lemma:neyman_pearson_upper}, i.e.\ replacing "$\leq$" with "$\geq$" and vice-versa. (see~\cref{lemma:neyman_pearson_upper}).
In the case of translation invariance, this results in upper bound 
$\Phi\left(\Phi^{-1}(p_{\mX,y'}) + \frac{||\Delta - \ones_N \overline{\Delta}||_2}{\sigma}\right))$.
In the cases involving rotation invariance, this results in bounds of the form
\begin{align*}
    &\Pr_{V \sim \upsilon^{(1)}} \left[\rho(V) \geq \kappa\right] \\
    \text{with } \kappa \in \sR \text{ such that } &\Pr_{V \sim \upsilon^{(2)}} \left[\rho(V) \geq \kappa\right] = p_{\mX,y'} \\
    \text{and } &p_{\mX,y^*} \vcentcolon= \Pr_{\mZ \sim \mu_\mX}\left[g(\mZ) = y'\right],
\end{align*}
i.e.~ the same certificates, but with ``$\geq \kappa$'' instead of ``$\leq \kappa$''.

Our Monte Carlo certification procedure proposed in~\cref{appendix:monte_carlo_certification} can also easily be adapted to computing an upper bound by replacing "$\leq$" with "$\geq$" and vice-versa.:
\begin{algorithm}[H]
\caption{Monte Carlo certification procedure (Upper bound)}
\label{algorithm:prob_cert_upper}
\begin{algorithmic}
\Function{ProbCertifyUpper}{$y^*$, $g$, $\mu_{\mX}$, $\upsilon^{(1)}$, $\upsilon^{(2)}$, $N_1$, $N_2$, $N_3$, $\alpha$}
  \State $\mZ^{(1)}, \dots, \mZ^{(N_1)} \gets$ \Call{Sample}{$\mu_{\mX}$, $N_1$}
  \State $\mathrm{count}_1 \gets \sum_{n=1}^{N_1} \indicator\left[g(\mZ^{(n)}) = y^*\right]$
  \State $\overline{p_{\mX,y'}} \gets$ \Call{UpperConfBound}{$\mathrm{count}_1$, $N_1$, $1 - \alpha$} \Comment{First bound}
  \State $n^* \gets \max \left\{n \mid \Call{BinPValue}{n, N_2, \leq, 1 - \overline{p_{\mX,y'}}} < \frac{\alpha}{2}\right\}$
  \State $V^{(1)}, \dots, V^{(N_2)} \gets$ \Call{Sample}{$\upsilon^{(2)}$, $N_2$}
  \State $R^{(1)}, \dots, R^{(N_2)} \gets$ \Call{SortAscending}{$\rho(V^{(1)}), \dots, \rho(V^{(N_2)})$}
  \State $\underline{\kappa} \gets R^{(n^*)}$ \Comment{Second bound}
  \State $V^{(1)}, \dots, V^{(N_3)} \gets$ \Call{Sample}{$\upsilon^{(1)}$, $N_3$}
  \State $\mathrm{count}_2 \gets \sum_{n=1}^{N_3} \indicator\left[\rho(V) \geq \underline{\kappa} \right]$
  \State \Return \Call{UpperConfBound}{$\mathrm{count}_2$, $N_3$, $1 - \frac{\alpha}{3}$} \Comment{Third bound}
\EndFunction
\end{algorithmic}
\end{algorithm}
Note that we still need to lower-bound threshold $\kappa$, because this increases the probability  $\Pr_{V \sim \upsilon^{(1)}} \left[\rho(V) \geq t\right]$
and thus leads to more pessimistic, sound multi-class certificates.

Finally note that, because the multi-class certificates are just a combination of two binary certificates, our evaluation of binary invariance-aware certificates in  in~\cref{section:main_experiments} is a good indicator of the performance of invariance-aware multi-class certificates.

\clearpage

\section{Inverse certificates}\label{appendix:inverse_certificates}
In~\cref{section:main_experiments_inverse}, we evaluate inverse certificates, i.e.\ compute the smallest prediction probability $p_\mathrm{min}$ for which robustness can still be certified, given the remaining certificate parameters such as $||\mDelta||_2$ or $||\mX||_2$.

In the case of the black-box randomized smoothing baseline, $p_\mathrm{min}$ can be calculated by solving
$\Phi\left( \Phi^{(-1)}(p) - \frac{||\mDelta||_2}{\sigma} \right) = \frac{1}{2}$ for $p$, which results in
$p_\mathrm{min} = \Phi\left(\frac{||\mDelta||_2}{\sigma}\right)$.

Similarly, the gray-box certificates for translation invariance has
$p_\mathrm{min} = \Phi\left(\frac{||\mDelta - \ones_N \overline{\mDelta}||_2}{\sigma}\right)$, where $\overline{\mDelta} \in \sR^{1 \times D}$ are the column-wise averages.

The tight gray-box certificates involving rotation invariance do not have closed-form analytic expressions.
Instead, they are of the form
\begin{align*}
    \min_{h \in \sH_\sT}\Pr_{\mZ \sim \mu_{\mX'}}\left[h(\mZ) = y^* \right]] = &\Pr_{V \sim \upsilon^{(1)}} \left[\rho(V) \leq \kappa\right] \\
    \text{with } \kappa \in \sR \text{ such that } &\Pr_{V \sim \upsilon^{(2)}} \left[\rho(V) \leq \kappa\right] = p_{\mX,y^*}.
\end{align*}
Here, an inverse certificate can be computed by first finding a threshold $\kappa$ such that $\Pr_{V \sim \upsilon^{(1)}} \left[\rho(V) \leq \kappa\right] = \frac{1}{2}$ and then  evaluating $\Pr_{V \sim \upsilon^{(2)}} \left[\rho(V) \leq \kappa\right]$.
That is,
\begin{align*}
    p_\mathrm{min} = &\Pr_{V \sim \upsilon^{(2)}} \left[\rho(V) \leq \kappa\right] \\
    \text{with } \kappa \in \sR \text{ such that } &\Pr_{V \sim \upsilon^{(1)}} \left[\rho(V) \leq \kappa\right] = \frac{1}{2}.
\end{align*}
To ensure a fair comparison with our baselines, we modify the Monte Carlo certification procedure proposed in~\cref{appendix:monte_carlo_certification} to compute a probabilistic upper bound on $p_\mathrm{min}$ that holds with high probability $1-\alpha$.

\begin{algorithm}[H]
\caption{Monte Carlo inverse certification procedure}
\label{algorithm:inverse_prob_cert}
\begin{algorithmic}
\Function{InverseProbCertify}{$\upsilon^{(1)}$, $\upsilon^{(2)}$, $N_1$, $N_2$, $\alpha$}
  \State $n^* \gets \min \left\{n \mid \Call{BinPValue}{n, N_1, \leq, 0.5} < \alpha\right\}$
  \State $V^{(1)}, \dots, V^{(N_2)} \gets$ \Call{Sample}{$\upsilon^{(1)}$, $N_1$}
  \State $R^{(1)}, \dots, R^{(N_2)} \gets$ \Call{SortAscending}{$\rho(V^{(1)}), \dots, \rho(V^{(N_2)})$}
  \State $\overline{\kappa} \gets R^{(n^*)}$ \Comment{First bound}
  \State $V^{(1)}, \dots, V^{(N_2)} \gets$ \Call{Sample}{$\upsilon^{(2)}$, $N_3$}
  \State $\mathrm{count}_2 \gets \sum_{n=1}^{N_2} \indicator\left[\rho(V) \leq \overline{\kappa} \right]$
  \State \Return \Call{UpperConfBound}{$\mathrm{count}_2$, $N_2$, $1 - \frac{\alpha}{2}$} \Comment{Second bound}
\EndFunction
\end{algorithmic}
\end{algorithm}
Note that, different from~\cref{appendix:monte_carlo_certification} we compute a probabilistic upper bound on threshold $\kappa$ by finding the smallest order statistic $R^{(n)}$ such that $\left[\rho(V) \leq R^{(n)}\right] \geq \frac{1}{2}$ holds with high probability.

\clearpage

\section{Parameter space of the tight certificate for rotation invariance in 2D}\label{appendix:tight_certificate_parameters}
Our tight certificates for 2D rotation invariance depend on $||\mX||_2$, $||\mDelta||_2$,
as well as parameters
$\epsilon_1 = \left\langle \mX, \mDelta \right\rangle_\mathrm{F}$, 
    $\epsilon_2 = \left\langle \mX \mR\left(-\frac{\pi}{2}\right)^T, \mDelta \right\rangle_\mathrm{F}$,
    which capture the orientation of the perturbed pointcloud, relative to the clean point cloud.
    
In this section, we determine the feasible range of $\epsilon_1$ and $\epsilon_2$ and calculate the two pairs of values corresponding to adversarial rotations, which is relevant for our experiments in~\cref{section:main_experiments_inverse}.

\subsection{Feasible parameter range}
Let $\hat{\mX} = \frac{\mX}{||\mX||_2}$.
Then, parameters $\epsilon_1$ and $\epsilon_2$ can be equivalently stated as follows:
\begin{align*}
    &\epsilon_1 =
\left\langle \mX, \mDelta \right\rangle_\mathrm{F}
= \mathrm{vec}(\mX)^T \mathrm{vec}(\mDelta)
= \mathrm{vec}(\hat{\mX})^T \mathrm{vec}(||\mX||_2 \mDelta),
\\
&\epsilon_2 =
\left\langle \mX\mR\left(-\frac{\pi}{2}\right)^T, \mDelta \right\rangle_\mathrm{F}
= \mathrm{vec}\left(\hat{\mX}\mR\left(-\frac{\pi}{2}\right)^T\right)^T \mathrm{vec}(||\mX||_2 \mDelta).
\end{align*}
Note that  $\mathrm{vec}(\hat{\mX})$ and $\mathrm{vec}\left(\hat{\mX}\mR\left(-\frac{\pi}{2}\right)^T\right)$
have norm $1$ and are orthogonal to each other.
Thus, they are the first two elements of an orthonormal basis of $\sR^{2N}$.
Parameters $\epsilon_1$ and $\epsilon_2$ are projections of $\mathrm{vec}(||\mX||_2 \mDelta)$ onto the first two elements of this basis.
Basis changes preserve the norm of vectors.
Thus, the values $\epsilon_1,\epsilon_2$ must fulfill 
$\sqrt{\epsilon_1^2 + \epsilon_2^2} \leq ||\mX||_2 \cdot ||\mDelta||_2$.

\subsection{Adversarial rotations}
\textbf{Known rotation angle.} Consider an arbitrary $\mX \in \sR^{N \times 2}$.
Further consider an adversarially perturbed $\mX' = \mX \mR(\theta)^T$ that is the result of rotating all rows of $\mX$ by angle $\theta$. Let $\mDelta = \mX' - \mX$.
Recall that, for any vector $\va, \vb \in \sR^D$, $\va^T \vb^T = ||\va||_2 ||\vb||_2 \cos(\angle\va\vb)$.
Therefore
\begin{align*}
    \epsilon_1
    =
    \left\langle \mX, \mDelta \right\rangle_\mathrm{F}
    = 
    \sum_{n=1}^N \mX_n^T (\mX'_n - \mX_n)
    = \sum_{n=1}^N ||\mX_n^T||_2^2 \cos(\theta) - ||\mX_n^T||_2^2
    = ||\mX||_2^2 (\cos(\theta) - 1),
\end{align*}
where the last equality follows from the definition of the Frobenius norm.
Similarly, we have
\begin{align*}
    \epsilon_2
    = ||\mX||_2^2 (\cos(\theta + \nicefrac{\pi}{2}) - \cos(\nicefrac{\pi}{2}))
    = - ||\mX||_2^2  \sin(\theta).
\end{align*}

\textbf{Unknown rotation angle.}
Now assume that we know $||\mX||_2$ and $||\mDelta||_2$, but do not know the rotation angle $\theta$.
We have 
\begin{align*}
    ||\mDelta||_2^2 = &\left\langle\mX \mR(\theta)^T - \mX,\mX \mR(\theta)^T - \mX \right\rangle_\mathrm{F}
    \\
    =
    &
    ||\mX \mR(\theta)^T||_2^2 + ||\mX||_2^2 - 2 \left\langle\mX,\mX \mR(\theta)^T  \right\rangle_\mathrm{F}
    \\
    =
    &
    ||\mX ||_2^2 + ||\mX||_2^2 - 2 ||\mX||_2^2  \cos(\theta) 
    \\
    =
    &
    2 ||\mX||_2^2 (1 - \cos(\theta)).
\end{align*}
Evidently, such an adversarial rotation is only possible if $||\mDelta|| \leq 2 ||\mX||$, which corresponds to a rotation by angle $\pi$.
Solving for $\theta$ yields
\begin{align*}
    \theta = \pm \arccos\left(1 - \frac{ ||\mDelta||_2^2}{2 ||\mX||_2^2}\right)
\end{align*}
Inserting into our formulae for $\epsilon_1$ and $\epsilon_2$ and using that $\sin(\arccos(a)) = \sqrt{1 - a^2}$ shows that
\begin{align*}
    \epsilon_1 &= -\frac{1}{2}||\mDelta||_2^2,\\
    \epsilon_2 &= \pm \frac{1}{2} \sqrt{||\mDelta||_2^2 \left( 4 ||\mX||_2^2 - ||\mDelta||_2^2\right)}.
\end{align*}


\end{document}